\documentclass[onecolumn,10pt]{asme2ej}

\usepackage{cite}
\usepackage{amsmath,amssymb,amsfonts}
\usepackage{textcomp}
\usepackage{subcaption}
\usepackage{tabularx, booktabs, graphicx, xcolor, multirow, mathrsfs, caption, array, mathtools, manyfoot, listings}
\usepackage[linesnumbered,ruled,vlined]{algorithm2e}
\usepackage{amsmath,amssymb}
\usepackage{bm}
\usepackage{url}
\usepackage{tikz}
\usetikzlibrary{arrows.meta, positioning, calc}
\usepackage{JMRSupport}

\def\BibTeX{{\rm B\kern-.05em{\sc i\kern-.025em b}\kern-.08em
    T\kern-.1667em\lower.7ex\hbox{E}\kern-.125emX}}

\begin{document}

\title{Multimodal Image-based Robotic Assistance and Guidance Environment (MIRAGE): A Novel Unified Framework for Surgical Navigation and Visualization
}
\author{%
{\normalfont\fontsize{11}{13}\selectfont
Ehsan Nasiri\thanks{Corresponding author: Ehsan Nasiri \texttt{ehsan.nasiri@dartmouth.edu}},\;
Yi Zheng, Michael A. Kokko, Xiaoyao Fan}\\
{\normalfont\fontsize{11}{13}\selectfont
William R. Warner, Keith D. Paulsen, Ryan J. Halter}\\[4pt]
{\normalfont\small Thayer School of Engineering at Dartmouth}\\
{\normalfont\small 15 Thayer Dr, Hanover, NH, USA}
}

\maketitle
\footnotetext{This work was supported in part by the Advanced Research Projects Agency for Health (ARPA-H) under award number 1AY2AX000043. The views and conclusions contained in this document are those of the authors and should not be interpreted as representing the official policies, either expressed or implied, of the United States Government.}
%
\begin{abstract}
This paper presents the computational architecture and quantitative preclinical evaluation of MIRAGE, a multimodal image-guided robotic surgery framework for real-time visualization and navigation. It integrates stereovision scene reconstruction, augmented-reality (AR) video overlay, quantified heads-up display (HUD) navigation, and intraoperative ultrasound within the da Vinci console's TilePro display (Intuitive Surgical). Additional modules include optical/electromagnetic tracker registration, stereo calibration, fiducial rigid registration with RMSE feedback, and 3D point-cloud reconstruction. Video I/O and rendering run in a C++ backend, while a Python GUI coordinates calibration, registration, reconstruction, visualization, and user interaction through an in-process zero-copy binding layer. The framework formalizes spatial transforms, module states, and synchronization of heterogeneous video and tracking streams. Preoperative calibration uses synchronized checkerboard image pairs and optimization, with estimated intrinsics, extrinsics, and rectification reused intraoperatively. MIRAGE was validated in phantom experiments across da Vinci S,Si, Xi, and SP platforms. Across clinically relevant working distances, mean 3D reprojection error was 0.6mm, with best 0.1mm RMSE at 100mm on Xi.  Results support MIRAGE as a preclinical technical-validation platform for accurate, responsive, and extensible multimodal surgical guidance.
\end{abstract}
\section {Introduction}
Robotic surgery has been shown to improve patient outcomes, with lower morbidity rates compared to traditional open surgery \cite{george2018,morris2005,Lanfranco2004,Reddy2023,Chuchulo2023,UFUrology2023 }.
Image-guided surgery, also known as surgical navigation, is another form of computer-assisted intervention designed to give surgeons real-time intraoperative guidance. Its purpose is to enable precise and efficient access to target regions, support surgical planning, and minimize the risk of damage to surrounding critical structures\cite{Bessen2021,Cleary2010,su2022s}.
 Imaging modalities integrated with surgical robots include fluorescence imaging, augmented-reality overlays, and navigational displays~\cite{vanLeeuwen2024,Wendler2021}. In urologic surgery for example, image-guided robotic systems employing near-infrared fluorescence imaging and augmented reality overlays have enhanced visualization of tumors and critical anatomy, improving surgical decision-making and functional outcomes \cite{Navaratnam2018, Porpiglia2019}. Similarly, in transoral robotic (TORS) procedures, the introduction of intraoperative imaging and navigational augmented reality helps surgeons overcome confined operative spaces and the lack of haptic feedback, thereby increasing the accuracy of tumor resection and safety near vital structures \cite{Pratt2018}. These advances, underscore a broader trend toward more precise, image-enhanced robotic interventions that improve oncologic and functional results for patients \cite{vanLeeuwen2024}.\par
Recent development in image-guided and robotic surgery have explored a wide spectrum of visualization strategies, ranging from integral videography (IV) autostereoscopic displays for 3D overlays without head-mounted devices \cite{liao2010} to goggle-based and monitor-based augmented reality (AR) systems that project preoperative models into the surgical field \cite{chan2021,schiavina2021}. In urology, numerous groups have applied computer vision and machine-learning techniques to automate 3D model registration during robotic partial nephrectomy \cite{Amparore2024,Piana2024,Giannone2021}, while systematic reviews highlight the broader use of extended reality (AR/VR) tools for planning, navigation, and education \cite{Checcucci2024,Manfredi2022}. 
Similar approaches have been extended to TORS to assist with tumor localization \cite{Chan20, Shi24} and to robotic prostatectomy, where AR improved anatomical orientation but remained hindered by organ motion \cite{schiavina2021}. Across these studies, the main benefits include enhanced depth perception, improved anatomical visualization, and reduced surgeon workload, whereas persistent drawbacks involve soft-tissue deformation, registration inaccuracy, high computational demands, hardware costs, and limited large-scale clinical validation.\par
Many image-guided robotic surgery systems are still evaluated as isolated demonstrations rather than integrated, time-synchronized navigation frameworks. This paper asks whether heterogeneous imaging, instrument tracking, modeling, reconstruction, and overlay streams can be synchronized, fused, and rendered within a unified robotic surgery framework while preserving measurable navigation accuracy and real-time responsiveness across multiple robotic platforms. \par
Prior multimodal guidance systems in robotic surgery (e.g., \cite{shapey2021}) have typically addressed only subsets of the sensing/visualization stack and left key integration gaps. In robotic prostatectomy, for example, Ukimura et al. \cite{Ukimura2014} reconstructed TRUS-derived, patient-specific 3D models and displayed them on the da Vinci console via TilePro, reporting favorable margin outcomes but relying on preoperative data and static alignment. Samei et al.\cite{Samei2020} integrated preoperative MRI with intraoperative 3D TRUS and live 2D TRUS in a “partial AR” scene shown beneath the endoscope feed on the surgeon’s console, improving situational awareness yet still constrained by registration quality and without optical/EM tracking fusion. Fusing optical and EM tracking instrument provides complementary robustness and tighter, continuous pose estimation of the endoscope and TRUS probe in a common frame, improving overlay accuracy under real operative dynamics. In \cite{Song2024} authors registered ultrasound/photoacoustic images   to the da Vinci stereo camera via an arc-to-line optimizer and achieved sub-centimeter target registration error (TRE) in simulation/bench experiments, but the workflow remains preclinical and depends on photoacoustics hardware. Meza et al.\cite{Meza2021} fused freehand ultrasound with structured-light surface capture (optically tracked) to one frame with sub-mm inter-modality error, demonstrating fusion accuracy but outside the robotic console and without endoscopic integration, so the surgeon is not able to see/fuse the modalities natively in the binocular console view. Classic stereo-vision pipelines (e.g., Stoyanov\cite{Stoyanov2010}) deliver real-time 3D tissue reconstruction from robotic endoscopy but are vision-only and lack fusion with other modalities (e.g., without any alternative imaging/sensing modules). In contrast, we introduce a unified, integrated framework that fuses preoperative stereo calibration, dual optical–EM tracking within a heads-up display(HUD)-TilePro navigation-visualization interface, live 3D stereo reconstruction (point clouds), 3D AR overlay, and live ultrasound streaming.  This integration supports dynamic intraoperative registration within a console workflow. Component calibration accuracy and end-to-end target accuracy are distinct endpoints, since the latter also depends on registration, tracking, motion, and temporal association.

The novelty of this work is not the individual use of stereo calibration, rigid registration, learned disparity estimation, or AR rendering, which are established in the literature. Instead, the contribution is a unified, time-aware, and quantitatively evaluated architecture that coordinates these components within a single console-attached robotic surgery workflow. By defining cross-modal states, timestamp association, transform propagation, buffer ownership, synchronization rules, and evaluation metrics, the framework enables reproducible testing of multimodal surgical guidance under realistic system-level constraints.

Recent consensus guidance recommends evaluating surgical robotics through staged translation rather than technical feasibility alone, spanning device performance, usability, patient acceptability, health-system impact, and long-term monitoring (IDEAL)~\cite{Marcus2024}. This is directly relevant to multimodal image-based robotic assistance and guidance environment (MIRAGE), where value depends not only on accuracy and frame rate, but also on workflow integration, user trust, failure mitigation, and reproducible evaluation across platforms and procedures.

Prior systems that utilize C++-Python software frameworks largely follow three major patterns: 
(1) Code-generated monoliths (e.g., large simulators) auto-generate both the C++ API and the pybind11 \cite{pybind11} glue to form a single expansive Python module; this scales to evolving engines but introduces a generator/continuous integration (CI) toolchain and is not tailored to low-latency, heterogeneous device streams \cite{Husar2024}. 
(2) Custom-designed interface over a single high-performance library deliver near–C++ speed, but typically bind \emph{one} math-control stack rather than multiple device-module subsystems \cite{Adorno2021}. 
(3) Thin Python fronts over real-time C++ processes (e.g., shared-memory or bursting-mode toolboxes) keep hard real-time loops out-of-process in C++ and expose a minimal Python API. This helps determinism but adds orchestration and deployment complexity beyond the process boundary \cite{Berenz2021}. 

In the proposed multi-threaded multimodal system, most high-level modalities are implemented in Python, while video acquisition, rendering, calibration support, and selected hardware-facing modules are implemented in C++ and bound \emph{in-process} via a custom-developed binding interface into a single Python orchestrator and GUI. The contribution is not the use of pybind11 alone, but the surrounding integration method: explicit buffer ownership, timestamped frame exchange, mutex-protected shared state, asynchronous worker management, and a unified state machine that coordinates C++ and Python modules. This design binds multiple heterogeneous C++ subsystems alongside Python-native modalities under one event loop, prioritizes in-process frame access over out-of-process shared memory, and exposes C++ frame buffers to Python through the buffer protocol when copy-free access is possible. Long-running C++ operations are isolated from the GUI event loop through worker threads and status callbacks. The system maintains a consistent timebase and timestamping policy to support Python-side data fusion and quantitative synchronization analysis.
\begin{figure}[!h]
 \centering
 \includegraphics[width=0.96\linewidth,height=0.64\textheight,keepaspectratio]{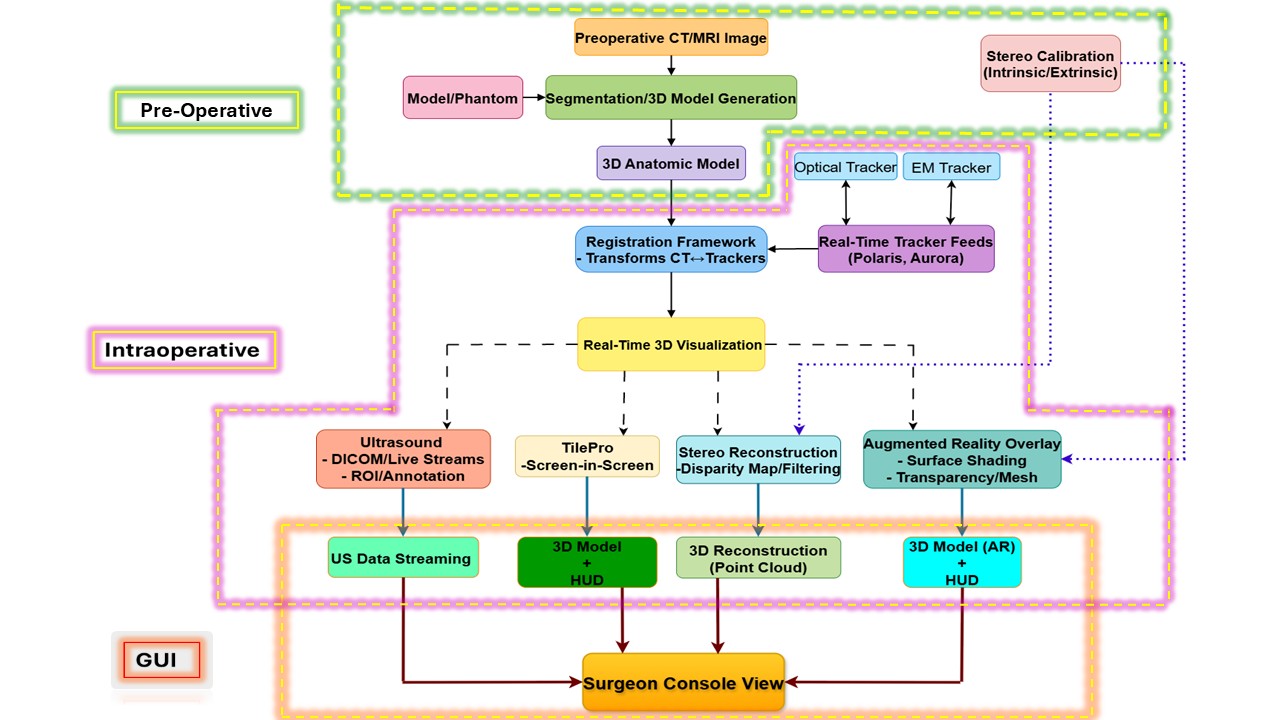}
 \caption{The overview of the framework pipeline.}
 \label{fig:block_diag}
\end{figure}

The binding layer connects C++ acquisition and rendering to the Python interface through timestamped image and pose exchange, shared configuration, and worker-status callbacks.

The formulation brings the spatial transforms, timestamp association, and module interfaces into a common description of the integrated workflow. Registration and calibration use established methods~\cite{Arun1987,Tsai1989,Zhang2000,Stoyanov2010}; the integration analysis concerns their coupled spatial, temporal, and configuration dependencies.

Compared with general-purpose robotic middleware and application frameworks such as the robot operating system (ROS/ROS2), the da Vinci research kit (dVRK) or NVIDIA Isaac, MIRAGE is deliberately implemented as a single workstation, console-attached framework focused on image guidance rather than robot actuation. In our configuration, the robot is driven entirely by its native clinical console, while MIRAGE concentrates on integrating stereo video, tracking, and multimodal visualization-navigation on the same machine. Introducing an additional middleware layer
would add serialization, inter-process communication, and dependency-management overhead without clear benefit for this tightly coupled setup.The single-process design prioritizes direct frame access and a common application state on the console workstation. Related infrastructure includes OpenIGTLink for data exchange~\cite{Tokuda2009}, Plus for tracked-image acquisition and calibration~\cite{Lasso2014}, 3D Slicer for extensible medical-image computing~\cite{Fedorov2012}, and ROS~2 for communication and component composition~\cite{Macenski2022}. Table~\ref{tab:software_framework_comparison} compares these roles with the application scope of MIRAGE. These complementary roles motivate comparisons with equivalent tasks, hardware, data-transfer policies, and timing boundaries.

\begin{table}[!htbp]
\centering
\caption{Roles and integration scope of MIRAGE and related software frameworks. The comparison describes documented capabilities, not measured performance rankings.}
\label{tab:software_framework_comparison}
\small\setlength{\tabcolsep}{5pt}\renewcommand{\arraystretch}{1.2}
\begin{tabularx}{\linewidth}{>{\raggedright\arraybackslash}p{0.17\linewidth} >{\raggedright\arraybackslash}p{0.25\linewidth} >{\raggedright\arraybackslash}X}
\toprule
\textbf{Framework} & \textbf{Primary role} & \textbf{Scope relevant to multimodal guidance}\\
\midrule
OpenIGTLink~\cite{Tokuda2009} & Communication protocol & Data exchange and interoperability across image-guided therapy components.\\
\addlinespace
Plus~\cite{Lasso2014} & US-guided intervention toolkit & Tracked ultrasound acquisition, calibration, reconstruction, and streaming.\\
\addlinespace
3D Slicer~\cite{Fedorov2012} & Medical-image computing platform & Provides visualization, segmentation, registration, and an extensible environment for image-analysis applications.\\
\addlinespace
ROS2~\cite{Macenski2022} & Robotics middleware & Provides communication and component composition for robotic applications; transport and quality-of-service settings depend on the application configuration.\\
\addlinespace
\textbf{MIRAGE} & Multimodal surgical image guidance framework & Unifies multimodal calibration, stereo vision, EM/optical tracking, tracked ultrasound, AR/HUD guidance, and 3D stereo reconstruction within a unified multimodal image-guidance framework.\\
\bottomrule
\end{tabularx}
\end{table}

The contributions of this research paper are summarized as follows:
\begin{itemize}
    \item \textbf{[1.]} A novel time-synchronized multimodal robotic surgery architecture that formalizes the coupling among stereo video, optical/electromagnetic tracking, AR overlay, stereo reconstruction, ultrasound, HUD navigation, and surgeon-console visualization within one state-driven framework.
    
    \item \textbf{[2.]} A novel cross-language real-time integration method that defines explicit function mappings, buffer ownership, synchronization invariants, thread-safety rules, and state-transition models between a C++ video/rendering backend and Python-based navigation, reconstruction, registration, and GUI modules.
    
    \item \textbf{[3.]} A temporal association and interpolation strategy for heterogeneous surgical data streams operating at different update rates, including video frames, tracker poses, rendering loops, and GUI events.
    
    \item \textbf{[4.]} A quantitative evaluation protocol for multimodal image-guided robotic surgery systems, including calibration/reprojection error, fiducial registration error, overlay reprojection error, stereo reconstruction error, end-to-end response time, synchronization error, and platform-specific repeatability.
    
    \item\textbf{[5.]} A comparative system-level analysis against representative prior image-guided robotic surgery approaches, identifying which capabilities are integrated within MIRAGE and which remain limitations.

    \item \textbf{[6.]} Preclinical validation with phantom experiments across the da Vinci S, Si, Xi, and SP platforms, demonstrating the feasibility of real-time multimodal guidance and defining the next evaluation steps needed for IDEAL-aligned translational assessment.
\end{itemize}\par
\subsection{Research Question and Evaluation Hypotheses}
This study asks whether heterogeneous imaging, tracking, reconstruction, and rendering modules can be integrated into a time-consistent robotic surgery navigation framework while preserving geometric accuracy and real-time responsiveness.

 The following hypotheses organize the evaluation of spatial accuracy, temporal association, frame access, and repeatability:
\begin{itemize}
\item \textbf{H1:} A unified transform chain preserves sub-millimeter to millimeter-scale calibration and registration accuracy across platforms and working distances.
\item \textbf{H2:} Timestamped multimodal association aligns video, tracking, rendering, and GUI events across different stream rates.
\item \textbf{H3:} In-process C++/Python integration enables real-time frame access and module switching without unnecessary frame duplication.
\item \textbf{H4:} The system supports reproducible quantitative evaluation of calibration, registration, overlay accuracy, reconstruction error, synchronization error, and latency.
\end{itemize}
Together, these hypotheses frame MIRAGE as a quantitatively evaluated multimodal navigation methodology, not merely a software interface.

\subsection{IDEAL-Aligned Evaluation Positioning}
This study is a preclinical technical and workflow validation, not evidence of clinical superiority or improved patient outcomes. Under IDEAL robotics~\cite{Marcus2024}, MIRAGE aligns primarily with stages 0--1, with early stage 2a elements, based on iterative development, integration with clinical robotic platforms, and phantom testing. Accordingly, evaluation focuses on integration, accuracy, real-time performance, reproducibility, and surgeon-facing visualization; clinical outcomes, learning curves, human factors, failure modes, health economics, and long-term surveillance remain future work.

\section{Multimodal Framework}
\begin{figure}[!t]
\centering
 \renewcommand{\thesubfigure}{\alph{subfigure}}
\begin{subfigure}[b]{\linewidth}
\centering
\includegraphics[width=0.95\linewidth,height=0.31\textheight,keepaspectratio]{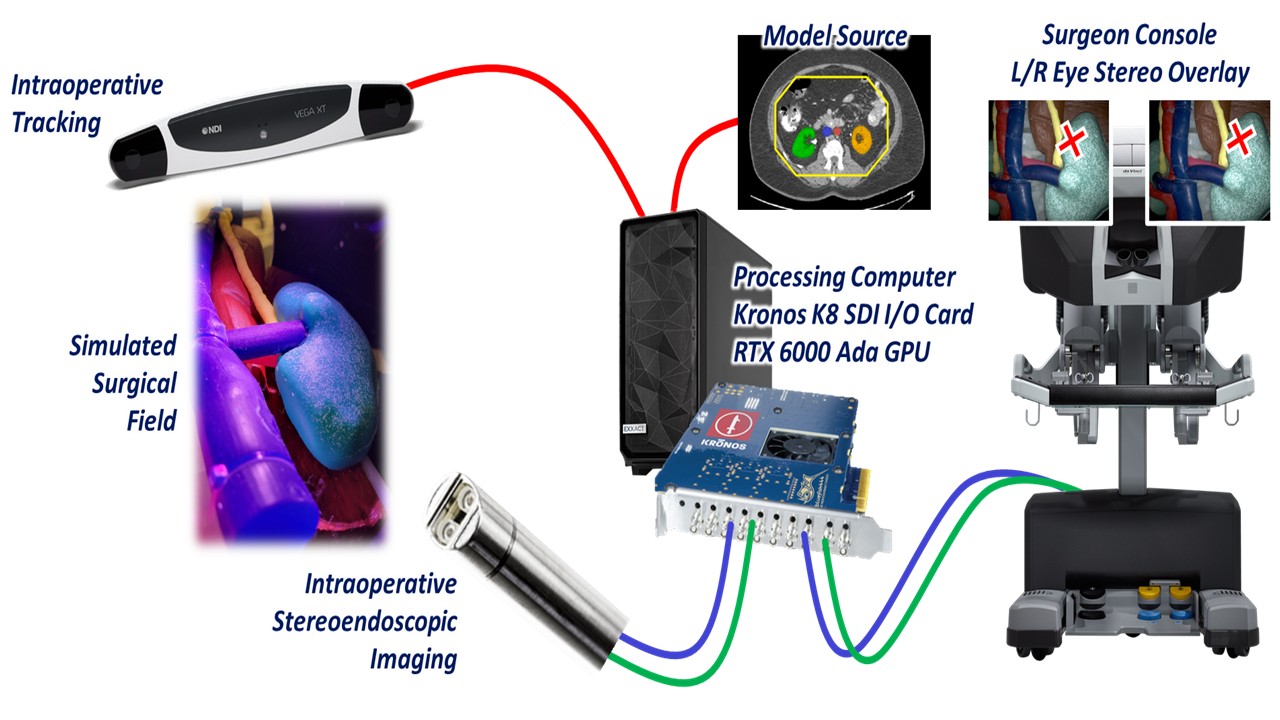}
 \caption{System integration and visualization: Real-time guidance architecture}
\label{fig:Multiple_views}
\end{subfigure}
\vfill
\begin{subfigure}[b]{\linewidth}
\centering
 \begin{adjustbox}{max width=0.95\linewidth,max totalheight=0.31\textheight}\input{Figures/RegistrationChain}\end{adjustbox}
\caption{The registration chain, where, \(\TF{A}{B}\) expresses the pose of the \{B\} frame in the \{A\} coordinate
system}
\label{fig:chain}
\end{subfigure}
\caption{System integration, calibration, registration and visualization pipeline.}
\label{fig:chain_pipeline}
\end{figure}
\subsection{Visualization - Navigation}
This section details the core principles and computational workflow underlying real-time visualization and navigation within the integrated image-guided surgery framework. As illustrated in Fig.~\ref{fig:block_diag}, the framework comprises preoperative (i) MR/CT imaging, (ii) stereo camera calibration, (iii) 3D segmentation, and (iv) generation of a patient-specific 3D anatomical model. Intraoperatively, the system integrates EM or optical tracking hardware with a real-time registration pipeline and provides a 3D visualization-navigation environment with multiple surgeon-selectable modes, all presented within a unified GUI on the surgeon's console. The upcoming subsections describe how spatial transformations are systematically chained to express the pose of every relevant component such as the instrument, tracking markers, and calibration references in a common coordinate system. For completeness, the mathematical formulation of these transformation chains, as well as their implementation in the registration and navigation pipelines, is presented. These processes provide the geometric basis for accurate model alignment, real-time scene rendering, and intuitive surgical guidance throughout all stages of the multimodal navigation system. Figure~\ref{fig:Multiple_views} depicts the main system architecture, and Fig.~\ref{fig:chain} illustrates the coordinate frames and transformation pathways.
The phantoms used for the experimental validation presented in this manuscript are shown in Fig.~\ref{fig:Phantom}.
\subsection{Registration Pipeline}
Throughout the paper, $\TF{A}{B}\in SE(3)$ maps homogeneous column coordinates from frame $\{B\}$ to frame $\{A\}$: $\bar{\mathbf{x}}_A=\TF{A}{B}\bar{\mathbf{x}}_B$, where $\bar{\mathbf{x}}=[\mathbf{x}^{\mathrm{T}},1]^{\mathrm{T}}$. Consequently, $\TF{A}{C}=\TF{A}{B}\TF{B}{C}$ and $\TF{B}{A}=(\TF{A}{B})^{-1}$. The rotation and translation blocks are $\RF{A}{B}\in SO(3)$ and $\TV{A}{B}\in\mathbb R^3$.

Figure~\ref{fig:chain} schematically illustrates the coordinate frames and transformation pathways underlying our navigation framework. For example, the transformation $\TF{PO}{LE}$ expresses the pose of the \emph{left eye} \{LE\} frame in the \emph{tracker} \{PO\} coordinate system, and is computed by chaining the transformations from the left eye to the \emph{checkerboard} \{CB\}, and then from the board to the tracker. 

Similarly, the transformation $\TF{PO}{BF}$ expresses the pose of the \emph{butterfly collar} \{BF\} frame in the tracker space. This is obtained by sequentially chaining the transformations from collar to left eye, left eye to checkerboard, and finally checkerboard to tracker. The systematic chaining of these transformations, as depicted by the arrows in Fig.~\ref{fig:chain}, enables every relevant component to be represented in a unified coordinate system for accurate model alignment and real-time navigation. The mathematical background underlying these registration pipelines is provided in \cite{Arun1987} 

In all cases, the chain of transformations follows the physical connection and registration path indicated by the red dashed arrows in Fig.~\ref{fig:chain}, ultimately allowing every frame of interest to be represented in the common tracker \{PO\} space,
\begin{align}
&\TF{PO}{LE} = \TF{PO}{CB}
\; \TF{CB}{LE}, \\[2ex]
&\TF{PO}{BF} = \TF{PO}{CB}
 \; \TF{CB}{LE}
\; \TF{LE}{BF}
\end{align}

\subsubsection{Optical Tracker}\label{op_regist}
Corresponding preoperative CT fiducials $\mathbf{p}_{CT}^{(i)}$ and optically tracked stylus samples $\mathbf{p}_{PO}^{(i)}$ are acquired through the GUI. Their least-squares rigid registration is,
\begin{align}
(\mathbf{R}_{\rm reg},\mathbf{t}_{\rm reg})
&=\arg\min_{\mathbf{R}_{\rm reg}\in SO(3),\,\mathbf{t}_{\rm reg}\in\mathbb{R}^{3}}
\sum_{i=1}^{N_{\rm fid}}
\left\|\mathbf{R}_{\rm reg}\mathbf{p}_{CT}^{(i)}+\mathbf{t}_{\rm reg}
-\mathbf{p}_{PO}^{(i)}\right\|^{2}, \\
\TF{PO}{CT}
&=
\begin{bmatrix}
\mathbf{R}_{\rm reg} & \mathbf{t}_{\rm reg}\\
\mathbf{0}_{1\times3} & 1
\end{bmatrix}
\end{align}
Registration quality is reported as,
\begin{equation}
\mathrm{RMSE}=
\sqrt{\frac{1}{N_{\rm fid}}\sum_{i=1}^{N_{\rm fid}}
\left\|\mathbf{R}_{\rm reg}\mathbf{p}_{CT}^{(i)}+\mathbf{t}_{\rm reg}
-\mathbf{p}_{PO}^{(i)}\right\|^{2}}
\end{equation}
Here $N_{\rm fid}$ is the number of corresponding fiducial pairs. The GUI displays and saves the transform and the root mean square error (RMSE). The centroid computation, SVD solution, and reflection correction follow \cite{Arun1987}; these steps and the complete acquisition workflow are detailed in Appendix~\ref{app:registration_sync}.

\subsubsection{EM Tracker}
The electromagnetic pipeline acquires the corresponding fiducials using the Field Generator $\{FG\}$ and applies the similar SVD-based rigid registration described for the optical tracker. 
The saved registrations provide the inter-tracker transform
\begin{equation}
\TF{PO}{FG}
=
\TF{PO}{CT}\,
\TF{CT}{FG}
\end{equation}
The complete optical/EM registration algorithm is given in Appendix~\ref{app:registration_sync}.

\subsubsection{Real-Time Navigation Pipeline}
The navigation pipeline combines the GUI, optical (Polaris Vega\textsuperscript{\tiny{\textregistered}} XT; Northern Digital, Waterloo, ON) and electromagnetic (Aurora\textsuperscript{\tiny{\textregistered}}; Northern Digital, Waterloo, ON) tracking, and a dynamic visualization toolkit (VTK) scene. A background thread composes each live tool pose with the saved CT registration, smooths the pose, projects its rotational component back onto $SO(3)$, and updates the anatomical models, instruments, fiducials, and HUD metrics in the common CT frame. The filtering and $SO(3)$-projection equations are detailed in Appendix~\ref{app:registration_sync}, with the SVD formulation in \cite{Arun1987}.

\subsection{Temporal Synchronization and Stream Association}\label{sec:temporal_sync}
All video, tracking, rendering, ultrasound, and GUI samples are timestamped by a monotonic host clock. For a fixed frame pair, $\mathbf T_i=\TF{A}{B}(t_i)$ has rotation $\mathbf R_i$ and translation $\mathbf p_i$; the interpolated pose $\hat{\mathbf T}(t_v)$ uses the same frame pair. For a video timestamp $t_v$ bracketed by tracker samples $(\mathbf{R}_i,\mathbf{p}_i,t_i)$ and $(\mathbf{R}_j,\mathbf{p}_j,t_j)$, the synchronized pose is (with rotations interpolated through their associated unit quaternions by SLERP)
\begin{align}
\beta&=\frac{t_v-t_i}{t_j-t_i},\qquad t_i\leq t_v\leq t_j,\quad t_i<t_j,\\
\hat{\mathbf{T}}(t_v)
&=
\begin{bmatrix}
\mathrm{SLERP}(\mathbf{R}_i,\mathbf{R}_j,\beta) &
(1-\beta)\mathbf{p}_i+\beta\mathbf{p}_j\\
\mathbf{0}_{1\times3} & 1
\end{bmatrix}
\end{align}
If no pose lies within $\Delta t_{\max}$, the overlay is marked stale and is frozen or disabled according to the safety policy. 

For tracker samples with $t_i<t_j$ bracketing the video timestamp $t_v$, the temporal support is characterized by
\begin{equation}
a_v=\max(t_v-t_i,t_j-t_v),\qquad h_v=t_j-t_i.
\end{equation}
Here $a_v$ is the largest absolute source-sample offset and $h_v$ is the interpolation bracket width. These quantities describe the supporting observations; assigning $t_v$ to the interpolated pose does not measure physical synchronization error. Acquisition time, host-arrival time, clock offset, and any wait for the later sample represent separate timing contributions.
 
 Appendix~\ref{app:registration_sync} provides the complete stream-association formulation.
\begin{figure}
\centering
\includegraphics[width=0.96\linewidth,height=0.64\textheight,keepaspectratio]{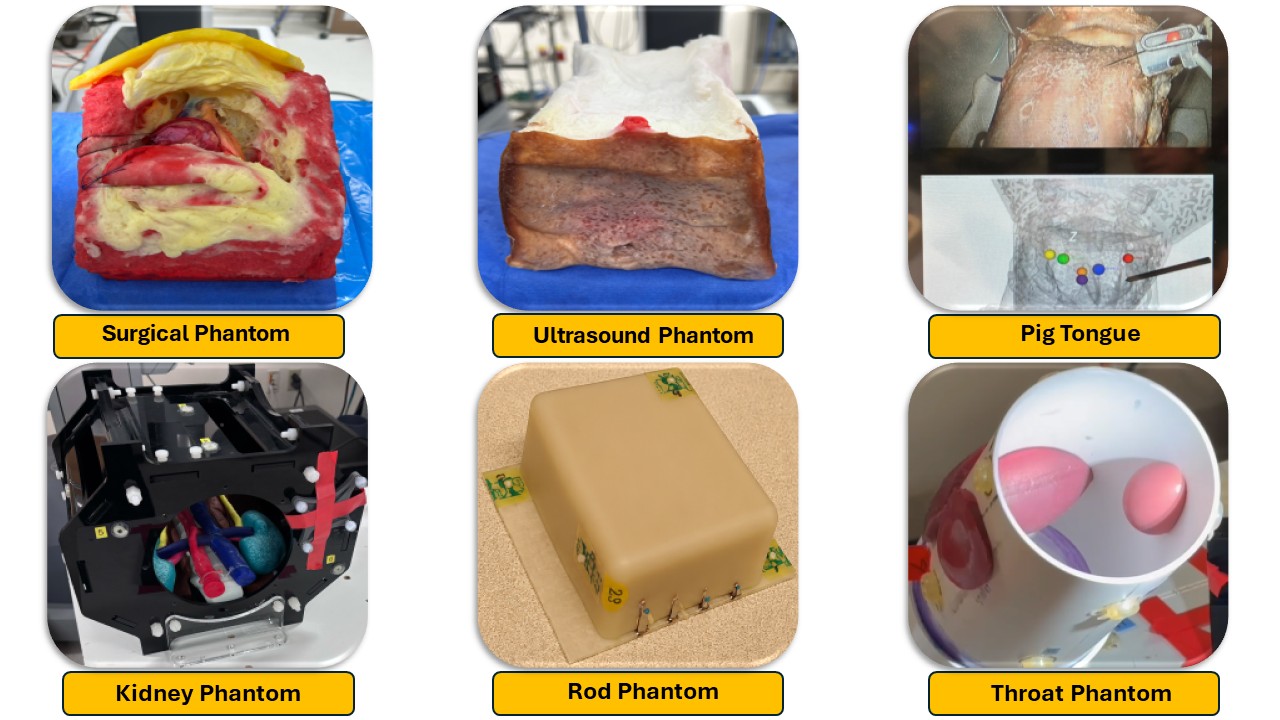}
\caption{Phantoms used in the experimental validation of this manuscript.}
\label{fig:Phantom}
\end{figure}
\begin{figure}[!t]
\centering
\renewcommand{\thesubfigure}{\alph{subfigure}}
\begin{subfigure}[b]{0.48\linewidth}
\centering
 \includegraphics[width=0.98\linewidth,height=0.25\textheight,keepaspectratio]{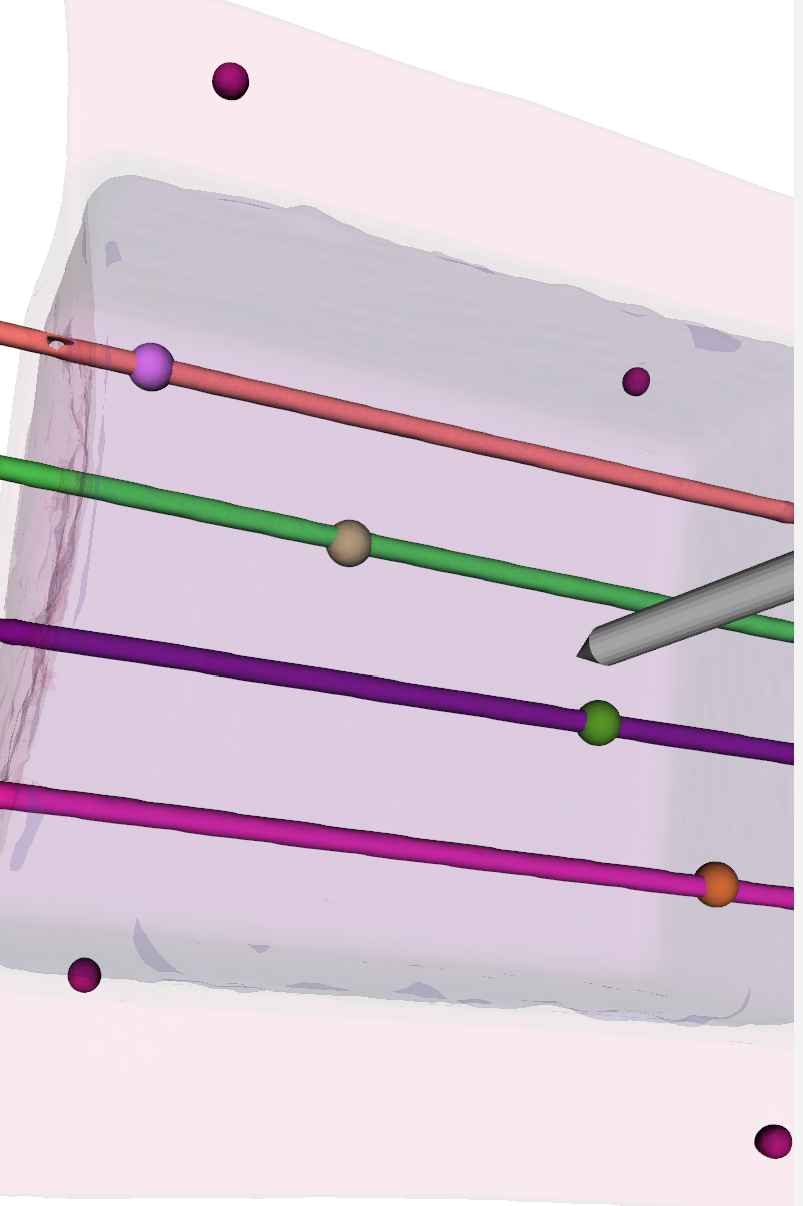}
\caption{CT-segmented 3D scene visualization of rod phantom model and EM-tracked needle}
 \label{fig:model_visual}
\end{subfigure}
 \hfill
\begin{subfigure}[b]{0.48\linewidth}
\centering
 \includegraphics[width=0.98\linewidth,height=0.31\textheight,keepaspectratio]{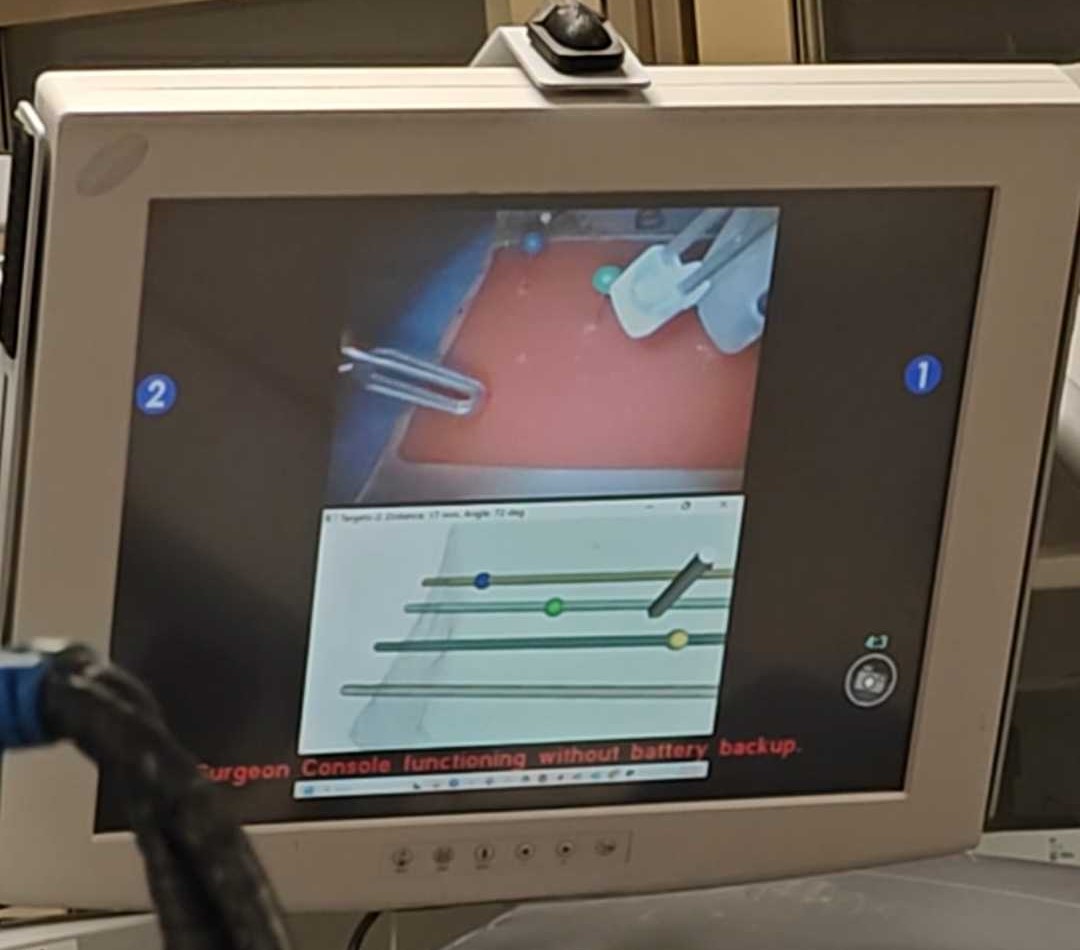}
 \caption{Tilepro visualization. Top: endoscopic surgical view. Bottom: Tracked and registered 3D model view}
 \label{fig:tilepro}
 \end{subfigure}
\caption{HUD-TilePro navigation-visualization on surgeon console. Note that the rod phantom (left) shows a set of embedded capillary tubes are not directly visible to the surgeon without use of the CT-segmented model (right).}
 \label{fig:tilpro-hud}
\label{fig:vis}
\end{figure}
\subsection{ Heads-Up Display (HUD)-TilePro}

The heads-up display framework provides tracked tool-to-target location guidance. Figure~\ref{fig:model_visual} shows the CT-segmented 3D rod phantom model and the EM-tracked needle. For each update cycle, the system automatically identifies the target closest to the instrument tip, using a combined score based on distance and approach angle. Only this best-matched target is fully highlighted in the 3D scene, while other targets are visually dimmed. The relevant guidance metrics are displayed on the VTK HUD console, ensuring the user receives immediate and intuitive feedback on navigation performance and instrument alignment. The full 3D scene and HUD outputs are refreshed in real time at rates up to 40Hz.

For real-time intraoperative feedback, the HUD computes and displays at each frame the distance and approach angle between the tracked tool and a selected target, as is explained in \cite{Shi24}, with several modifications and extensions. In particular, we introduce new filtering and smoothing techniques, as well as a new multi-criteria valuation metric for enhanced target highlighting.  Let $\mathbf{p}_{\rm target} \in \mathbb{R}^3$ denote the position of a target in CT space, and let the $4 \times 4$ homogeneous matrix $\TF{CT}{tool}$ describe the current pose of the surgical instrument in CT space. The tip position of the instrument is given by the translation component:
\begin{equation}
    \mathbf{p}_{\rm tip} = \TF{CT}{tool}[0:3,\, 3]
\end{equation}
The Euclidean distance from the instrument tip to the target, accounting for the target's physical radius $r_{target}$, is calculated as
\begin{equation}
    d_{\rm target} = \| \mathbf{p}_{\rm target} - \mathbf{p}_{\rm tip} \| - r_{target}
\end{equation}
where $\| \cdot \|$ denotes the Euclidean norm.

The instrument's approach angle is defined with respect to the vector from a fixed instrument-local wrist offset point to the target. The tool's local $z$-axis is obtained from the third column of the rotation part of $\TF{CT}{tool}$:
\begin{equation}
    \mathbf{z}_{tool} = \TF{CT}{tool}[0:3,\, 2]
\end{equation}
Let $\mathbf{p}_{\rm wrist}$ denote the instrument wrist offset point, computed as
\begin{equation}
    \mathbf{p}_{\rm wrist} = (\TF{CT}{tool} \; [0, -w_{\rm wrist}, 0, 1]^{\mathrm{T}})[0:3]
\end{equation}
where $w_{\rm wrist}$ is the offset (e.g., wrist length). The vector from the offset to the target is
\begin{equation}
    \mathbf{v}_{\rm target} = \mathbf{p}_{\rm target} - \mathbf{p}_{\rm wrist}
\end{equation}
The angle of approach $\theta_{\rm target}$ is then calculated as,
\begin{equation}
    \theta_{\rm target} = \arccos \left( \frac{ \mathbf{z}_{tool}^{\mathsf T}\mathbf{v}_{\rm target} }{ \| \mathbf{z}_{tool} \| \; \| \mathbf{v}_{\rm target} \| } \right) 
\end{equation}

To determine which target should be highlighted, all targets are evaluated using a combined normalized distance-angle score. For the $i$-th target, the score is defined as
\begin{equation}
s_{\rm target,i} = \left(\boldsymbol{\zeta}_i \boldsymbol{\zeta}_i^{\mathrm{T}} \right)^{1/2}
\end{equation}
where,
\begin{equation}
\boldsymbol{\zeta}_i =
\begin{bmatrix}
 d_{\rm target,i}/d_{\rm target,max} & \theta_{\rm target,i}/\theta_{\rm target,max}
\end{bmatrix}
\end{equation}
Here, $d_{\rm target,max}>0$ and $\theta_{\rm target,max}>0$ are normalization constants, with the same distance and angular units as their corresponding numerators. The target with the lowest score is selected, and its associated distance $d_{\rm target}$ and approach angle $\theta_{\rm target}$ are displayed as real-time HUD feedback, as shown in Fig.~\ref{fig:tilepro}.

This pipeline provides a quantitatively evaluable platform for registration assessment and navigation guidance in image-guided robotic surgery research, supporting both qualitative visualization and quantitative assessment of surgical accuracy. 

\subsection{Augmented Reality (AR) Video Overlay}\label{sec:AR_Overlay}
A real-time stereo overlay registers segmented preoperative CT anatomy to the tracked endoscopic scene and renders it directly over the surgical video. The framework was validated using multiple phantoms, including kidney, throat, and surgical phantoms, and supports surgeon-controlled visibility, mesh, and transparency modes.
\begin{figure}[!t]
\centering
 \renewcommand{\thesubfigure}{\alph{subfigure}}
\begin{subfigure}[b]{1\columnwidth}
    \centering
 \includegraphics[width=0.95\linewidth,height=0.31\textheight,keepaspectratio]{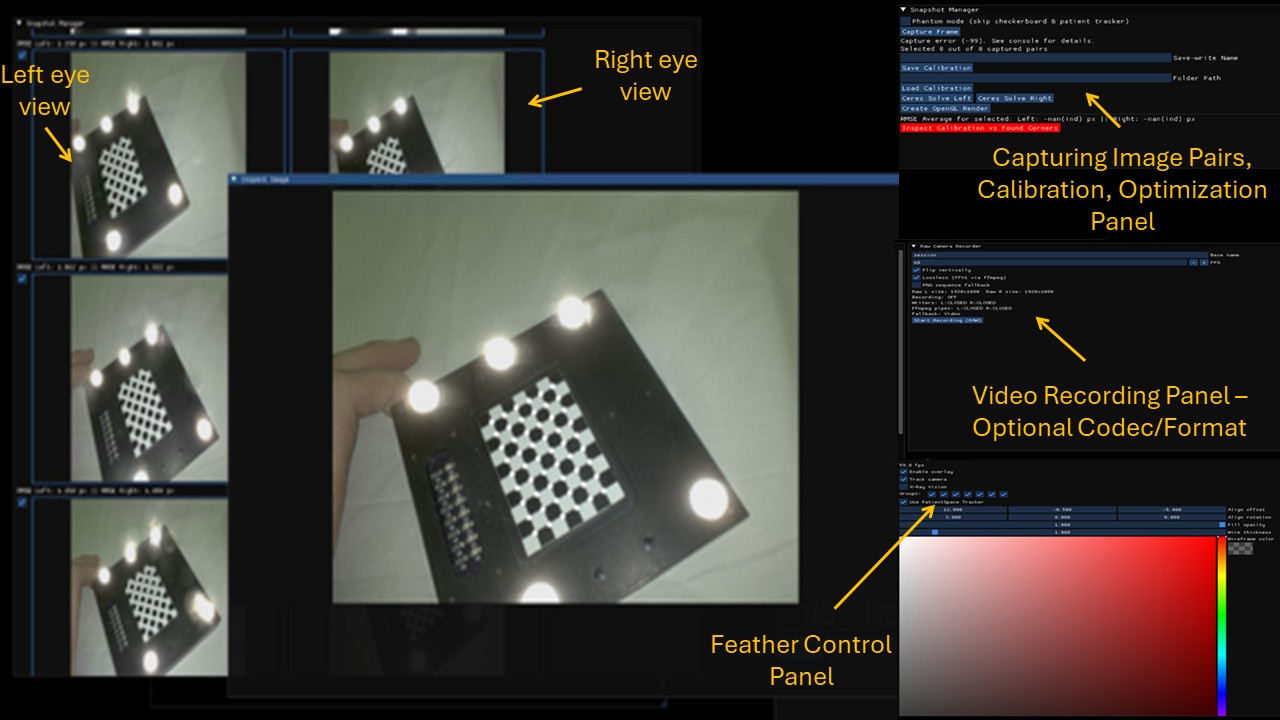}
 \caption{Stereo calibration framework includes capture of several poses of our calibration rig. Both stereo pairs (left and right eyes) and optical tracking data are capture simultaneously }
 \label{fig:Stereo_calibration}
\end{subfigure}
\vfill
\begin{subfigure}[b]{1\columnwidth}
  \centering
 \includegraphics[width=0.95\linewidth,height=0.31\textheight,keepaspectratio]{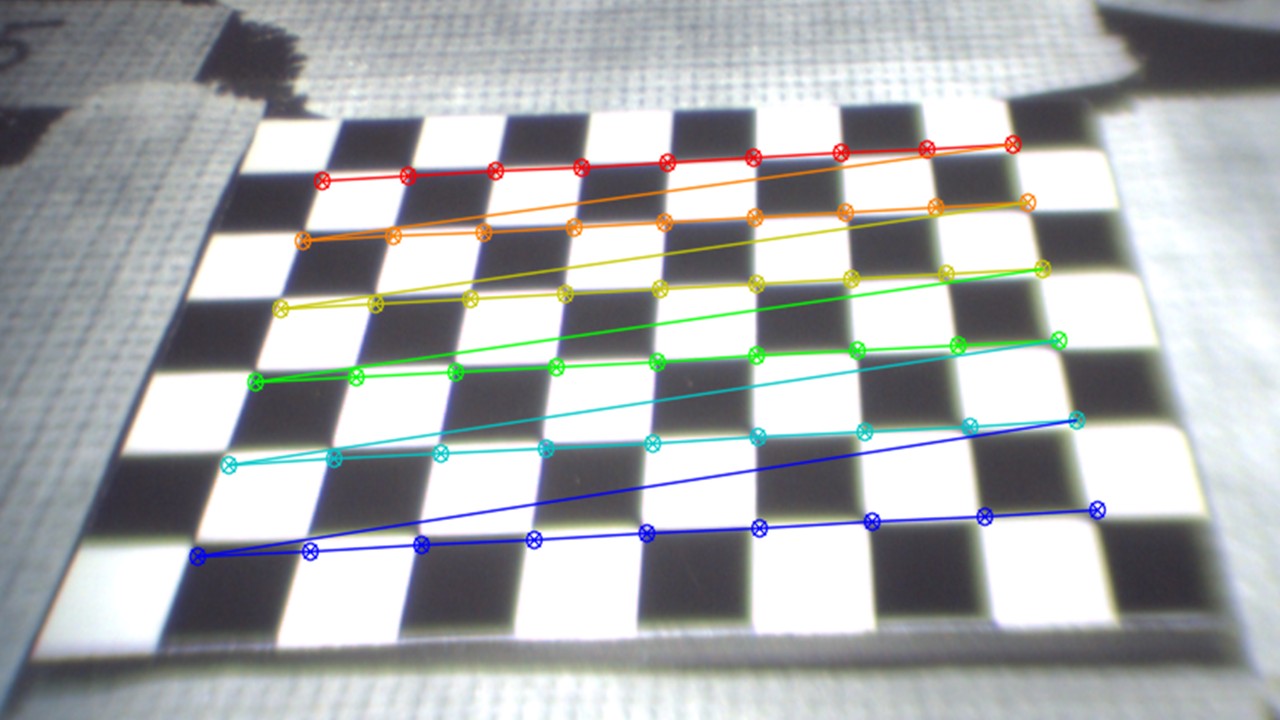}
 \caption{Errors for the camera as highlighted by matching of the checkerboard intersections}
 \label{fig:IDM_arch}
 \end{subfigure}
  \caption{Stereo calibration pipeline.}
 \label{fig:tilpro}
\label{fig:calibration_chekcerboard}
\end{figure}
An OpenCV stereo calibration estimates the intrinsic and tracked extrinsic parameters of both endoscope eyes using a custom checkerboard/retroreflector rig \cite{kokko2023} (Fig.~\ref{fig:calibration_chekcerboard}). Evaluation on da Vinci S and Xi cameras yielded reprojection RMSEs of approximately 6 and 2 pixels, respectively. Figure~\ref{fig:Stereo_calibration} illustrates the acquisition of stereo-image pairs and optical tracking data using the calibration rig, while Fig.~\ref{fig:IDM_arch} shows the corresponding checkerboard corner matches.
\subsubsection{System Overview} 
\begin{figure}
\centering
\includegraphics[width=0.96\linewidth,height=0.64\textheight,keepaspectratio]{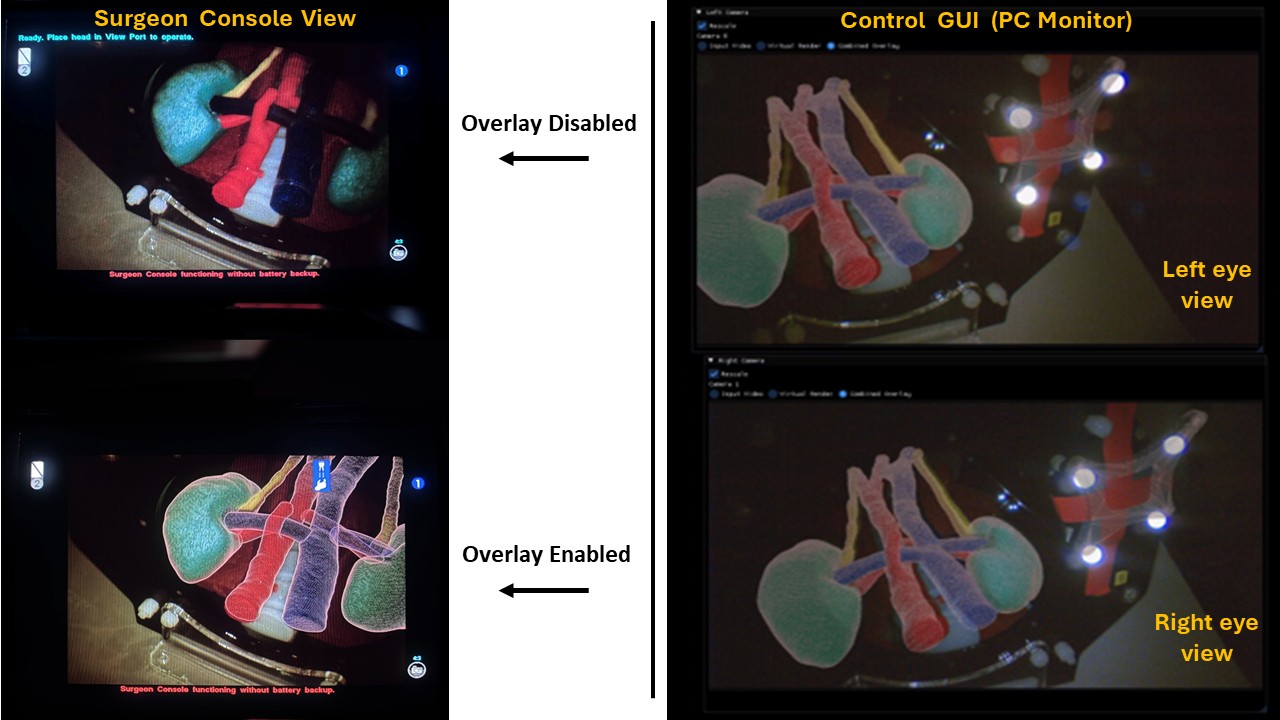}
\caption{The surgeon's console view (left) and overlay GUI (right) with the overlay toggled off (left top) and on (left bottom) for both the left and right endoscope eyes.}
\label{fig:ovelray_control}
\end{figure}
The pipeline combines optical tracking, YAML-loaded CT/tool registrations, calibrated stereo intrinsics/extrinsics, pose smoothing, OpenGL rendering, and SDI video I/O on an Exaact PC with a Kronos K8 card and RTX 6000 Ada GPU (Fig.~\ref{fig:chain_pipeline}). The composite CT-to-camera transform for either eye is,
\begin{equation}
\TF{CM}{CT}=\TF{CM}{PO}\,\TF{PO}{CT}
\end{equation}
Dynamic tool poses are updated from the current tracker measurement. The input, virtual-render, and overlay modes use,
\begin{equation}
\mathbf{I}_{\mathrm{display}}
=(1-\lambda)\mathbf{I}_{\mathrm{video}}
+\lambda\mathbf{I}_{\mathrm{render}}
\end{equation}
where the blend weight $\lambda\in[0,1]$ is adjusted from the GUI. Dedicated rendering, tracker, and Bluefish444 KRONOS K8\textsuperscript{\tiny{\textregistered}} video-I/O threads update both eye views in real time (Fig.~\ref{fig:ovelray_control}). Pose smoothing, camera-frustum construction, coordinate-transform composition, and real-time rendering are implemented within MIRAGE, while camera calibration and reprojection optimization follow the established formulation in \cite{Zhang2000}.

Appendix~\ref{app:ar_overlay} provides the complete pose-smoothing, camera-frustum, coordinate-transform, and rendering equations.
\subsection{Stereo Endoscope Calibration and Optimization Pipeline}\label{stereo_calib}
\label{sec:calib-opt}
\begin{figure}
\centering
\includegraphics[width=0.96\linewidth,height=0.64\textheight,keepaspectratio]{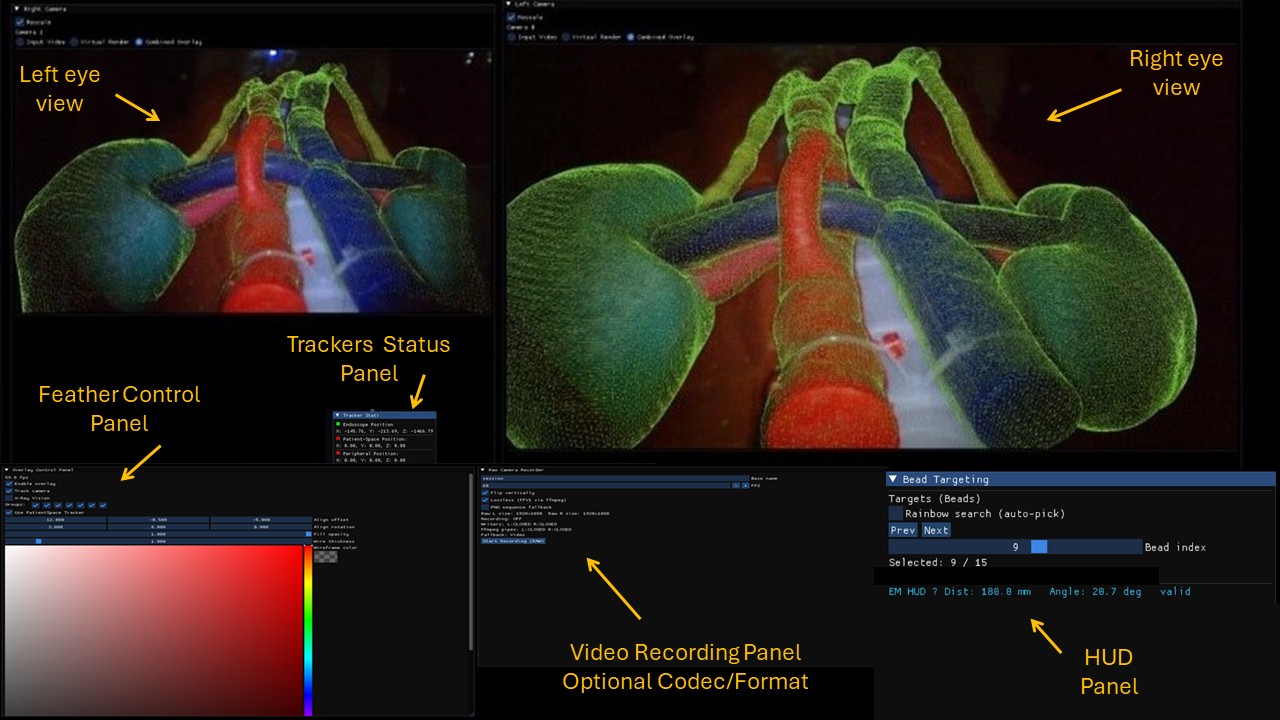}
\caption{Augmented reality 3D overlay (left-right endoscope eyes) displayed on the rendered kidney phantom model.}
\label{fig:ovelray_out}
 \vspace{-0.1cm}
\end{figure}
\paragraph{Calibration Model}
Synchronized stereo images and tracked endoscope/checkerboard poses are acquired at multiple rig poses. CLAHE, corner detection, and subpixel refinement provide observed checkerboard corners $\mathbf{c}_{i,k}^{(f)}$. Each checkerboard point $\mathbf{X}_k\in\mathbb R^3$ has homogeneous coordinates $\bar{\mathbf{X}}_k=[\mathbf{X}_k^{\mathrm{T}},1]^{\mathrm{T}}$ in $\{CB\}$ and is transformed into eye $CM_i$ (left/right eyes) for calibration frame $f$ as,
\begin{equation}
\bar{\mathbf{X}}_{i,k}^{(f)} = \TF{CM_i}{CB}^{(f)}\,\bar{\mathbf{X}}_k \label{eq:reporjection}
\end{equation}
and projected using the intrinsic and radial/tangential distortion parameters. The expanded camera projection and distortion model follows the established formulation in \cite{Zhang2000}.

The expanded projection model is given in Appendix~\ref{app:ar_overlay}.
\paragraph{Optimization}
The calibration parameters $\boldsymbol{\theta}_{\rm calib}=\{\mathbf{K}_i,\mathbf{D}_i,\TF{CM_i}{BF}\}_{i=1}$ are estimated by minimizing the reprojection error over all recorded frames and detected corners. The frame-dependent chain is $\TF{CM_i}{CB}^{(f)}=\TF{CM_i}{BF}(\TF{PO}{BF}^{(f)})^{-1}\TF{PO}{CB}^{(f)}$, using the recorded butterfly-marker and checkerboard poses:
\begin{align}
\mathbf{r}_{i,k}^{(f)}\bigl(\boldsymbol{\theta}_{\rm calib}\bigr)
  &= \mathbf{u}_{i,k}^{(f)} - \mathbf{c}_{i,k}^{(f)}                                            
= \begin{bmatrix}
u_{i,k}^{(f),\text{pred}}-u_{i,k}^{(f),\text{obs}}\\[4pt]
v_{i,k}^{(f),\text{pred}}-v_{i,k}^{(f),\text{obs}}
\end{bmatrix} \in \mathbb{R}^{2},        \\[1.2ex]
&\min_{\boldsymbol{\theta}_{\rm calib}}
      \sum_{f=1}^{N_{\rm cal}}
      \sum_{i=1}^{2}
      \sum_{k=1}^{N_{\mathrm{corners}}}
      \bigl\|
        \mathbf{r}_{i,k}^{(f)}\bigl(\boldsymbol{\theta}_{\rm calib}\bigr)
      \bigr\|_{2}^{2}
      \label{eq:optim1}
\end{align}

where $\mathbf{c}_{i,k}^{(f)}$ and $\mathbf{u}_{i,k}^{(f)}(\boldsymbol{\theta}_{\rm calib})$ are the observed and predicted pixel coordinates, respectively; $N_{\rm cal}$ is the number of calibration frames and $N_{\rm corners}$ the number of detected corners per frame. The two eye-specific parameter blocks are fitted using the same notation.

\subsubsection*{SE(3) Pose Optimization}
The fixed tracked-marker-to-eye extrinsic is parameterized by the six-dimensional twist $(\bm{\omega},\mathbf{v}_{\rm twist})$ and mapped to $SE(3)$ by,
\begin{equation}
\TF{CM_i}{BF} \;=\; \exp\!\Bigl(
\begin{bmatrix}
     [\bm{\omega}]_\times & \mathbf{v}_{\rm twist}\\[2pt]
     \mathbf{0}_{1\times3} & 0
  \end{bmatrix}
\Bigr)
\label{eq:optim2}
\end{equation}
where $[\,\cdot\,]_\times$ is the skew-symmetric operator; $\bm\omega\in\mathbb R^3$ and $\mathbf v_{\rm twist}\in\mathbb R^3$ are the rotational and translational twist coordinates for the selected eye. The residual block jointly updates the nine intrinsic/distortion parameters and six pose parameters; the logarithm map initializes the twist from the current transform.
\subsubsection*{Output and Evaluation}
The optimized intrinsics and extrinsics are exported in YAML format, and per-frame and average reprojection RMSEs are reported for each camera. Figure~\ref{fig:ovelray_out} shows representative left- and right-eye AR overlays of the kidney phantom.
\subsection{Ultrasound Streaming} \label{ultrasound}
This prototype module acquires BK Medical\textsuperscript{\tiny\textregistered} model 3000 ultrasound from DICOM files or a live OpenIGTLink stream. The GUI supports frame browsing, polygonal ROIs, labeled landmarks, live display, and marching-cubes visualization of multi-frame DICOM or cine-buffer pseudo-volumes. A bounded producer--consumer queue discards the oldest pending frame on overflow to prioritize low-latency display and provides timestamps for tracker synchronization.
\begin{figure}
\centering
\renewcommand{\thesubfigure}{\alph{subfigure}}
\begin{subfigure}[b]{\textwidth}
\centering
\includegraphics[width=0.95\linewidth,height=0.31\textheight,keepaspectratio]{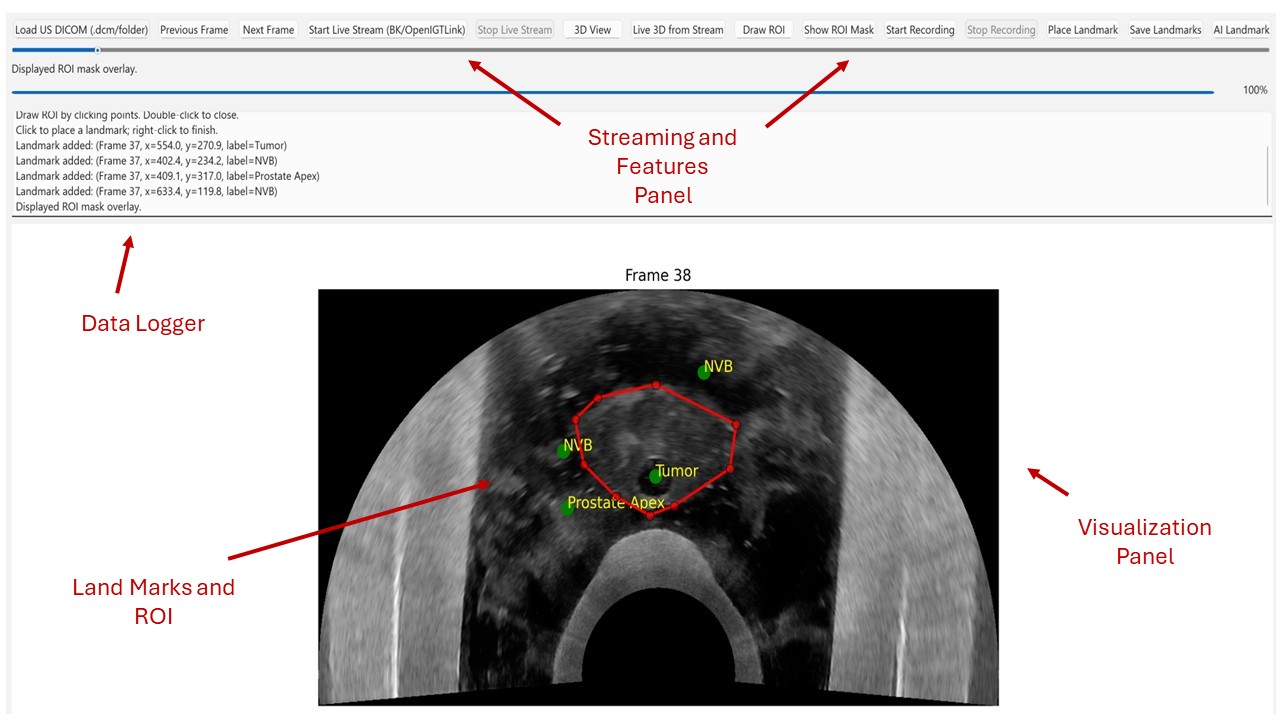}
\caption{Ultrasound GUI with live stream, ROI polygon, and landmarks}
\label{fig:US_gui}
\end{subfigure}
\medskip
\begin{subfigure}[b]{\textwidth}
\centering
\includegraphics[width=0.95\linewidth,height=0.31\textheight,keepaspectratio]{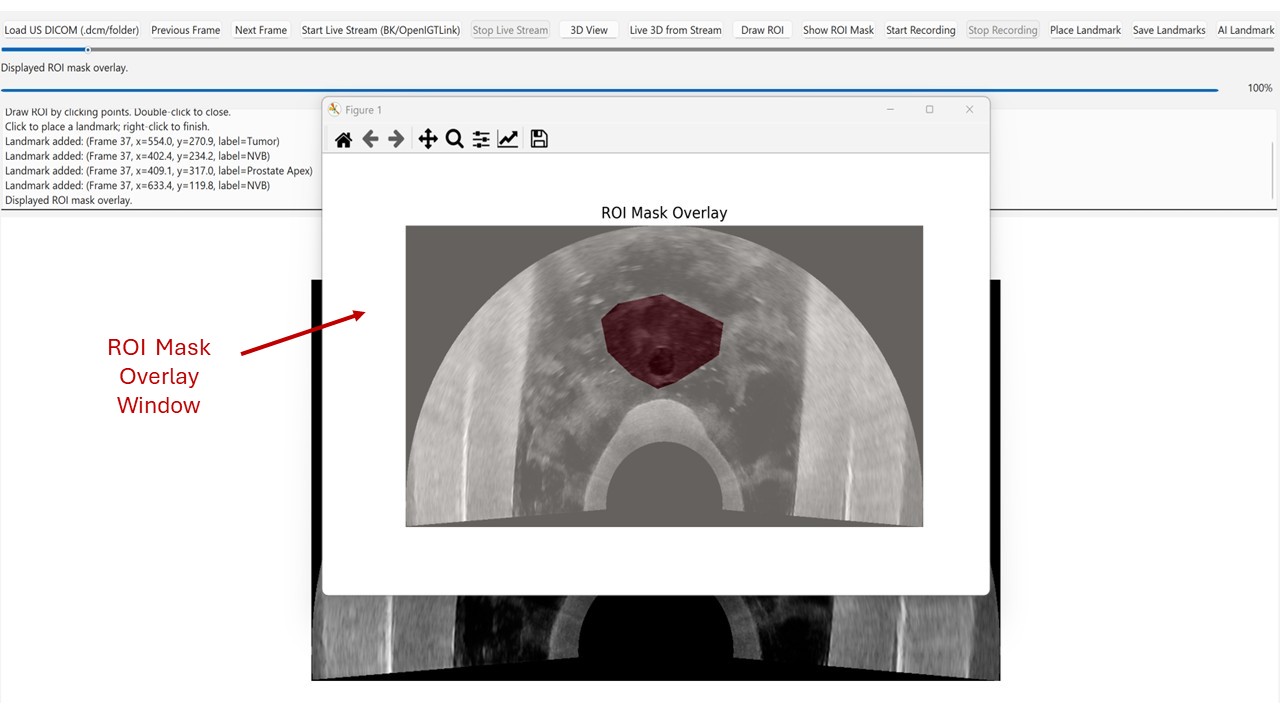}
\caption{ROI mask overlay}
\label{fig:US_roi}
\end{subfigure}
\caption{Ultrasound mode: acquisition, interaction, and visualization.}
\label{fig:US_mode_figs}
\end{figure}
\paragraph{Acquisition and Coordinates}
DICOM slices form $\mathcal{V}_{\rm US}\in\mathbb{R}^{N_{\rm US}\times H_{\rm img}\times W_{\rm img}}$ with row, column, and slice spacing $(s_r,s_c,s_z)$ in millimeters, where $N_{\rm US}$ is the number of slices and $H_{\rm img},W_{\rm img}$ are the image height and width in pixels, while live IMAGE messages provide timestamped frames $\mathbf{I}_t$ and a recent cine buffer. For pixel $(c,r)$ and slice $k$, the metric coordinate in the ultrasound frame is
\begin{equation}
\bar{\mathbf{x}}_{US} = 
\begin{bmatrix}
x\\y\\z\\1
\end{bmatrix}
=
\begin{bmatrix}
s_c & 0   & 0 & 0\\
0   & s_r & 0 & 0\\
0   & 0   & s_z & 0\\
0   & 0   & 0 & 1
\end{bmatrix}
\!\!
\begin{bmatrix}
c-c_0\\
r-r_0\\
k\\
1
\end{bmatrix}
\label{eq:us-voxel-metric}
\end{equation}
where $(c_0,r_0)$ is the image origin and $k=0$ for a 2D frame. DICOM orientation and origin tags define the IJK-to-LPS mapping, and coordinates are converted to centimeters for rendering.

\paragraph{Interaction, Reconstruction, and Registration}
A user polygon generates an optionally smoothed binary ROI mask; landmarks are stored as frame, image coordinate, and anatomical label and exported to CSV. A DICOM stack or cine buffer forms a pseudo-volume, from which marching cubes extracts a metric-scaled surface. With probe tracking, the ultrasound-to-CT alignment is

\begin{equation}
\TF{CT}{US}=\TF{CT}{PO}\,\TF{PO}{PR}\,\TF{PR}{US}
\end{equation}
where $\TF{PR}{US}$ is the probe calibration and $\TF{PO}{PR}$ is the live optical-tracker pose. Outputs comprise landmark CSV files, optional per-frame ROI masks, and 3D meshes (Fig.~\ref{fig:US_mode_figs}).
Figure~\ref{fig:US_gui} shows the ultrasound GUI with ROI and landmark annotations, and Fig.~\ref{fig:US_roi} shows the corresponding ROI mask overlay.
MIRAGE implements bounded latest-frame queueing, ROI processing, pseudo-volume assembly, and metric scaling for ultrasound data, live image streaming follows the OpenIGTLink protocol \cite{Tokuda2009}, physical image coordinates and voxel spacing are derived from DICOM metadata \cite{DICOM2026}, and 3-D surface extraction uses the established marching-cubes methodology \cite{Lorensen1987}.

Appendix~\ref{app:ultrasound_streaming} provides the complete queueing, DICOM-coordinate, ROI, pseudo-volume, scaling, and parameter formulations.
\subsection{3D Stereo Reconstruction Point Cloud}
\paragraph{Overview}
A synchronized, rectified stereo pair is acquired, disparity is then estimated with a learned model (e.g., RAFT-Stereo \cite{lipson2021} or PSMNet \cite{chang2018}). A colored 3D point cloud is reconstructed and cleaned, then cloud is visualized alongside preoperative CT meshes in a visualization toolkit (VTK) scene. Live 6-DoF poses from the optical tracker \{PO\} update the phantom model STL pose so meshes remain registered in \{CT\}. The loop repeats upon user request. Both learned stereo models (RAFT-Stereo and PSMNet) are implemented within this unified framework, allowing either to be selected as needed. While RAFT-Stereo generally yields superior results with shorter runtimes, PSMNet can be more accurate for certain surface reconstructions of specific objects.
Further details are provided in Section \ref{experimental_validation}.

\subsubsection{Calibration and Rectification}
Intrinsic and stereo parameters, \(\{\mathbf{K}_L,\mathbf{D}_L,\mathbf{K}_R,\mathbf{D}_R,
\RF{L_r}{L},\RF{R_r}{R},\mathbf{P}_L,\mathbf{P}_R,\mathbf{Q}_{\rm stereo}, \RF{R}{L},\TV{R}{L}\}\), are used, where \(\RF{L_r}{L},\RF{R_r}{R}\) are the rectification rotations into frames $\{L_r\}$ and $\{R_r\}$, respectively, with $L\equiv LE$ and $R\equiv RE$, and \((\RF{R}{L},\TV{R}{L})\) are the original right-from-left extrinsics prior to rectification. The matrix \(\mathbf{Q}_{\rm stereo}\in\mathbb{R}^{4\times 4}\) is the stereo \emph{reprojection} (disparity-to-3D) matrix for the left-rectified camera, mapping \([u,v,d_{\rm cv},1]^\mathrm{T}\) to homogeneous \([X,Y,Z,1]^\mathrm{T}\) up to the nonzero scale $\kappa$. It encodes \(f_{\rm stereo},c_x,c_y\) and the baseline (so \(Z\) has the same length units as the baseline).
Each pixel \(\mathbf{u}\) is mapped to the rectified domain by the rectification operator
\begin{equation}
\mathbf{u}^{\mathrm{rect}} = \mathrm{Rect}\!\left(\mathbf{u};\,\mathbf{K},\mathbf{D},\mathbf{R}_{\rm rect},\mathbf{P}\right)
\end{equation}
Let \(\mathbf{I}_L\) denote the rectified left image. The matrices \(\mathbf{P}_L,\mathbf{P}_R\in\mathbb{R}^{3\times 4}\) are the rectified projection matrices; their \(3\times 3\) upper-left blocks
\(\mathbf{K}_{L_r}=\mathbf{P}_L[0:3,0:3]\) and \(\mathbf{K}_{R_r}=\mathbf{P}_R[0:3,0:3]\)
are the effective intrinsics after rectification. In the generic rectification operator, $\mathbf{R}_{\rm rect}$ is the corresponding eye-specific rectification rotation.
\subsubsection{Disparity Estimation}
Given \(\mathbf{I}_L\) and the rectified right image \(\mathbf{I}_R\), a dense disparity \(d_{\rm disp}(\mathbf{u})\) is computed by RAFT-Stereo or PSMNet. RAFT-Stereo outputs right-to-left disparity. OpenCV left-to-right convention is recovered by,
\begin{equation}
d_{\text{cv}}(\mathbf{u}) =
\begin{cases}
-d_{\text{RAFT}}(\mathbf{u}), & \text{RAFT}\\
\;\;\;d_{\text{PSM}}(\mathbf{u}), & \text{PSMNet}
\end{cases}
\end{equation}
\subsubsection{Back-Projection to 3D}
Disparity is reprojected into the left-rectified eye using stereo reprojection matrix \(\mathbf{Q}_{\rm stereo}\):
\begin{equation}
\kappa
\begin{bmatrix} X \\ Y \\ Z \\ 1 \end{bmatrix}
=
\mathbf{Q}_{\rm stereo}
\begin{bmatrix} u \\ v \\ d_{\text{cv}} \\ 1 \end{bmatrix},
\quad \kappa \neq 0
\label{eq:Q-reprojection}
\end{equation}
Under the rectified pinhole stereo model with equal horizontal and vertical focal lengths and aligned principal points, with baseline \(b_{\rm stereo}\) and focal length \(f_{\rm stereo}\),
\begin{align}
Z &= \frac{f_{\rm stereo}\,b_{\rm stereo}}{d_{\text{cv}}}, \\
X &= \frac{(u - c_x)}{f_{\rm stereo}}\,Z, \qquad
Y = \frac{(v - c_y)}{f_{\rm stereo}}\,Z
\end{align}
Point color is sampled from \(\mathbf{I}_L\). For visualization in Open3D/VTK, the frame-flipped coordinate
\(\tilde{\mathbf{p}}=\big[X,\,-Y,\,-Z\big]^\mathrm{T}\)
is used; \(\tilde{\mathbf{p}}\) denotes this visualization-convention variant.

\subsubsection{Validity Masking and Cleanup}
Disparity is denoised by a bilateral filter and capped at the \(99.8\)th percentile. The valid set combines a minimum disparity threshold, a photometric threshold on \(\mathbf{I}_L\), and a morphological opening:
\begin{equation}
\mathcal{V}_{\rm valid} = \Big\{ \mathbf{u} \;\big|\; d_{\text{cv}}(\mathbf{u}) > d_{\rm disp,min},\;
d_{\text{cv}}(\mathbf{u}) < \tau_{\rm disp,99.8},\;
\|\mathbf{I}_L(\mathbf{u})\|_{\infty} > \iota \Big\}
\end{equation}
Here $d_{\rm disp,min}$ is the minimum disparity, $\tau_{\rm disp,99.8}$ the 99.8th-percentile disparity cutoff, and $\iota$ the image-intensity threshold. Let $\mathcal{Z}_{\rm depth}=\{z_i\}_{i=1}^{N_{\rm depth}}$ be the set of reconstructed depths. Define;
\begin{align}
&q_1 := \mathrm{quantile}(\mathcal{Z}_{\rm depth},0.25),\\
&q_3 := \mathrm{quantile}(\mathcal{Z}_{\rm depth},0.75),\\
&\Delta_{\rm IQR} := q_3 - q_1
\end{align}
Points are kept if,
\begin{equation}
q_1 - k_{\rm IQR}\,\Delta_{\rm IQR} \;\le\; z_i \;\le\; q_3 + k_{\rm IQR}\,\Delta_{\rm IQR}
\end{equation}
with \(k_{\rm IQR}>0\) (e.g., \(k_{\rm IQR}{=}3.0\)).
 The cloud is further cleaned by statistical outlier removal (mean distance, standard-deviation gate) and radius outlier removal.

\subsubsection{Projection Back to the Rectified Planes}
The relative pose \emph{between rectified cameras} is;
\begin{align}
&\RF{R_r}{L_r} = \RF{R_r}{R}\,\RF{R}{L}\,\RF{L_r}{L}^{\mathrm{T}}, \\
&\TV{R_r}{L_r} = \RF{R_r}{R}\,\TV{R}{L}
\end{align}
For a point \(\mathbf{X}_{L_r}\) in the left-rectified frame,
\begin{align}
&\mathbf{X}_{R_r} = \RF{R_r}{L_r}\,\mathbf{X}_{L_r} + \TV{R_r}{L_r},\\
&\mathbf{u}_L = \pi\!\left(\mathbf{K}_{L_r}\,\mathbf{X}_{L_r}\right), \\
&\mathbf{u}_R = \pi\!\left(\mathbf{K}_{R_r}\,\mathbf{X}_{R_r}\right)
\end{align}
with \(\pi([x,y,z]^\mathrm{T})=[x/z,\;y/z]^\mathrm{T}\).

\subsubsection{Live Registration to CT Space}
The camera frame \{CM\} denotes the physical camera on the shaft. The left-eye endoscope frame \{LE\} is located at the tip. A calibrated rigid transform
$\TF{LE}{CM}=\begin{bmatrix}\RF{LE}{CM}&\TV{LE}{CM}\\\mathbf{0}_{1\times3}&1\end{bmatrix}$
aligns \{CM\} to \{LE\}. Let $\TF{PO}{BF}$ be the live pose of the butterfly tool \{BF\} in the optical tracker \{PO\} frame, \(\TF{LE}{BF}\) the fixed calibration from \{BF\} to \{LE\}, and \(\TF{PO}{CT}\) the preoperative extrinsic. A scene bridge \(\TF{CM}{O3D}\) places the renderer in \{CM\} frame. The runtime composition is as follow:

\begin{align}
\TF{CT}{CM}
&=\TF{CT}{PO}\,\TF{PO}{BF}\,\TF{BF}{LE}\,\TF{LE}{CM},\\[3pt]
\TF{CT}{O3D}
&=\TF{CT}{CM}\,\TF{CM}{O3D}
\end{align}

Here $\TF{BF}{LE}=(\TF{LE}{BF})^{-1}$ and $\TF{CT}{PO}=(\TF{PO}{CT})^{-1}$. The tracked pose and fixed calibration factors therefore share the same destination-from-source composition rule. The bridge includes the rectified-eye and renderer conventions: $\TF{CM}{O3D}=\TF{CM}{LE}\TF{LE}{L_r}\TF{L_r}{O3D}$, where the rotation of $\TF{L_r}{O3D}$ is $\operatorname{diag}(1,-1,-1)$ and its translation is zero.

\begin{figure}
\centering
\renewcommand{\thesubfigure}{\alph{subfigure}}
    \begin{subfigure}[b]{\textwidth}
        \centering
        \includegraphics[width=0.95\linewidth,height=0.31\textheight,keepaspectratio]{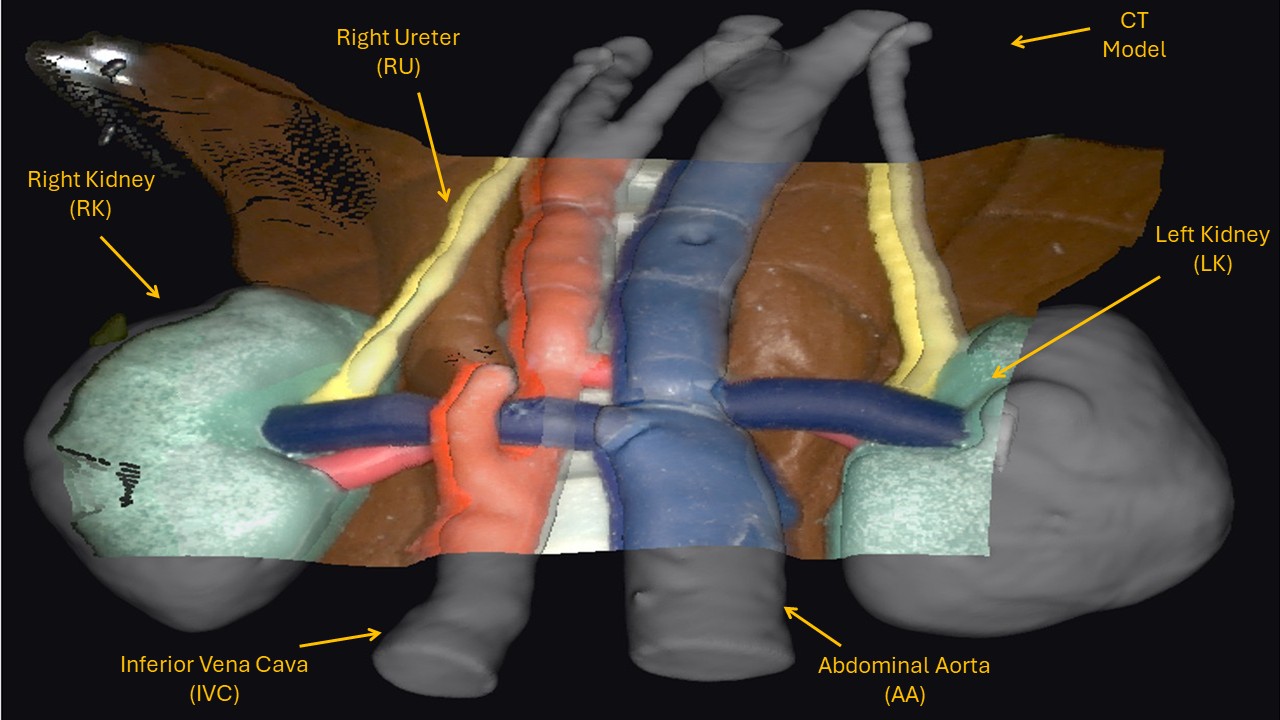}
        \caption{3D stereo reconstruction point-cloud overlay and CT model visualization}
        \label{fig:Stere_recon}
    \end{subfigure}
    \medskip
    \begin{subfigure}[b]{\textwidth}
        \centering
        \includegraphics[width=0.95\linewidth,height=0.31\textheight,keepaspectratio]{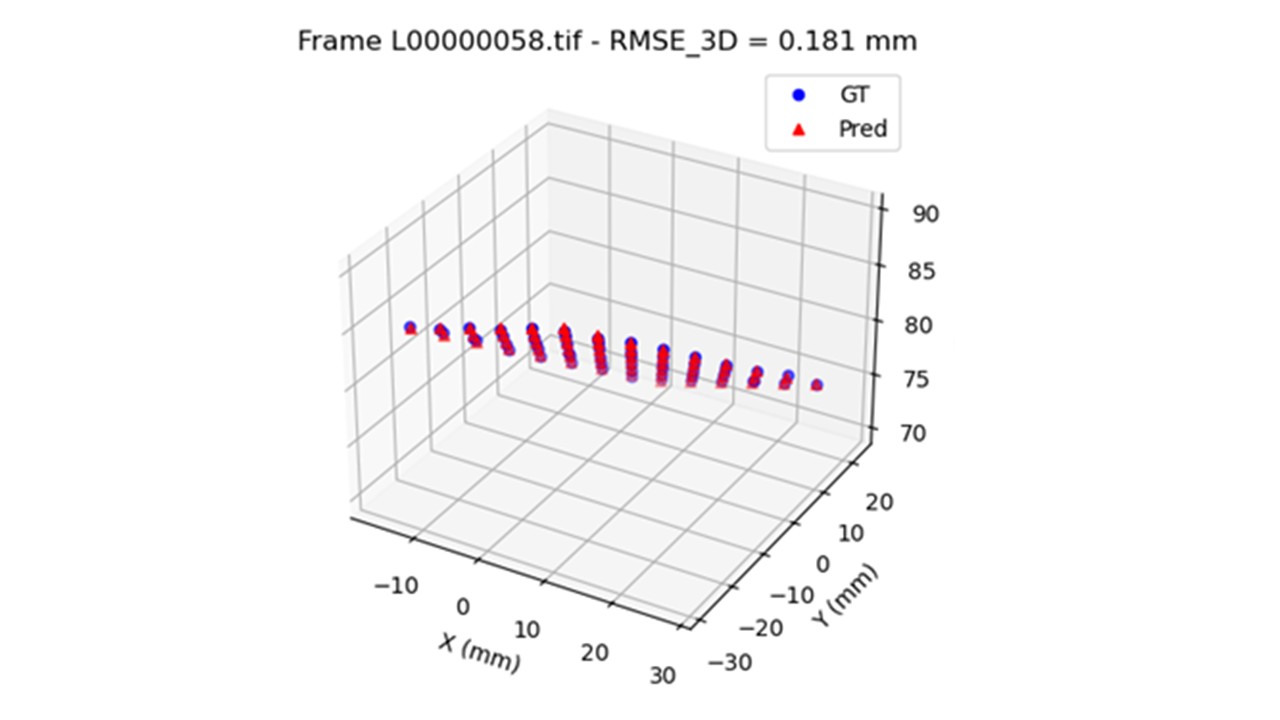}
        \caption{Stereo calibration 3D error between ground-truth (GT) and predicted (Pred) checkerboard corner locations}
        \label{fig:3D_error}
    \end{subfigure}

    \caption{3D stereo reconstruction CT overlay visualization and stereo calibration 3D error.}
    \label{fig:Stereo_Reconstruction_res}
\end{figure}
\subsubsection{VTK Scene and User Interaction}
A colored point cloud is rendered with translucent CT meshes as shown in Fig.~\ref{fig:Stere_recon}. Hotkeys toggle visibility, reset the camera, switch the driving tool, capture a new stereo pair, or quit. The capture UI is terminated before opening the VTK window to ensure exclusive control of tracker streaming.
\subsubsection{Control Flow}
Each loop launches the capture UI and waits until a \emph{new} rectified stereo pair stabilizes, computes disparity and reconstructs/cleans the point cloud, opens the VTK window and streams live \{PO\} poses to keep meshes aligned in \{CT\} space, and then returns to the capture stage when another acquisition is requested.
\subsubsection{Outputs and evaluations}
As detailed in Section~\ref{sec:calib-opt}, the calibration procedure is here extended to evaluate the endoscope performance across different \emph{working distances}, using the da Vinci Xi and SP robots. The calibration error for the 100mm working distance is illustrated in Fig.~\ref{fig:3D_error}, while the complete calibration results across all working distances are summarized in Table~\ref{table:stereo_calib_res}.
\begin{table}[H]
\centering
\caption{Pixel size and physical error at different working distances.}
\label{table:stereo_calib_res}
\begin{tabular}{ccc}
\toprule
\textbf{Working distance} & \textbf{Pixel error} & \textbf{Physical error} \\
\midrule
100\,mm & 2\,px & 0.1\,mm \\
300\,mm & 2.5\,px & 0.3\,mm \\
500\,mm & 2.5\,px & 0.6\,mm \\
\bottomrule
\end{tabular}
\end{table}
\section{{Framework Architecture:Implementation}}
The augmented-reality video overlay described in Section~\ref{sec:AR_Overlay} and ultrasound streaming explained in Section~\ref{ultrasound} are implemented in C++ because they require continuous video acquisition, rendering, and hardware-level interaction. Complementary modules including rigid registration, external hardware tracking (e.g., EM and optical), 3D stereo reconstruction point-cloud overlay, HUD navigation, and the main GUI are implemented in Python. The integration challenge addressed in this section is not simply exposing C++ functions to Python, but maintaining consistent state, timestamped frame access, thread-safe parameter updates, and reproducible module switching across heterogeneous subsystems.

The proposed binding layer uses pybind11 as the low-level interface mechanism, but the methodological contribution lies in the surrounding architecture: explicit buffer ownership, timestamped frame exchange, mutex-protected shared state, asynchronous worker management, and a unified state machine that coordinates C++ and Python modules. This design allows frame buffers generated by the C++ backend to be exposed to Python without additional image copies when possible, while long-running C++ operations remain isolated from the GUI event loop. The following sections detail the binding mechanism and overall system integration.

\begin{figure}[ht]
  \centering
  \includegraphics[width=0.96\linewidth,height=0.64\textheight,keepaspectratio]{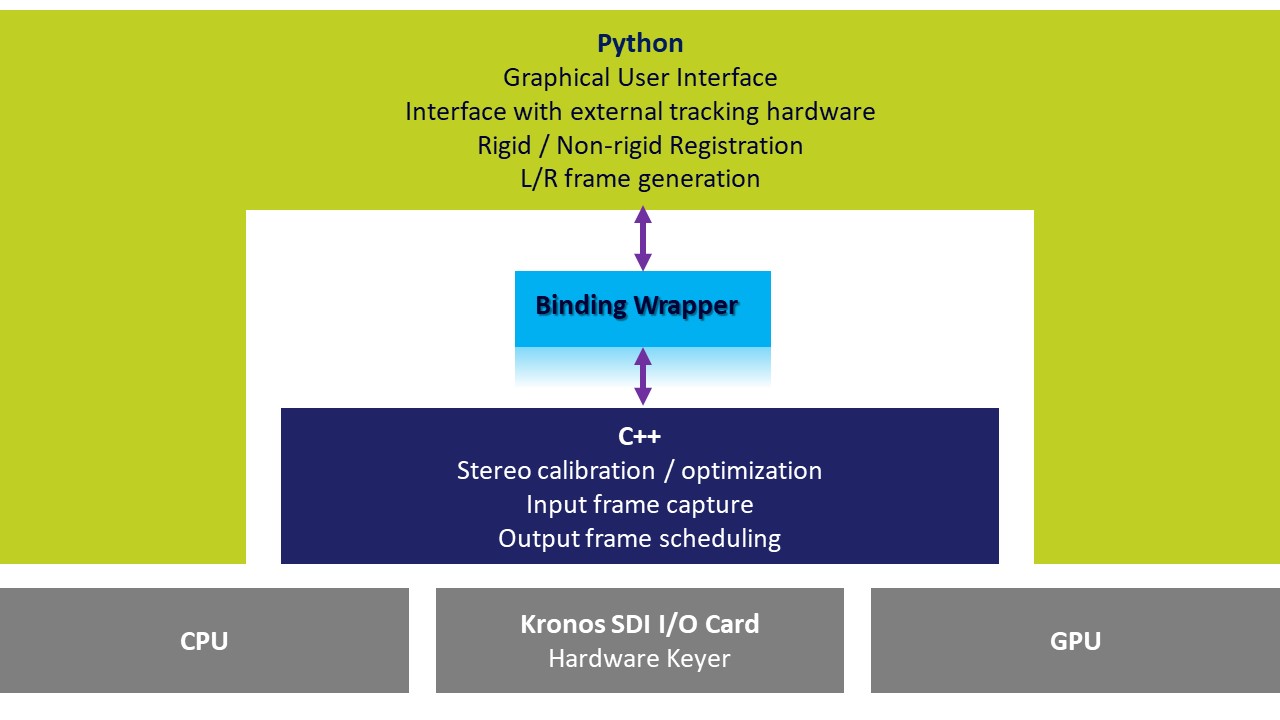}
  \caption{Software-hardware strategy.}
  \label{fig:Software_architecture}
\end{figure}

\subsubsection{Cross-Language State and Buffer Binding}

The cross-language binding layer, referred to as \emph{OverlayEngine}, is designed to expose selected C++ runtime state to Python while preserving frame timing, memory validity, and thread safety. The layer provides: (1) Python-callable control functions for starting, stopping, and configuring C++ modules; (2) timestamped access to left/right eye video buffers; (3) synchronized tracker and rendering state; (4) mutex-protected parameter updates; (5) asynchronous execution of blocking hardware loops; and (6) a consistent API for higher-level Python modules. Pybind11 is used as the binding mechanism; the system-level contribution is the design of a deterministic data and control pathway between the real-time backend and the Python orchestration layer. The software-hardware strategy is illustrated in Fig.~\ref{fig:Software_architecture}. This approach leverages the performance advantages of native C++ alongside the flexibility of Python.
\paragraph{System Architecture}

The pipeline is implemented in C++ and exposes its main functionality to Python via a Pybind11-based interface. The architecture comprises:

\begin{enumerate}
    \item \textbf{Low-level backend:} High-performance C++ modules for hardware communication and rendering, including:
    \begin{itemize}
        \item \textbf{Bluefish} for real-time stereo video acquisition (left/right RGBA frames).
        \item \textbf{Tracker Interface} for optical/EM tracker data acquisition.
        \item \textbf{OpenGL} for rendering and interactive GUI.
    \end{itemize}
    \item \textbf{Core} "video overlay (AR)","ultrasound" and "calibration" pipelines: Central C++ control logic for streaming, registration, calibration, and visualization.
    \item \textbf{Bindings:} A \emph{binding wrapper} is developed to expose C++ classes, functions, and runtime state to Python, enabling direct access and control from high-level scripts and interactive GUI applications.
     \item \textbf{High-level frontend:} The framework's module launcher and all other components including stereo reconstruction, tracker registration, and the HUD are implemented in Python.
\end{enumerate}

\paragraph{Python API}

The exposed Python interface enables users to flexibly control the system. For instance, overlays and calibration pipelines can be launched as background processes, automatically initiating all necessary C++ backend components responsible for video streaming, hardware tracking, and graphical user interface management. Users can retrieve the most recent left and right video frames in real time as NumPy arrays, which allows for additional processing or visualization in Python. The interface also allows starting and stopping of pipeline threads, switching between modes, and dynamic configuration of runtime parameters such as camera tracking, rendering options, and GUI settings. Furthermore, registration sessions as well as calibration and augmented reality overlay workflows can be executed directly from Python main GUI, providing seamless integration and control over the entire multimodal pipeline.

    The main GUI in Python, provides methods for all essential operations, including launching or terminating the C++ backend threads and main event loop, retrieving the current left/right camera images(RGBA), updating paths to YAML configuration files and data, dynamically controlling rendering, modes, visualization options, and hardware settings, as well as standalone function bindings to allow direct execution of registration or analysis pipelines in blocking or background thread modes.

\paragraph{Pybind11-Based Communication Workflow} \label{lable:binded_interface}

The developed platform, enables the C++ implementation to exposes C++ classes and global application state including frame buffers and runtime flags to Python, ensuring thread safety through the use of mutex locking mechanisms. To avoid ambiguity about buffer lifetime, each exported frame is associated with a validity interval. Python receives a view of the current C++ buffer together with a timestamp and frame index.  A zero-copy view shares the underlying buffer storage. Consistent frame access requires the owner to retain that storage and prevent conflicting writes throughout the consumer read; retaining a frame beyond that interval requires ownership protection or a copy~\cite{EditorialPybind11}. Long-running C++ loops release the Python interpreter lock during execution and communicate status changes through atomic flags or mutex-protected state variables. GUI operations therefore remain responsive while the backend continues frame acquisition, rendering, and tracking. Data is efficiently marshaled between C++ and Python environments, with image buffers converted into NumPy arrays that preserve the correct dimensions and memory layout for seamless access. From Python, background threads can be launched and managed to handle rendering, video capture, and tracking tasks, allowing the system to maintain real-time performance. This design allows high-level Python scripts to control every aspect of the overlay pipeline, including camera selection, 3D visualization parameters, and hardware tracking or calibration routines.

\begin{figure}[htbp]
\centering
\resizebox{0.98\textwidth}{!}{
\begin{tikzpicture}[
node distance=0.70cm and 1.55cm,
block/.style={draw, rounded corners, align=center, minimum width=3.05cm, minimum height=0.58cm, inner sep=2pt, font=\small},
wide/.style={draw, rounded corners, align=center, minimum width=4.75cm, minimum height=0.58cm, inner sep=2pt, font=\small},
arrow/.style={-{Latex[length=2mm]}, thick}
]
\node[wide, fill=blue!6] (cpp) {C++ backend\\ Video I/O, rendering, tracker drivers};
\node[wide, fill=green!6, right=3.0cm of cpp] (py) {Python frontend\\ GUI, registration, HUD, reconstruction};

\node[block, fill=orange!10, below=0.50cm of cpp] (buf) {Timestamped\\ frame buffers};
\node[block, fill=orange!10, below=0.50cm of py] (pose) {Timestamped\\ tracker poses};

\node[wide, fill=yellow!12, below=0.65cm of $(buf)!0.5!(pose)$] (sync) {Synchronization/interpolation module\\ interpolated pose at video timestamp};

\node[block, fill=gray!10, below left=0.55cm and -0.2cm of sync] (ar) {AR 3D overlay};
\node[block, fill=gray!10, below=0.55cm of sync] (hud) {3D model + HUD};
\node[block, fill=gray!10, below right=0.55cm and -0.2cm of sync] (us) {Ultrasound};
\node[block, fill=gray!10, right=0.65cm of us] (rec) {Stereo reconstruction};

\draw[arrow] (cpp) -- node[above, font=\scriptsize]{control/status} (py);
\draw[arrow] (py) -- node[below, font=\scriptsize]{commands/parameters} (cpp);

\draw[arrow] (cpp) -- (buf);
\draw[arrow] (cpp) -- (pose);
\draw[arrow] (py) -- (pose);

\draw[arrow] (buf) -- (sync);
\draw[arrow] (pose) -- (sync);

\draw[arrow] (sync) -- (ar);
\draw[arrow] (sync) -- (hud);
\draw[arrow] (sync) -- (us);
\draw[arrow] (sync) -- (rec);
\end{tikzpicture}
}
\caption{Cross-language state and buffer binding architecture. The C++ backend handles video I/O, rendering, and hardware communication, while the Python frontend coordinates registration, HUD, stereo reconstruction, ultrasound, and GUI control. Timestamped frame buffers and tracker poses are exchanged through a thread-safe binding layer and synchronized before display.}
\label{fig:GUIs}
\end{figure}
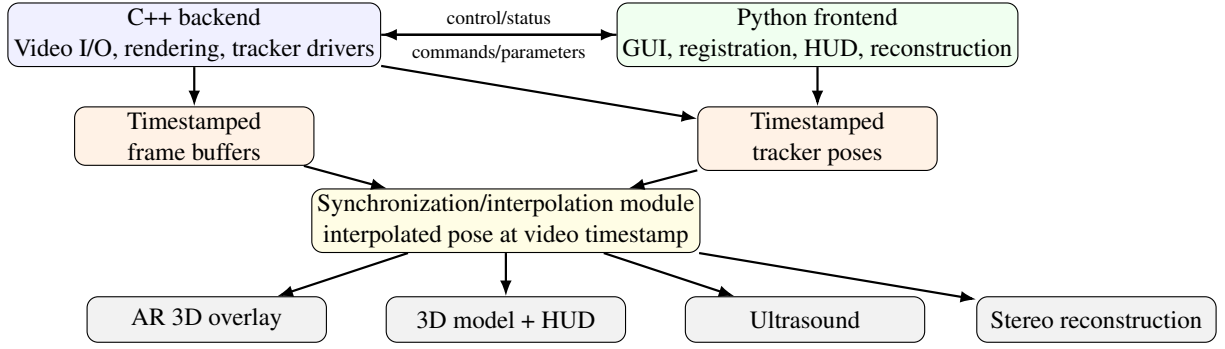

Figure~\ref{fig:GUIs} illustrates the cross-language state and buffer binding architecture. The C++ backend manages video acquisition, rendering, and hardware communication, while the Python frontend coordinates registration, stereo reconstruction, HUD navigation, ultrasound interaction, and GUI control. Timestamped frame buffers and tracker poses are exchanged through a thread-safe binding layer. The synchronization module associates each rendered frame with the closest or interpolated tracker pose before display.

\subsubsection{Mathematics of Bindings}
The interaction between the C++ backend ($\mathcal{C}$) and the Python frontend ($\mathcal{P}$) is formalized as a collection of mathematical mappings, state transitions, and synchronization operations. The integration leverages Pybind11 to enable real-time, zero-copy data exchange and multithreaded control with formal guarantees.

\paragraph{Bidirectional Interface Mapping}

Let $\mathcal{F}_{\mathcal C}^{\rm exp}$ be the exported C++ functions and $\mathcal{F}_{\mathcal P}^{\rm api}$ their Python-facing interfaces. The binding map $\phi:\mathcal F_{\mathcal C}^{\rm exp}\to\mathcal F_{\mathcal P}^{\rm api}$ associates each exported function $f_j$ with its Python interface $f_j^{\prime}$. The binding relation identifies the implementation invoked by each exposed call. Optional callbacks form a separately specified interface, $f_{\rm callback}:\mathcal{F}_{\mathcal P}^{\rm cb}\to\mathcal{F}_{\mathcal C}^{\rm handler}$; they need not be an inverse of the forward binding map.

\paragraph{Function and Data Flow}
\noindent Python bindings are thin wrappers such as: calling $f'_j$ is semantically identical to calling $f_j$ in C++, with identical domains. This ensures parity of arguments, outputs, and side effects.
Each Python-invoked method $f'_j$ directly evaluates the underlying C++ implementation:
\begin{equation}
    y = f'_j(x) = f_j(x), \qquad x \in \mathcal{X}_j
\end{equation}
Here, $\mathcal{X}_j$ denotes the set of all valid input arguments (domain) for the $j$-th C++ function $f_j$ and its Python binding $f'_j$.\\

For data buffers (e.g., image frames) $\mathbf{B} \in \mathbb{R}^{H_{\rm img}\times W_{\rm img}\times N_{\rm ch}}$, where $N_{\rm ch}$ is the number of channels, the buffer-protocol map is:
\begin{equation}
    \mathbf B^{(\mathcal P)}(t)=\psi\!\left(\mathbf B^{(\mathcal C)}(t)\right)
\end{equation}
with
\begin{equation}
    \forall\, (r_{\rm px},c_{\rm px},j_{\rm ch}):\; \mathbf{B}^{(\mathcal{P})}(t)[r_{\rm px},c_{\rm px},j_{\rm ch}] = \mathbf{B}^{(\mathcal{C})}(t)[r_{\rm px},c_{\rm px},j_{\rm ch}]
\end{equation}
and the memory invariant:
\begin{equation}
    \operatorname{addr}\!\left(\mathbf{B}^{(\mathcal{P})}(t)\right) = \operatorname{addr}\!\left(\mathbf{B}^{(\mathcal{C})}(t)\right)
\end{equation}
\paragraph{Thread Safety and Synchronization}
\noindent Shared state is guarded by mutexes or atomics to prevent race conditions. Mutual exclusion ensures updates are serialized or performed atomically.
For each shared state variable $\xi \in \Xi$, synchronization is enforced via mutex $\mu_\xi$ or atomic operations. The update is:
\begin{align}
    \text{lock}(\mu_\xi) & \implies \xi \gets \xi' ,\\
    \text{unlock}(\mu_\xi)
\end{align}
or, if atomic:
\begin{equation}
    \xi \gets_{\text{atomic}} \xi'
\end{equation}

The expected mutual exclusion property, with $q$ indexing threads, is:
\begin{equation}
    \forall t:\quad\#\{q:\ q\text{ owns mutex }\mu_\xi\text{ at time }t\}\leq1
\end{equation}
This ensures reliable and safe operation of the system in a multithreaded (real-time) environment.
\paragraph{Frame Delivery, Consistency, and Latency}

\noindent Zero-copy guarantees byte-wise equality between C++ and Python frame buffers, while latency $\Delta t_{\rm binding}^{(n)}$ is bounded by a small $\epsilon_{\rm binding}$. This enables real-time visualization and analysis.
Define $\mathbf{B}_n^{(\mathcal{C})}$ as the $n$-th frame generated by C++, with $\mathbf{B}_n^{(\mathcal{P})}$ the corresponding Python-visible buffer. A zero-copy, real-time invariant holds:
\begin{equation}
    \mathbf{B}_n^{(\mathcal{P})} = \mathbf{B}_n^{(\mathcal{C})}, \quad \forall n
\end{equation}
with latency:
\begin{equation}
    \Delta t_{\rm binding}^{(n)} = t_{\text{Python}}^{(n)} - t_{\text{C++}}^{(n)} \leq \epsilon_{\rm binding}
\end{equation}
 (e.g., \(\epsilon_{\rm binding}\) is in milliseconds).

\paragraph{Event-Driven and Blocking/Non-Blocking Operations}
\noindent Synchronous commands block in Python, while asynchronous ones spawn managed threads with explicit state-transition dynamics. Only one worker per async command instance is allowed.
Let $C_{\mathrm{async}}$ and $C_{\mathrm{sync}}$ denote the set of asynchronous (threaded) and synchronous (blocking) commands:
\begin{align}
    C_{\mathrm{sync}} &\subset \mathcal{F}_{\mathcal P}^{\rm api}, \\
    C_{\mathrm{async}} &\subset \mathcal{F}_{\mathcal P}^{\rm api}
\end{align}

Each asynchronous call $c_a \in C_{\mathrm{async}}$ spawns a new thread $\mathsf{Thread}_{c_a}$:
\begin{equation}
    \mathsf{Thread}_{c_a}:~ k_{\rm e}\mapsto s_{c_a}(k_{\rm e}+1) = f_{\rm worker,c_a}\!\left(s_{c_a}(k_{\rm e}),\, u_{c_a}(k_{\rm e})\right)
\end{equation}
where $f_{\rm worker,c_a}$ denotes the per-command state-transition map. If $N_{\rm worker}(c_a,k_{\rm e})$ is the number of active workers for command instance $c_a$, thread lifetime is constrained by
\begin{equation}
    N_{\rm worker}(c_a,k_{\rm e})\in\{0,1\}
\end{equation}

\paragraph{Global Application State Evolution}
\noindent The application is modeled as a discrete-time dynamical system whose next state depends on the current state and input from both Python and C++. This abstraction unifies heterogeneous subsystems under a single update rule.
We aggregate all subsystems into the state vector $S(k_{\rm e}) = \left[s_1(k_{\rm e}),\, \ldots,\, s_{N_{\rm sub}}(k_{\rm e}),\, \Theta(k_{\rm e})\right]$, where $N_{\rm sub}$ is the number of subsystems and
$\Theta(k_{\rm e})$ denotes a parameter vector comprising persistent configuration, calibration parameters, and runtime tunables. When a module requires only calibration constants, we write the calibration block $\Theta_{\mathrm{calib}}$. Then the evolution is:
\begin{equation}
    S(k_{\rm e}+1) = f_{\rm state}\!\left(S(k_{\rm e}),\, U^{(\mathcal{P})}(k_{\rm e}),\, U^{(\mathcal{C})}(k_{\rm e})\right)
\end{equation}
where $U^{(\mathcal{P})}(k_{\rm e})$ are Python actions (e.g., GUI input) and $U^{(\mathcal{C})}(k_{\rm e})$ are internal/automated C++ events.

\paragraph{Closed-Loop Feedback}
\noindent Python computes control actions from current C++ outputs, and C++ emits new outputs after state updates, forming a feedback loop. This supports interactive GUIs and adaptive control.
The Python frontend acts both as a controller and observer:
\begin{equation}
    \begin{cases}
        U^{(\mathcal{P})}(k_{\rm e}) = f_{\rm control}\!\left(Y^{(\mathcal{C})}(k_{\rm e})\right) \\
        Y^{(\mathcal{C})}(k_{\rm e}+1) = f_{\rm output}\!\left(S(k_{\rm e}+1)\right)
    \end{cases}
\end{equation}
where $f_{\rm control}$ is the Python control policy (e.g., GUI response), and $f_{\rm output}$ is the C++ output mapping (frames, status, errors).

\paragraph{Parameter Update Flow}
\noindent Parameter changes originate in Python and are propagated to C++ with appropriate locking/atomicity. This preserves consistency between user intent and engine configuration.
When Python requests a parameter change:
\begin{equation}
    \chi\!\left(\Theta^{(\mathcal P)}(k_{\rm e})\right)=\Theta^{(\mathcal C)}(k_{\rm e})
\end{equation}
with thread-safe, blocking, or atomic guarantees.

\paragraph{API Contract (Interface Invariants)}
\noindent Pre/post-conditions specify valid inputs and expected outputs for each bound function. Buffer validity is guaranteed between updates to prevent use-after-free.
Specifically, the API contract $\mathcal{I}$ comprises the following membership requirements and pre/post-conditions for each exposed function:
\begin{align}
    \forall j:~ &f'_j\in\mathcal F_{\mathcal P}^{\rm api},\quad f_j\in\mathcal F_{\mathcal C}^{\rm exp} \\
    &\text{pre:}~ x \in \operatorname{dom}(f_j),\ \text{post:}~ y = f_j(x)
\end{align}
For the same frame storage, the buffer-lifetime invariant is:
\begin{equation}
    \text{storage retained and protected from conflicting writes}
\ \Longrightarrow\ \mathbf B^{(\mathcal P)}(t)\text{ remains valid}
\end{equation}

\paragraph{Summary}
\noindent The mappings $(\phi,\psi,\chi)$ unify function calls, buffers, and parameters across languages under real-time constraints.  Together they describe the data dependencies and state updates of the integrated application.
The binder thus provides a real-time, zero-copy, multithreaded mapping:
\begin{align}
    \phi &: \mathcal{F}_{\mathcal C}^{\rm exp} \to \mathcal{F}_{\mathcal P}^{\rm api} \\
    \psi\!\left(\mathbf B^{(\mathcal C)}\right)&=\mathbf B^{(\mathcal P)} \\
    \chi\!\left(\Theta^{(\mathcal P)}\right)&=\Theta^{(\mathcal C)}
\end{align}
 The overall overlay and analysis system combines discrete configuration transitions with continuously updated image and pose data. 

\subsubsection{Reproducibility and Implementation Details}

To support reproducibility beyond the study hardware, MIRAGE separates device-dependent components, including Bluefish video capture, tracker drivers, and da Vinci video configuration, from device-independent components, including transform chains, timestamp association, calibration schemas, stereo reconstruction, registration, HUD metrics, and GUI-state logic.
Implementation details are reported as follows: frame buffers are stored as timestamped stereo pairs with indices; tracker poses as timestamped $4 \times 4$ homogeneous matrices; calibration files include intrinsics, distortion, stereo extrinsics, rectification, projection, and transform-chain definitions; module switching uses a state machine; blocking hardware operations run in worker threads; and logs record frame rate, dropped frames, tracker rate, stale poses, synchronization error, and module status.

Although some proprietary hardware interfaces cannot be fully released, the mathematical formulation, pseudocode, configuration schema, and device-independent evaluation protocol allow reproduction of the same architecture using alternative capture devices, trackers, or robotic platforms.

The device-independent architecture defines transform composition, stream association, calibration schemas, and evaluation endpoints. Reproduction on another platform additionally depends on compatible device interfaces, configuration data, and a reproducible acquisition or replay pathway.

\section{Multimodal framework: Integration}

This framework unifies multiple real-time modalities including \textbf{HUD navigation-visualization} ($\mathcal{H}$), \textbf{AR Overlay} ($\mathcal{O}$), \textbf{3D Stereo-Reconstruction} ($\mathcal{R}$), and \textbf{Ultrasound} ($\mathcal{U}$) along with \textbf{rigid registration} ($\mathcal{RR}$) and \textbf{stereo calibration} ($\mathcal{S}$) features. Modalities communicate either with the C++ backends$\mathcal{O}$ and $\mathcal{U}$ and external devices via well-defined Python interfaces (Section~\ref{lable:binded_interface}), or directly through the Python GUI $\mathcal{R}$ and $\mathcal{H}$, etc. Algorithm~\ref{alg:MIRAGE_full} outlines the pipeline of the unified framework.
\subsection{Modes and State Switching}
\noindent Modes are discrete GUI configurations; the selector $f_{\rm select}$ maps the current mode and user action to the next mode, turning toolbar/shortcut inputs into deterministic transitions.

Let $\mathcal{M} = \{\mathcal{H},\,\mathcal{O},\,\mathcal{R},\,\mathcal{U}, \ldots \}$ denote the available GUI modes (HUD-TilePro, video overlay, stereo reconstruction, ultrasound, \ldots). Let $m(k_{\rm e}) \in \mathcal{M}$ denote the selected mode at update $k_{\rm e}$.

Mode switching is modeled as a discrete-time Markov process:
\begin{equation}
    m(k_{\rm e}+1) = f_{\rm select}\!\left( m(k_{\rm e}),\, u_{\mathrm{user}}(k_{\rm e}) \right)
\end{equation}
where $f_{\rm select}$ is the mode selector function, and $u_{\mathrm{user}}(k_{\rm e})$ is the user action.

\subsection{Per-Mode Subsystem States}
\noindent Each mode $m$ maintains its own state $S_m(k_{\rm e})$, bundling the signals it needs (e.g., tracker poses, GUI panels, overlay buffers) so updates and rendering stay isolated yet composable.

Write $\mathbf B(k_{\rm e})=\mathbf B(t_{k_{\rm e}})$ for the buffer available at event time $t_{k_{\rm e}}$. Each mode $m$ has a subsystem state $S_m(k_{\rm e})$, for instance:
\begin{align}
    S_{\mathcal{H}}(k_{\rm e}) &= \left[ \mathcal{T}_{\mathrm{pose}}(k_{\rm e}),\, \mathcal{M}_{\mathrm{phantom}}(k_{\rm e}),\, \mathcal{Q}_{\mathrm{console}}(k_{\rm e}) \right] \\
    S_{\mathcal{O}}(k_{\rm e}) &= \left[ \mathbf{B}^{(\mathcal{P})}_{L}(k_{\rm e}),\, \mathbf{B}^{(\mathcal{P})}_{R}(k_{\rm e}),\, \eta_{\mathrm{overlay}}(k_{\rm e}),\, \mathcal{V}_{\mathrm{panel}}(k_{\rm e}) \right]
\end{align}

where,
\begin{itemize}
\item $\mathcal{T}_{\mathrm{pose}}(k_{\rm e})$: tracker pose data (HUD-navigation)
    \item $\mathcal{M}_{\mathrm{phantom}}(k_{\rm e})$: model transforms (visualization)
    \item $\mathbf{B}^{(\mathcal{P})}_{L/R}(k_{\rm e})$: left/right overlay video frames visible in Python (AR overlay)
    \item $\eta_{\mathrm{overlay}}(k_{\rm e})$: overlay parameters (opacity, group, etc.)
    \item $\mathcal{V}_{\mathrm{panel}}(k_{\rm e})$: vector of all GUI panel controls and interactive settings (such as layer visibility, opacity, colors, toggles, and sliders) for the overlay mode at update $k_{\rm e}$
    \item $\mathcal{Q}_{\mathrm{console}}(k_{\rm e})$: queue or buffer of recent console messages, log outputs, and user/system notifications displayed in the HUD-navigation console at update $k_{\rm e}$.
\end{itemize}

\subsection{Unified Application State and Controller}
\noindent Tracker-driven visualization and zero-copy overlay rendering are expressed as state updates; the calibration block $\Theta_{\mathrm{calib}}$ of $\Theta$ ensures all assets are placed consistently across modules. As representative examples, we detail the control workflows for the following two modes.

\paragraph{Navigation-Visualization}
The HUD-TilePro subsystem receives tracker pose data and transforms all VTK actors:
\begin{align}
    \mathcal{T}_{\mathrm{pose}}(k_{\rm e}+1) &= \mathcal{T}_{\mathrm{tracker}}(k_{\rm e}+1), \\
    \mathcal{M}_{\mathrm{phantom}}(k_{\rm e}+1) &= \mathcal{F}_{\mathrm{model}}\!\big( \mathcal{T}_{\mathrm{pose}}(k_{\rm e}+1), \Theta_{\mathrm{calib}} \big)
\end{align}
where $\Theta_{\mathrm{calib}}$ denotes the calibration block of $\Theta$, including $\boldsymbol{\theta}_{\rm calib}$ from Eqs.~\eqref{eq:optim1} and~\eqref{eq:optim2}.
\paragraph{AR Overlay}
The overlay subsystem receives C++-generated frame buffers via the developed binding wrapper:
\begin{align}
    \big(\mathbf{B}^{(\mathcal{P})}_{L}(k_{\rm e}+1),\, \mathbf{B}^{(\mathcal{P})}_{R}(k_{\rm e}+1)\big) &= \psi_{\mathrm{overlay}}\!\big( \mathbf{B}^{(\mathcal{C})}_{L}(k_{\rm e}+1),\, \mathbf{B}^{(\mathcal{C})}_{R}(k_{\rm e}+1) \big)
\end{align}
with zero-copy property:
\begin{align}
&\operatorname{addr}\!\left(\mathbf{B}^{(\mathcal{P})}_{L}(k_{\rm e})\right) = \operatorname{addr}\!\left(\mathbf{B}^{(\mathcal{C})}_{L}(k_{\rm e})\right),\\
&\operatorname{addr}\!\left(\mathbf{B}^{(\mathcal{P})}_{R}(k_{\rm e})\right) = \operatorname{addr}\!\left(\mathbf{B}^{(\mathcal{C})}_{R}(k_{\rm e})\right)
\end{align}
A similar control pattern applies to the other modalities as well.
\subsection{Cross-Modal Actions and Event Loop}
\noindent A single dispatcher $\mathcal{D}_{\mathrm{loop}}$ processes inputs $U(k_{\rm e})$ from GUI and devices, routes them to the active mode, and schedules updates/events so all modalities remain synchronized.

All modes share a unified event loop, $\mathcal{D}_{\mathrm{loop}}$. Event dispatch is:
\begin{align}
&\mathcal{E}(k_{\rm e}+1) = \mathcal{D}_{\mathrm{loop}}\!\big( S(k_{\rm e}),\, U(k_{\rm e}) \big),\\
U(k_{\rm e}) &= \left[ u_{\mathrm{panel}}(k_{\rm e}),\, u_{\mathrm{track}}(k_{\rm e}),\,\ldots \right]
\end{align}
By convention, $U(k_{\rm e}) = \big[ U^{(\mathcal{P})}(k_{\rm e}),\, U^{(\mathcal{C})}(k_{\rm e}) \big]$ to match notation in the Overlay Engine mathematics.

\subsection{Frame Synchronization and FPS Monitoring}
For each mode with live frames, define the frame rate averaged over the observation window:
\begin{equation}
    \overline{\mathrm{FPS}}_m[t_1,t_2] = \frac{N_{\mathrm{frames}, m}}{t_2 - t_1}
    \label{eq:fps}
\end{equation}
for $N_{\mathrm{frames}, m}$ frames received between physical times $t_1<t_2$ (in seconds).

\subsection{Summary of Unified Multimodal System}
The system acts as a hybrid automaton, switching between modes, with each subsystem evolving as a dynamical system, all orchestrated via a unified event loop and state machine. All real-time data flows are managed via thread-safe, zero-copy interfaces, and the full state is observed and controlled via Python GUI display \(\Gamma()\).
\begin{align}
    \text{Modes:} \quad m(k_{\rm e}) &\in \{\mathcal{H},\,\mathcal{R},\,\mathcal{O},\,\mathcal{U},\,\ldots\} \label{eq:modes} \\
    \text{State:} \quad S(k_{\rm e}+1) &= f_{\rm state}\!\left( S(k_{\rm e}),\, U^{(\mathcal{P})}(k_{\rm e}),\, U^{(\mathcal{C})}(k_{\rm e}) \right) \label{eq:states}\\
    \text{Display:} \quad Y^{(\mathcal{P})}(k_{\rm e}) &= \Gamma\!\left( S(k_{\rm e}) \right) \label{eq:display}
\end{align}
This formalism enables patentable, high-performance, extensible multimodal surgical navigation systems supporting future device, modality, and automation integrations. 
\begin{algorithm}[H]
\caption{Unified Multimodal Framework Pipeline}
\label{alg:MIRAGE_full}
\small
\DontPrintSemicolon
\setlength{\algomargin}{0.5em}
\SetInd{0.6em}{0.8em}

\KwIn{Intrinsic/Extrinsic parameters, rom/stl files;  stereo weights \& calibration parameters;}
\KwOut{Registration pose; stereo calibration; real-time HUD-TilePro; 3D overlay; 3D Stereo-reconstruction; Ultrasound, etc.}
\textbf{Boot \& Config:}
Init UI; start optical \& EM trackers;\;
create \emph{Engine} (ENG); start ENG;\;
Arm CPU taps (L/R); enable tracking; 
Create views: HUD, Calibration, Overlay, Stereo-Recon., Ultrasound etc.; 
stack in \emph{MainWindow};  start FPS/health monitors;\;
\textbf{Main Loop:}
Read \texttt{mode} $\in$ \{\emph{Registration}, \emph{Stereo-Calib}, \emph{HUD-Tilepro}, \emph{Overlay}, \emph{Stereo-Recon}, \emph{Ultrasound}, ... \}\;
\Switch{mode}{
  \Case{\emph{(AR) Overlay}}{
     Pull L/R from ENG; apply opacity/X-ray/group; render overlay\;
    \If{Run Registration}{
      stop ENG/tracking; set tool ROMs \& FCSV; run REG UI;\;
      save REG output; restart ENG/tracking\;
    }
    \If{Screenshot/Save}{write current frames with timestamp\;}
  }
  \Case{\emph{HUD}}{
    In background: poll tracker poses; compose chains
    ($\TF{CT}{PO}$ or $\TF{CT}{FG}$, and $\TF{LE}{BF}$);\;
    low-pass transforms (re-orthonormalize);\;
    update 3D actors (CT/phantom, tool, targets);\;
    compute HUD metrics (distance/angle)\;
  }
  \Case{\emph{Stereo Calibration }}{
    capture L/R $\to$ dataset; detect chessboard corners;\;
    generate plate points; project with $(K,D)$ \& transforms;\;
    compute per-eye RMSE (report avg);\;
    save: \emph{images, transform}\;
  }
  \Case{\emph{Stereo Reconstruction}}{
    Load model (RAFT/PSMNet) \& calib params; build rectification maps;\;
    For each pair: rectify $\to$ infer disparity $\to$ reproject to dense point cloud;\;
    compose $\TF{CT}{O3D}$; map PCD \& pre-op meshes to CT; visualize; log stage timings/FPS\;
  }
  \Case{\emph{Ultrasound}}{
    Show US panel; overlay guidance cues as configured\;
  }
  \Case{\emph{...}}{
  ...
  }
}
Handle global actions (mode switch, errors, quit); update status bar; FPS/state\;
\textbf{Shutdown:}
Stop trackers; stop ENG;\;
\normalsize
\end{algorithm}
For frame $n$, a coherent rendering input comprises the stereo pair $(I_{L,n},I_{R,n})$, its timestamp $t_n$, the associated tracker pose $\widehat{\mathbf{T}}_n$, and the calibration and registration configuration $\Theta_n$. The displayed output can be expressed as
\begin{equation}
Y_n=f_{\rm render}(I_{L,n},I_{R,n},\widehat{\mathbf{T}}_n,\Theta_n,m_n),
\end{equation}
where $m_n$ is the selected mode. Frame identity, source-sample association, and configuration consistency determine which measurements contribute to each display update.
\section{Quantitative Experimental Validation} \label{experimental_validation}
\begin{figure}[ht]
\centering
    \begin{subfigure}{\linewidth}
        \centering
        \FigTwelvePanel{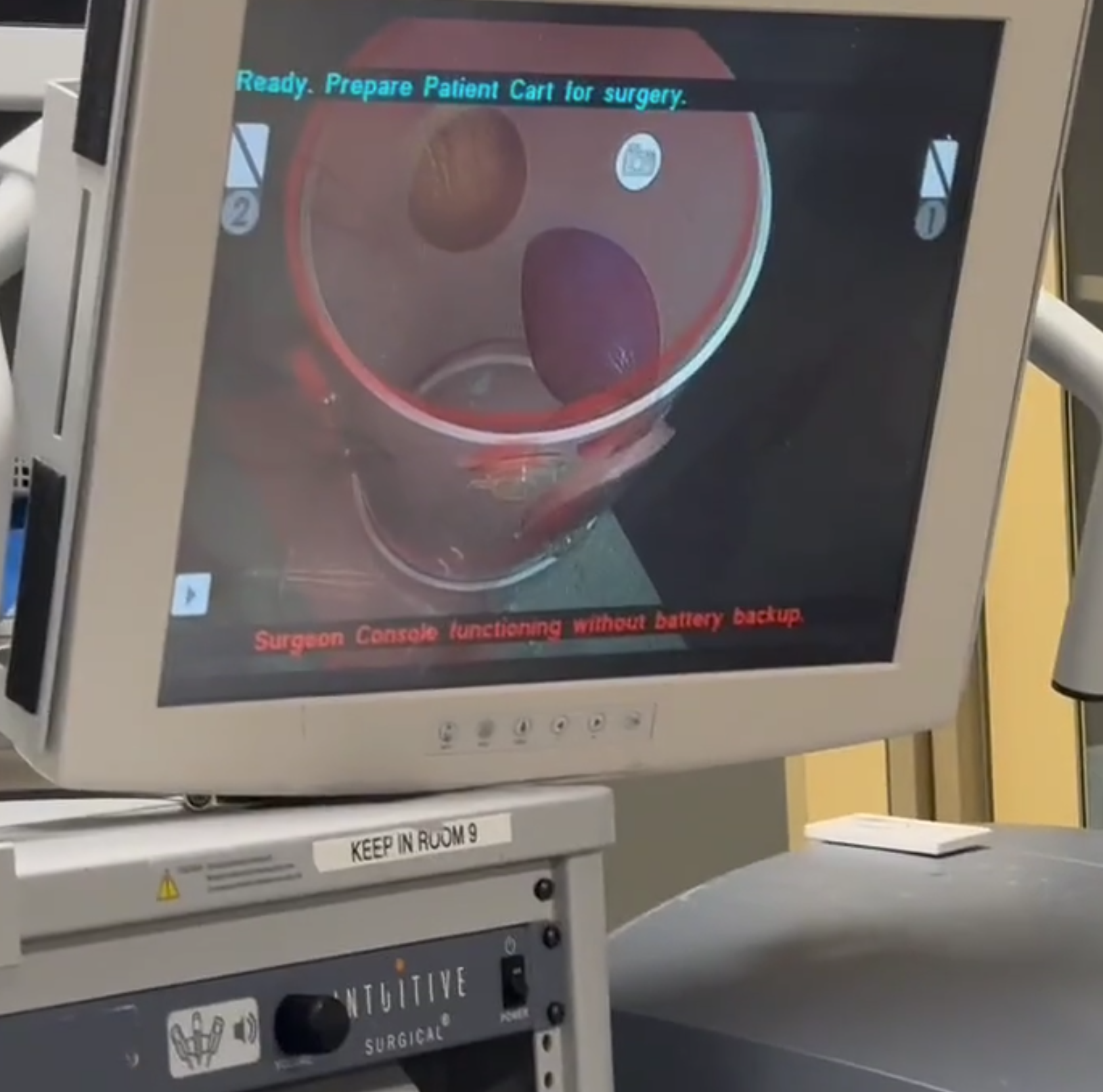}
        \caption{Augmented reality 3D overlay mode on surgeon console view}
        \label{fig:fig1}
    \end{subfigure}
    \begin{subfigure}{\linewidth}
        \centering
        \FigTwelvePanel{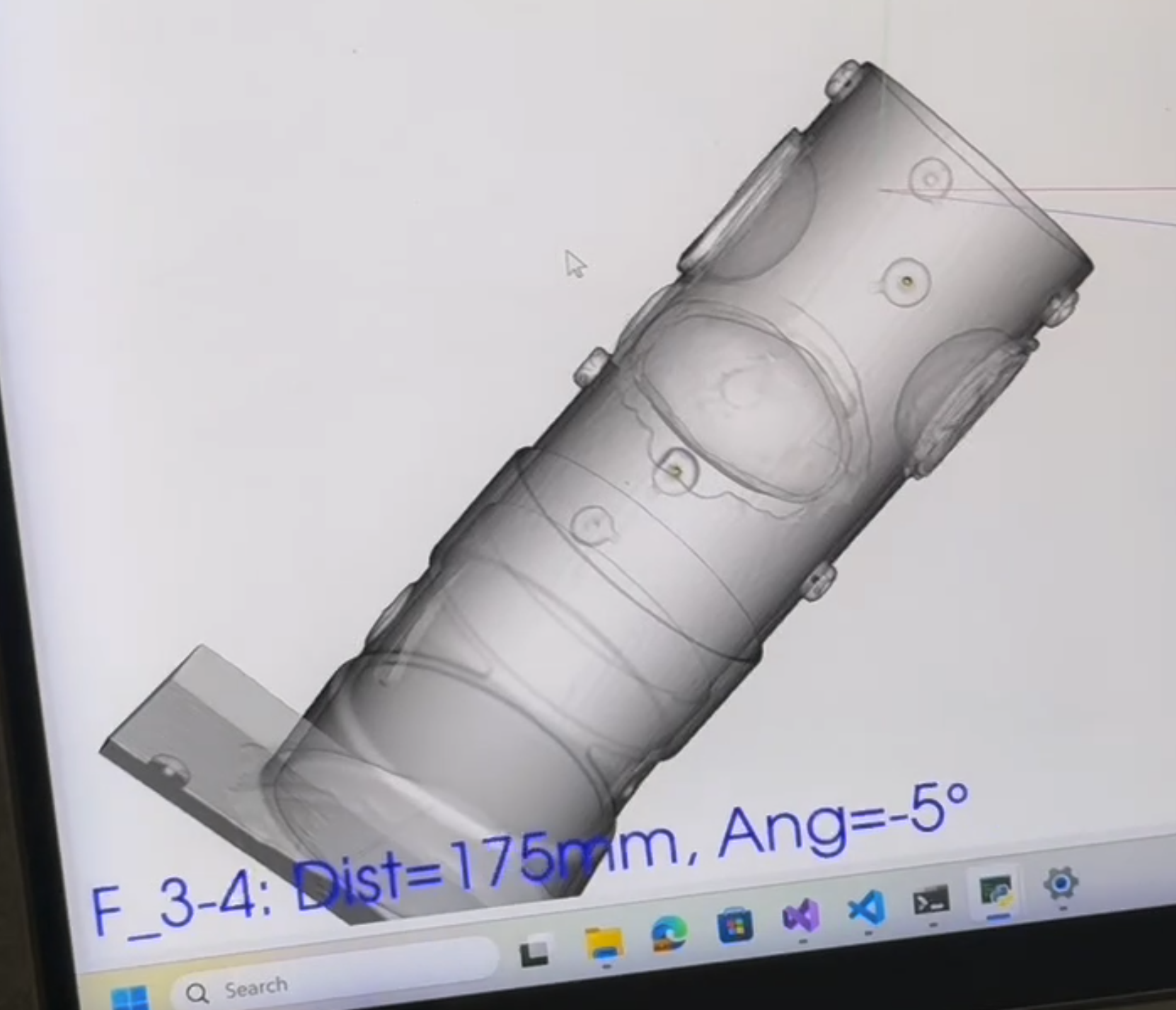}
        \caption{HUD-navigation mode on surgeon console view}
        \label{fig:fig2}
    \end{subfigure}
    \begin{subfigure}{\linewidth}
        \centering
        \FigTwelvePanel{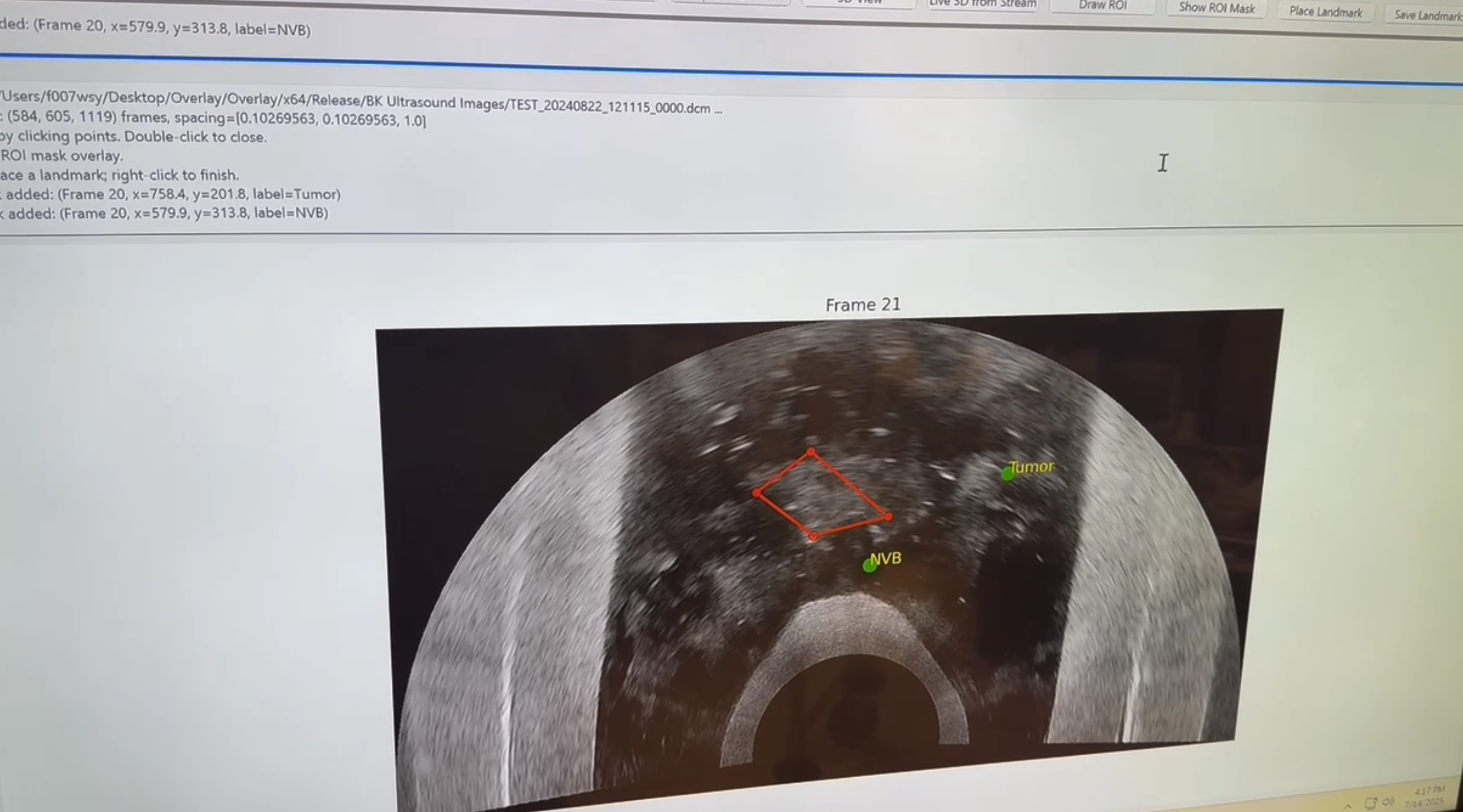}
        \caption{Ultrasound streaming mode on surgeon console view}
        \label{fig:fig3}
    \end{subfigure}
    \caption{Representative multimodal outputs for AR overlay accuracy, HUD tool-to-target guidance, ultrasound ROI/landmarks, and system performance logs.}
\label{fig:sys_integrated_Throat}
    \end{figure}
The MIRAGE framework was evaluated as a preclinical, system-level robotic surgery navigation platform spanning stereo video, tracking, AR overlay, stereo reconstruction, HUD navigation, and ultrasound. Representative surgeon-console views of the AR overlay, HUD navigation, and ultrasound modes are shown in Fig.~\ref{fig:sys_integrated_Throat}. Validation tested whether the architecture preserved geometric accuracy, temporal consistency, and real-time responsiveness across seven endpoints: stereo calibration/reprojection error, fiducial registration error, AR overlay projection error, stereo reconstruction error, temporal synchronization error, hardware-to-GUI latency, and platform-specific repeatability across da Vinci S,Si, Xi, and SP systems.
\subsection{Comparative Positioning Against Existing Systems}
Table~\ref{tab:comparison_prior_systems} positions MIRAGE against representative image-guided robotic surgery systems using system-level capabilities rather than isolated algorithms. The comparison shows that MIRAGE is not intended to replace individual calibration, registration, reconstruction, or ultrasound-fusion methods, instead, it integrates stereo reconstruction point clouds, AR video overlay, HUD-TilePro navigation, ultrasound, and video/rendering into a single console-attached, time-aware robotic surgery workflow.
\begin{table}[H]
\centering
\caption{System-level comparison with representative prior image-guided robotic surgery approaches.}
\label{tab:comparison_prior_systems}
\small\setlength{\tabcolsep}{3pt}
\newcolumntype{Y}{>{\raggedright\arraybackslash}X}
\begin{tabularx}{\linewidth}{>{\raggedright\arraybackslash}p{0.17\linewidth} >{\raggedright\arraybackslash}p{0.27\linewidth} Y Y Y}
\toprule
\textbf{Study/system} & \textbf{Primary capability} & \textbf{Console view} & \textbf{Live tracking} & \textbf{Stereo recon.} \\
\midrule
Ukimura et al.~\cite{Ukimura2014} & TilePro 3D prostate model & Yes & No & No \\
\addlinespace[3pt]
Samei et al.~\cite{Samei2020} & MRI/TRUS partial AR guidance & Yes & Yes & No \\
\addlinespace[3pt]
Song et al.~\cite{Song2024} & Arc-to-line US/PA registration & No & PM/TRUS & No \\
\addlinespace[3pt]
Meza et al.~\cite{Meza2021} & Freehand US + structured light & No & US probe & SL surface \\
\addlinespace[3pt]
Stoyanov~\cite{Stoyanov2010} & Real-time stereo reconstruction & No & No & Yes \\
\addlinespace[3pt]
\textbf{MIRAGE} & Time-synchronized multimodal navigation & Yes & Yes (EM/optical) & Yes \\
\bottomrule\end{tabularx}
\par\medskip
\begin{tabularx}{\linewidth}{>{\raggedright\arraybackslash}p{0.17\linewidth} Y Y Y >{\raggedright\arraybackslash}p{0.17\linewidth} Y}
\toprule
\textbf{Study/system} & \textbf{AR overlay} & \textbf{US} & \textbf{Time sync.} & \textbf{Latency} & \textbf{Multi-platform} \\
\midrule
Ukimura et al.~\cite{Ukimura2014} & Static/side-by-side & Preop TRUS/MRI & No & NA & No \\
\addlinespace[3pt]
Samei et al.~\cite{Samei2020} & Partial scene & Live TRUS & Limited & 200 ms & S/Si \\
\addlinespace[3pt]
Song et al.~\cite{Song2024} & No & TRUS/PA & No & NA & Si only \\
\addlinespace[3pt]
Meza et al.~\cite{Meza2021} & No & 3D freehand US & Simult. acq. & Offline & No \\
\addlinespace[3pt]
Stoyanov~\cite{Stoyanov2010} & No & No & No & 10--15 FPS & No \\
\addlinespace[3pt]
\textbf{MIRAGE} & Yes & Yes & Yes & 16.7 ms/stereo overlay; 200 ms/stereo reconstruction & S/Si/Xi/SP \\
\bottomrule\end{tabularx}
\par\medskip
\end{table}
At startup and after calibration, MIRAGE detects and configures available EM/optical tracking hardware, loads platform-specific imaging parameters, and launches the unified processing and visualization engine, as shown in figures~\ref{fig:block_diag} and~\ref{fig:chain_pipeline}, and~\ref{fig:GUIs} and formalized in \eqref{eq:modes}-\eqref{eq:display}. Rigid fiducial registration aligns the patient/phantom and preoperative MR/CT volume through an embedded GUI with real-time RMSE feedback and visual overlays; in practice, convergence is typically achieved in less than 30s.
\subsection{Fiducial Registration Accuracy}
 The fit residual in Section~\ref{op_regist} is the fiducial registration error (FRE). Error at a held-out point measures a different endpoint: leave-one-out evaluation refits the transform without that point, whereas independent-target TRE evaluates targets excluded from fitting~\cite{Fitzpatrick1998}. For $N_{\rm fid}$ fiducials, the $i$-th point is excluded, the transform is estimated from the remaining $N_{\rm fid}-1$ points, and the excluded CT point is transformed into tracker space:
\begin{equation}
e_{\mathrm{LOO},i}=
\left\|
\mathbf{R}_{-i}\mathbf{p}^{(i)}_{\mathrm{CT}} + \mathbf{t}_{-i} - \mathbf{p}^{(i)}_{\mathrm{tracker}}
\right\|_2
\end{equation}
Here $e_{\rm LOO,i}$ is the leave-one-out residual, and $\mathbf R_{-i}$ and $\mathbf t_{-i}$ are estimated without the $i$-th fiducial. Table~\ref{tab:fiducial_reg_accuracy} reports registration RMSE as mean~$\pm$~SD across trials. The original experiment logs stored RMSE feedback during registration.
\begin{table}[H]
\centering
\caption{Fiducial registration accuracy reporting structure across experiments.}
\label{tab:fiducial_reg_accuracy}
\small
\begin{tabular}{lc}
\toprule
\textbf{Platform}  & \textbf{RMSE mean $\pm$ SD (mm)} \\
\midrule
da Vinci S/Si  & 1.2 $\pm$ 0.20 \\
da Vinci Xi  & 0.9 $\pm$ 0.20 \\
da Vinci SP  & 0.9 $\pm$ 0.20\\
\bottomrule
\end{tabular}
\end{table}
Stereo calibration is performed preoperatively by acquiring multiple synchronized checkerboard image pairs and executing the stereo calibration/optimization pipeline to estimate intrinsic and extrinsic parameters (typically 1-2 min). The resulting intrinsics, distortion models, and rectification transforms are persisted and reused intraoperatively. 

Table~\ref{table:stereo_calib_res} reports physical errors of 0.1, 0.3, and 0.6~mm at working distances of 100, 300, and 500~mm, respectively. In the Xi optimization experiment (Table~\ref{tab:sys_calib}), 2D pixel error decreased from 4.3 to 2.5~px and 3D spatial error from 3.0 to 0.6~mm. These are separate working-distance and optimization results.

\begin{table}[H]
  \centering
  \caption{Calibration result before and after optimization on the Xi robot.}
  \label{tab:sys_calib}
  \begin{tabular}{lcc}
    \hline
    \textbf{Metric} & \textbf{Before Optimization} & \textbf{After Optimization} \\
    \hline
    2D pixel error   & 4.3 px & 2.5 px \\
    3D spatial error & 3.0 mm & 0.6 mm \\
    \hline
  \end{tabular}
\end{table}
Referring to Fig.~\ref{fig:calibration_chekcerboard}, Fig.~\ref{fig:Stereo_Reconstruction_res}, and Table~\ref{tab:sys_calib}, the system achieves low reprojection errors in both the AR overlay and stereo reconstruction modes. For stereo depth, both PSMNet and RAFT-Stereo were evaluated across multiple phantoms (throat, kidney, etc.). In general, RAFT-Stereo was faster and slightly more accurate overall.  For the right kidney, PSMNet yielded a lower reconstruction error than RAFT-Stereo (2.15 versus 2.25~mm); RAFT-Stereo had the lower error for the other three listed structures (Table~\ref{tab:stereo_recon_err}). Typical inference times were approximately 0.20s for RAFT-Stereo and 0.60s for PSMNet per stereo pair, motivating the choice of RAFT-Stereo as the default model.
\begin{table}[H]
\centering
\caption{3D reconstruction error (mm) between point cloud and Kidney phantom surfaces. Inference time is per stereo pair}
\label{tab:stereo_recon_err}
\begin{tabular}{lcc}
\toprule
\multicolumn{1}{c} {\textbf{Structure}} & \multicolumn{1}{c}{\textbf{PSMNet}} & \multicolumn{1}{c}{\textbf{RAFT-Stereo}} \\
\midrule
Abdominal Aorta (AA)   & 2.63 & 2.17 \\
Inferior Vena Cava (IVC)  & 2.30 & 1.79 \\
Right Kidney (RK)   & 2.15 & 2.25 \\
Right Ureter (RU)   & 2.43 & 2.13 \\
\midrule
Inference time & $\sim$0.60\,s & $\sim$0.20\,s \\
\bottomrule
\end{tabular}
\end{table}
\subsection{3D Overlay Accuracy}
AR overlay accuracy was evaluated by comparing projected model or calibration-target landmarks with manually identified corresponding landmarks in the endoscopic image. Let $\mathbf{X}^{CT}_k$ denote a 3D landmark in CT/model coordinates and $\mathbf{u}^{obs}_k$ its observed 2D image location. 

For a homogeneous CT landmark $\overline{\mathbf{X}}_k^{CT}=[(\mathbf{X}_k^{CT})^{\mathrm {T}},1]^{\mathrm {T}}$ and an undistorted pinhole image,
\begin{equation}
\mathbf{u}_k^{\rm pred}=\pi\!\left(\mathbf{K}_i
\begin{bmatrix}\mathbf{I}_3&\mathbf{0}\end{bmatrix}
\TF{CM_i}{CT}\overline{\mathbf{X}}_k^{CT}\right)
\end{equation}
For the selected eye $CM_i$, $\mathbf{K}_i\in\mathbb{R}^{3\times3}$, $\TF{CM_i}{CT}\in\mathbb{R}^{4\times4}$, and $\pi([x,y,z]^{\mathrm{T}})=[x/z,y/z]^{\mathrm{T}}$. Raw-image comparison includes the calibrated distortion model; rectified-image comparison uses the corresponding rectified camera parameters.
 The overlay projection error is
\begin{equation}
e_{\mathrm{overlay},k}=\left\|\mathbf{u}_k^{pred}-\mathbf{u}_k^{obs}\right\|_2
\end{equation}
Pixel error can be converted to approximate physical error using the calibrated working-distance-dependent pixel scale. Table~\ref{tab:ar_overlay_accuracy} reports the overlay-accuracy structure across platforms.
\begin{table}[H]
\centering
\caption{Mean AR overlay accuracy across phantoms.}
\label{tab:ar_overlay_accuracy}
\small
\setlength{\tabcolsep}{5pt}
\begin{tabular}{lcc}
\toprule
\textbf{Platform} & \textbf{Working distance (mm)} & \textbf{Error (px)} \\
\midrule
da Vinci S & 100 & 6.0 \\
da Vinci Si & 100 & 3.0 \\
da Vinci Xi    & 100 & 2.5 \\
da Vinci SP    & 100 & 2.5 \\
\bottomrule
\end{tabular}
\end{table}
Referring to Fig.~\ref{fig:chain_pipeline} and Fig.~\ref{fig:Software_architecture}, the system binds heterogeneous C++ subsystems with Python-native modalities under a single event loop and GUI. The optical tracker operates at 400~Hz, while the overlay pipeline streams left/right stereo frames at 60~FPS via the Bluefish frame grabber and renders the tracker-based overlay in real time. In-process frame access, copy-free C++ buffer exposure through the Python buffer protocol, worker threads, and batched callbacks reduce Python overhead in performance-critical loops.
Experimental videos \footnote{\url{https://sites.google.com/dartmouth.edu/halter-lab/resources/mirage}} 
demonstrate real-time operation of the framework and its modules, including user mode switching during experiments on different phantom models via the main GUI.
\begin{figure}[t]
\centering
\includegraphics[width=0.96\linewidth,height=0.64\textheight,keepaspectratio]{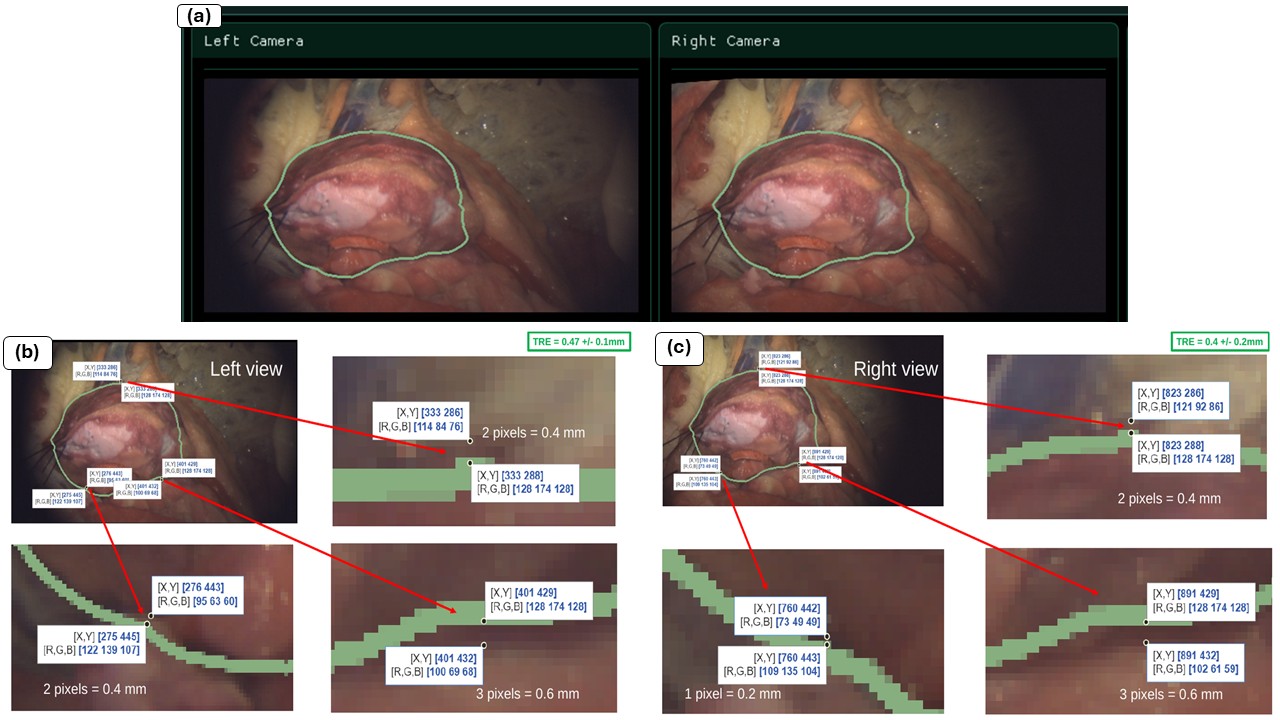}
\caption{Overlay accuracy analysis for Si: (a) Left-Right  live overlay views in MIRAGE environment. (b) Left-eye contour checks: 0.4, 0.4, and 0.6 mm, reported as $0.47\pm0.10$ mm. (c) Right-eye checks: 0.4, 0.2, and 0.6 mm, reported as $0.40\pm0.20$ mm. The pixel checks use the documented MIRAGE scale of 0.20 mm/pixel.}
\label{fig:IV&V_validation}
\end{figure}
\subsection{Temporal Synchronization and End-to-End Latency}
System responsiveness was evaluated from timestamped video-acquisition, tracker, rendering, and GUI-display logs when available.  Backend-arrival-to-display latency is defined as,
\begin{equation}
\Delta t_{\mathrm{total}} = t_{\mathrm{display}} - t_{\mathrm{arrival}}
\end{equation}
where $t_{\mathrm{arrival}}$ is assigned when a frame reaches the acquisition backend and $t_{\mathrm{display}}$ when the corresponding frame or overlay is presented. For non-overlapping consecutive stages after backend arrival, the latency is decomposed as,
\begin{equation}
\Delta t_{\mathrm{total}} = \Delta t_{\mathrm{acq}} + \Delta t_{\mathrm{pose}} + \Delta t_{\mathrm{render}} + \Delta t_{\mathrm{transfer}} + \Delta t_{\mathrm{display}}
\end{equation}
Here $\Delta t_{\rm acq}$ is backend acquisition/queue time after arrival; the remaining terms are pose association, rendering, buffer transfer, and presentation time, respectively. The pose-association timestamp offset is the absolute time difference between each video frame and the tracker pose used for rendering:
\begin{equation}
e_{\mathrm{assoc}} = |t_v - t_{\hat{T}}|
\end{equation}

Here $t_{\hat{T}}$ is the timestamp assigned to the selected or interpolated pose. For interpolation at $t_v$, $e_{\rm assoc}=0$ by construction and is not an independent measure of physical synchronization accuracy; source offsets $a_v$ and $h_v$ retain the definitions in Section~\ref{sec:temporal_sync}.

The nominal stream rates were 60~FPS for video and 400~Hz for optical tracking. Table~\ref{tab:sync_latency} summarizes the video frame rates and tracker update rates for each operating mode. A 60~FPS stream has a 16.7~ms paired-frame display latency. The reported 200~ms reconstruction time corresponds to approximately 0.20~s of inference per stereo pair.

\begin{figure}[t]
\centering
\includegraphics[width=0.96\linewidth,height=0.64\textheight,keepaspectratio]{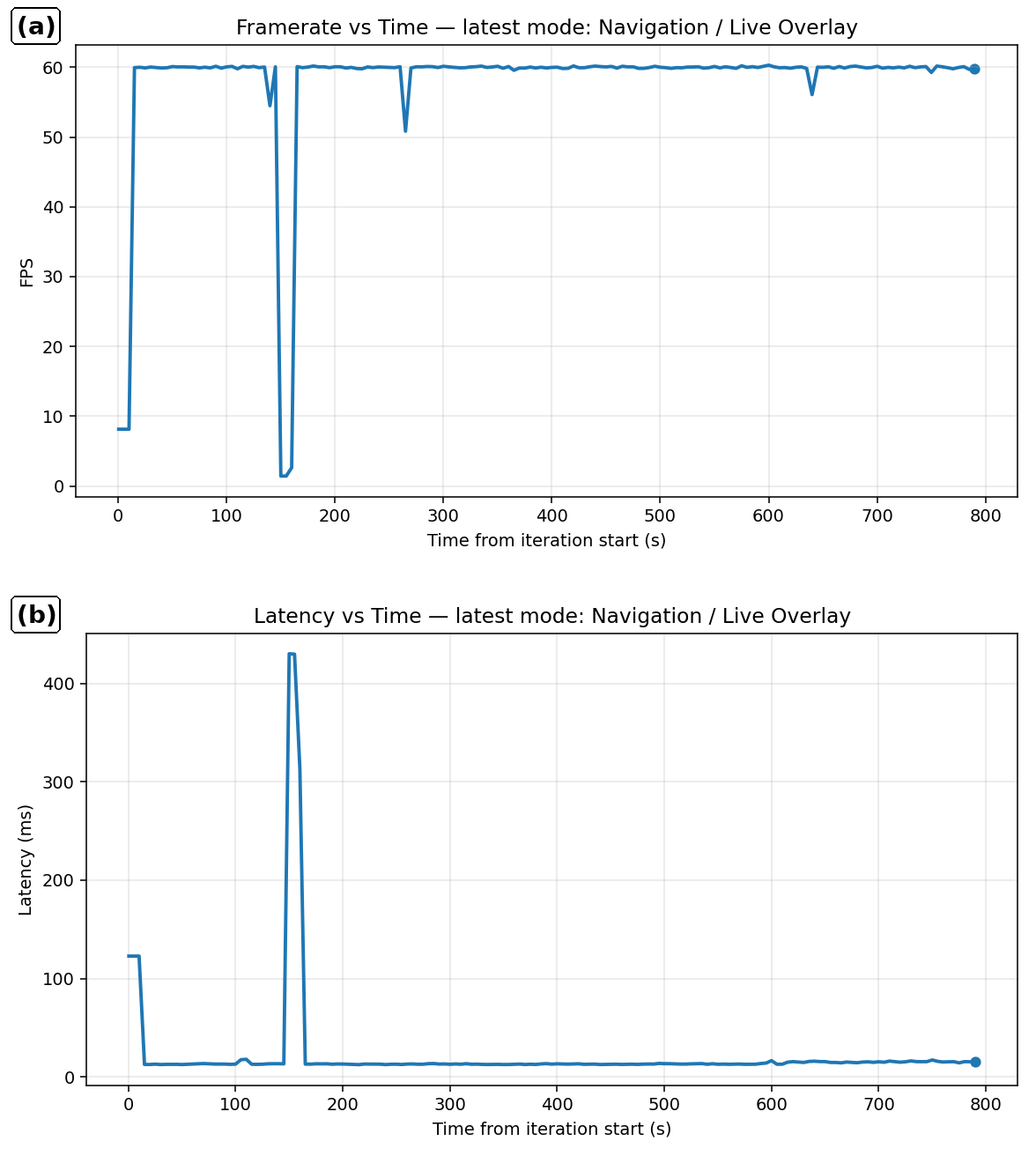}
\caption{Supplied live overlay runtime traces: (a) navigation frame rate and (b) navigation-processing time.}
\label{fig:runtimeplots}
\end{figure}

\begin{figure}
\centering
\includegraphics[width=0.96\linewidth,height=0.64\textheight,keepaspectratio]{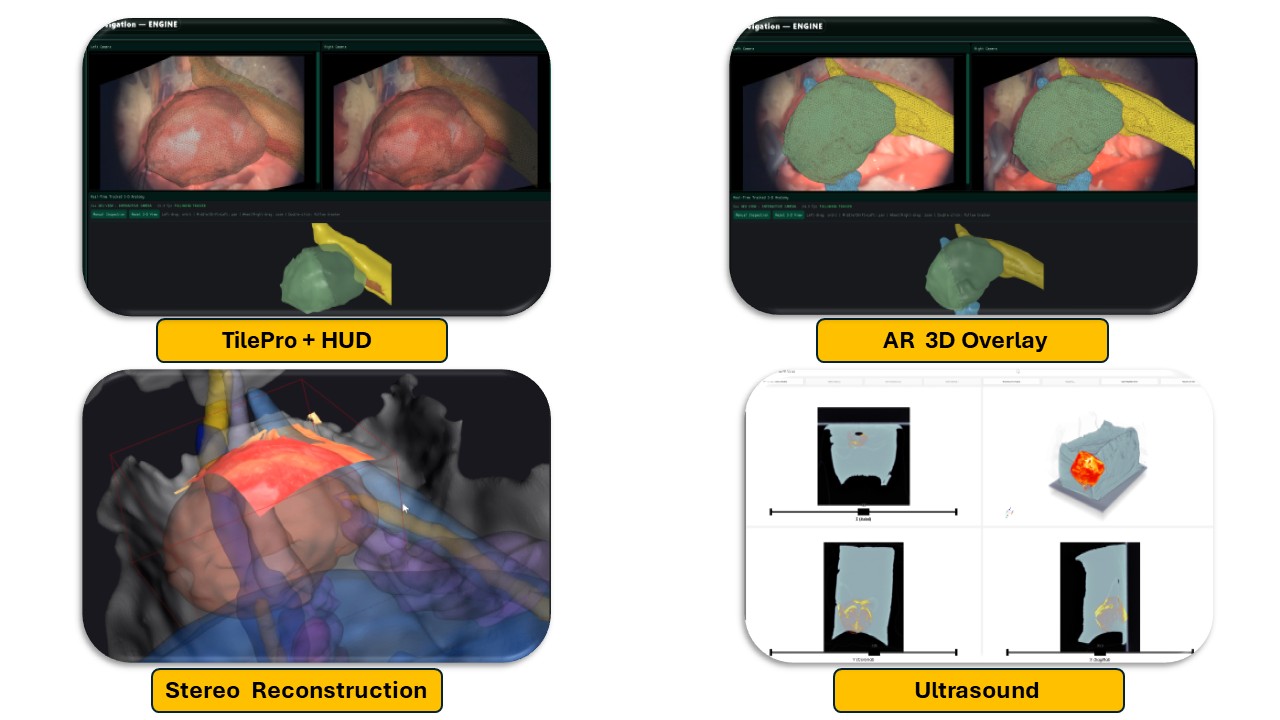}
\caption{MIRAGE integrated modules in the surgical phantom experiment.}
\label{fig:All-modes}
\end{figure}

Figures~\ref{fig:IV&V_validation} and~\ref{fig:runtimeplots} show the accuracy and performance of MIRAGE navigation on the surgeon console. Gaps in the plot correspond to periods during which the user paused the real-time overlay to perform another registration using ultrasound or fiducial-based registration, or to switch between modalities, such as from video overlay to point-cloud reconstruction, which required restarting the trackers.
\begin{table}[H]
\centering
\caption{Synchronization and latency reporting structure.}
\label{tab:sync_latency}
\begin{tabular}{lcc}
\toprule
\textbf{Mode} & \textbf{Video FPS} & \textbf{Tracker rate} \\ 
\midrule
 AR overlay & 60 & 400 Hz \\
 HUD & 60 & 40 Hz \\
 Stereo reconstruction & 60 & 400 Hz \\
 Ultrasound & 60 & 40 Hz \\
\bottomrule
\end{tabular}
\end{table}
The complete MIRAGE framework and binding-wrapper pipelines, including registration, stereo calibration, HUD navigation, AR overlay, stereo reconstruction, and ultrasound, were evaluated in multiple experiments involving several phantom models and four da Vinci platforms: S, Si, Xi, and SP. Referring to Fig.~\ref{fig:Phantom}, the experiments were conducted using multiple phantom models.  At the listed 100~mm working distance, mean overlay errors were 6.0~px for S, 3.0~px for Si, and 2.5~px for Xi and SP (Table~\ref{tab:ar_overlay_accuracy}). These platform-specific values characterize the tested configurations; differences in camera calibration, image format, and trial conditions remain relevant to their comparison. 
MIRAGE automatically identifies the endoscope streaming format and size and configures the pipeline accordingly. Representative throat and surgical phantom experiments are shown in Figs.~\ref{fig:sys_integrated_Throat} and~\ref{fig:All-modes}.

\section{Limitations and Translational Evaluation Requirements}
This phantom-based study provides preclinical technical and workflow validation, but does not establish clinical effectiveness, patient benefit, or superiority over existing workflows. The evaluation does not yet include formal human-factors testing, surgeon workload analysis, learning curves, clinical outcomes, or long-term monitoring. Several modules use mature methods, including rigid registration, camera calibration, stereo reconstruction, and projection-based AR rendering; the contribution is their synchronized robotic-console integration, quantitative evaluation, and cross-language binding architecture rather than a new isolated calibration or reconstruction algorithm. Platform-specific validation is also limited by the number of experiments per da Vinci system, especially SP.
Future translational evaluation should include structured surgeon usability studies, formal latency testing under clinical workflow conditions, failure-mode and effects analysis, overlay-failure detection, stale-transform warning thresholds, learning-curve assessment, patient-centered outcomes, cost-effectiveness, sustainability, and long-term monitoring. Consistent with IDEAL Robotics recommendations~\cite{Marcus2024}, subsequent work should progress from technical preclinical validation toward prospective workflow evaluation, clinical usability assessment, and real-world surveillance.
\section{Conclusion}
This work presents MIRAGE, a time-synchronized multimodal image-guided robotic surgery framework for real-time robotic surgical navigation. The system integrates stereo reconstruction point cloud visualization, video (AR) overlay, HUD-TilePro navigation, ultrasound imaging, optical/electromagnetic tracking, stereo calibration, and rigid registration within a unified surgeon-facing workflow. The central contribution is not an isolated calibration, registration, or reconstruction algorithm, but a quantitatively evaluated architecture for synchronizing and coordinating heterogeneous imaging and tracking streams during robotic surgery.

At system startup, and following calibration, the framework initializes all essential user interface and data processing modules, beginning with the detection and configuration of available tracking hardware, including both electromagnetic and optical tracking systems. Users are then prompted to provide configuration files containing necessary parameters for the robotic platform and associated imaging sources. Once configuration is confirmed, the core data processing and visualization engine is launched, enabling high-speed data exchange and control between low-level modules and the user interface.

The unified framework supports multiple real-time modalities within a single, interactive interface. Each modality offers dedicated visualization and control panels, allowing users to enable or disable live 3D stereo video overlay, show or hide specific overlay components, and adjust parameters such as transparency and texture. Users can also interactively select anatomical regions of interest, view live distance and angle readouts from instrument tracking, display input imagery through TilePro, and capture video or individual frames. Additional features may be integrated based on clinical feedback and evolving requirements.
The framework was validated through phantom experiments conducted on the da Vinci S, Si, Xi, and SP platforms. The results demonstrate technical feasibility, cross-platform deployment, real-time multimodal visualization, calibration accuracy, stereo reconstruction capability, and integrated navigation.  The evaluation separates calibration, registration, overlay, reconstruction, and runtime endpoints to characterize the integrated workflow at the component and system levels.
MIRAGE should therefore be interpreted as a preclinical technical and workflow-validation platform. Consistent with IDEAL robotics recommendations~\cite{Marcus2024}, future work should extend beyond technical validation to include formal human-factors assessment, surgeon learning curves, safety and failure-mode analysis, patient-centered evaluation, health-economic assessment, sustainability analysis, and long-term real-world monitoring.
\section{future works}
Ongoing work will extend MIRAGE with additional modalities, including multi-excitation wavelength emission-ratio (MEWER) depth mapping~\cite{Torres2025}, related video streams, video-based calibration, continuous point-cloud reconstruction, and automatic overlay-quality assessment. Current calibration uses discrete checkerboard image pairs to estimate intrinsic/extrinsic parameters; the next iteration will ingest continuous video to support calibration monitoring and real-time point-cloud updates.
Future evaluation will follow a staged translational pathway, beginning with simulated-procedure studies with surgeons to quantify usability, cognitive workload, task efficiency, overlay trust, and workflow integration. Additional studies will assess learning curves, stale-transform detection, overlay-failure alerts, dropped-frame behavior, and registration drift. If the system progresses toward clinical use, prospective protocols will evaluate patient-centered outcomes, surgeon training requirements, consent language, cost-effectiveness, sustainability, and long-term real-world performance.
Further technical evaluation will compare matched image-guidance tasks using common hardware, data, timing boundaries, and target definitions. Controlled ablations of temporal association and frame-transfer policy will examine their effects on dynamic overlay error, source-data age, display latency, and frame loss. Repeated trials across platforms will quantify variation with working distance, motion, calibration, and configuration, using independent targets and trial-level uncertainty estimates.

\section*{Funding Data}
This work was supported in part by the Advanced Research Projects Agency for Health (ARPA-H) under award number 1AY2AX000043. The
views and conclusions contained in this document are those of the authors and should not be interpreted as representing the official policies, either expressed or implied, of the United States Government.

\section*{Conflict of Interest}
U.S. provisional patent applications covering aspects of this work have been filed (U.S. Provisional Application Nos. 63/920,165 and 64/166,677).

\section*{Data Availability Statement}
The article includes the reported quantitative results. A public repository for the raw acquisition logs and the complete MIRAGE implementation is not provided with this submission. The implementation includes proprietary hardware interfaces.

\appendices
\section{Rigid Registration, Navigation, and Temporal Synchronization}
\label{app:registration_sync}

\subsection{Optical Tracker Registration}
The rigid registration process begins by initializing the GUI and hardware tracker. Preoperatively acquired CT fiducials are loaded as $\mathbf{P}_{CT}\in\mathbb{R}^{N_{\rm fid}\times3}$, and the corresponding stylus-tip positions are collected as $\mathbf{P}_{PO}\in\mathbb{R}^{N_{\rm fid}\times3}$. The GUI displays both sets and allows the operator to edit, remove, or clear acquired points.

The centroids of the two point sets are
\begin{equation}
\bar{\mathbf{p}}_{CT}
=\frac{1}{N_{\rm fid}}\sum_{i=1}^{N_{\rm fid}}\mathbf{p}_{CT}^{(i)},
\end{equation}
\begin{equation}
\bar{\mathbf{p}}_{PO}
=\frac{1}{N_{\rm fid}}\sum_{i=1}^{N_{\rm fid}}\mathbf{p}_{PO}^{(i)}
\end{equation}
The centered point sets are
\begin{equation}
\mathbf{X}_{\rm reg}=\mathbf{P}_{CT}-\mathbf{1}_{N_{\rm fid}}\bar{\mathbf{p}}_{CT}^{\mathrm{T}},
\end{equation}
\begin{equation}
\mathbf{Y}_{\rm reg}=\mathbf{P}_{PO}-\mathbf{1}_{N_{\rm fid}}\bar{\mathbf{p}}_{PO}^{\mathrm{T}}
\end{equation}
Here $\mathbf{1}_{N_{\rm fid}}$ is the all-ones column vector; each row of a point-set matrix contains the transpose of a 3D point. The covariance matrix and its singular value decomposition are
\begin{equation}
\mathbf{H}_{\rm reg}=\mathbf{X}_{\rm reg}^{\mathrm{T}}\mathbf{Y}_{\rm reg},
\qquad
\mathbf{H}_{\rm reg}\in\mathbb{R}^{3\times3},
\end{equation}
\begin{equation}
\mathbf{H}_{\rm reg}=\mathbf{U}_{\rm reg}\boldsymbol{\Sigma}_{\rm reg}\mathbf{V}_{\rm reg}^{\mathrm{T}}
\end{equation}
which gives
\begin{equation}
\mathbf{R}_{\rm reg}=\mathbf{V}_{\rm reg}\mathbf{U}_{\rm reg}^{\mathrm{T}}
\end{equation}
If $\det(\mathbf{R}_{\rm reg})<0$, the sign of the last column of $\mathbf{V}_{\rm reg}$ is reversed and $\mathbf{R}_{\rm reg}$ is recomputed to ensure a proper rotation. The translation is
\begin{equation}
\mathbf{t}_{\rm reg}
=\bar{\mathbf{p}}_{PO}
-\mathbf{R}_{\rm reg}\bar{\mathbf{p}}_{CT}
\end{equation}
The CT-to-optical-tracker transform is
\begin{equation}
\TF{PO}{CT}
=
\begin{bmatrix}
\mathbf{R}_{\rm reg} & \mathbf{t}_{\rm reg}\\
\mathbf{0}_{1\times3} & 1
\end{bmatrix}
\end{equation}
and registration quality is quantified by
\begin{equation}
\mathrm{RMSE}
=
\sqrt{\frac{1}{N_{\rm fid}}\sum_{i=1}^{N_{\rm fid}}
\left\|
\mathbf{R}_{\rm reg}\mathbf{p}_{CT}^{(i)}
+\mathbf{t}_{\rm reg}
-\mathbf{p}_{PO}^{(i)}
\right\|^{2}}
\end{equation}
The GUI displays and saves the resulting transformation and RMSE.

\subsection{Optical and Electromagnetic Registration Workflow}
The electromagnetic pipeline acquires the same CT fiducials with the Field Generator $\{FG\}$ and applies the preceding SVD solution. The complete acquisition and transform-composition workflow is summarized in Algorithm~\ref{alg:rigid_reg_trackers_appendix}.

\begin{algorithm}[H]
\caption{Rigid Registration Pipeline for Optical and Electromagnetic Trackers}
\label{alg:rigid_reg_trackers_appendix}
\small
\KwIn{CT fiducials $\mathbf{P}_{CT}$, optical tracker, electromagnetic tracker}
\KwOut{$\TF{PO}{CT}$, $\TF{FG}{CT}$, $\TF{PO}{FG}$, RMSE}
\vspace{0.7em}
\hspace{2em}\textbf{"Optical Tracker Registration":}\\
Load $\mathbf{P}_{CT}$ from file\;
Initialize PO list\;
\While{collected PO points $<$ required}{
    Collect tip position using optical tracker\;
    Append to $\mathbf{P}_{PO}$\;
    \If{user edits/removes/clears}{
        Update $\mathbf{P}_{PO}$\;
    }
}
\If{collected PO points == required}{
    Compute rotation (SVD)\;
    Adjust if improper rotation\;
    Compute translation\;
    Form $\TF{PO}{CT}$\;
    Calculate RMSE\;
    Display/save $\TF{PO}{CT}$, RMSE\;
}
\vspace{0.25em}
\noindent\makebox[0.85\linewidth]{\rule{0.7\linewidth}{0.7pt}}
\vspace{0.25em}
\hspace{2em}\textbf{"EM Tracker Registration":}\\
Load $\mathbf{P}_{CT}$ from file\;
Initialize FG list\;
\While{collected FG points $<$ required}{
    Collect tip position using EM tracker\;
    Append to $\mathbf{P}_{FG}$\;
    \If{user edits/removes/clears}{
        Update $\mathbf{P}_{FG}$\;
    }
}
\If{collected FG points == required}{
    Compute rotation (SVD)\;
    Adjust if improper rotation\;
    Compute translation\;
    Form $\TF{FG}{CT}$\;
    Calculate RMSE\;
    Display/save $\TF{FG}{CT}$, RMSE\;

}
\textbf{Composing Transforms:}\\
Compute $\TF{PO}{FG}
=\TF{PO}{CT}\TF{CT}{FG}$
\end{algorithm}
The EM point fit yields $\TF{FG}{CT}$. The inverse needed for tracker-to-tracker composition is,
\begin{align}
\TF{CT}{FG}=\left(\TF{FG}{CT}\right)^{-1},\\
\TF{PO}{FG}=\TF{PO}{CT}\TF{CT}{FG}
\end{align}
The fitted and inverse transforms therefore have distinct source and destination frames.

\subsection{Real-Time Navigation Filtering}\label{app:pose_filtering}
Both trackers continuously provide homogeneous tool poses. After composition with the saved CT registrations, the pose update is smoothed as
\begin{equation}
\widetilde{\TF{CT}{tool}}
=
\alpha_{\rm nav}\TF{CT}{tool}^{\rm cur}
+(1-\alpha_{\rm nav})\widehat{\TF{CT}{tool}}^{\rm prev}
\end{equation}
where $\TF{CT}{tool}^{\rm cur}$ is the most recent transform, $\widehat{\TF{CT}{tool}}^{\rm prev}$ is the previously filtered transform, and $\alpha_{\rm nav}$ is the current-sample weight, with $\alpha_{\rm nav}\in[0,1]$. Because element-wise filtering does not preserve rotational orthonormality, the rotational block is projected back onto $SO(3)$. If
\begin{equation}
\widetilde{\TF{CT}{tool}}[:3,:3]
=
\mathbf{U}_{\rm nav}\boldsymbol{\Sigma}_{\rm nav}\mathbf{V}_{\rm nav}^{\mathrm{T}},
\end{equation}
then
\begin{equation}
\mathbf{R}_{\rm nav}
=
\mathbf{U}_{\rm nav}\operatorname{diag}\!\left(1,1,\det(\mathbf{U}_{\rm nav}\mathbf{V}_{\rm nav}^{\mathrm{T}})\right)\mathbf{V}_{\rm nav}^{\mathrm{T}}
\end{equation}
The rigid filtered pose $\widehat{\TF{CT}{tool}}$ uses $\mathbf R_{\rm nav}$ as its rotation, the translation of the intermediate matrix $\widetilde{\TF{CT}{tool}}$, and bottom row $[0,0,0,1]$. The filtered transforms update the anatomical models, instruments, and targets in the common CT frame.

\subsection{Temporal Synchronization and Stream Association}
Video, tracking, rendering, ultrasound, and GUI streams are timestamped with a monotonic host clock upon acquisition or arrival. Each sample is stored as
\begin{equation}
\mathcal{D}_{{\rm sample},i}=(x_i,t_i,\ell_{{\rm stream},i})
\end{equation}
where $x_i$, $t_i$, and $\ell_{{\rm stream},i}$ denote the payload, timestamp, and stream source, respectively.

For each video timestamp $t_v$, tracker samples $\mathbf{T}_i$ and $\mathbf{T}_j$ for the same fixed frame pair, as defined in Section~\ref{sec:temporal_sync}, are selected such that
\begin{equation}
t_i\leq t_v\leq t_j,\qquad t_i<t_j,
\end{equation}
with interpolation factor
\begin{equation}
\beta=\frac{t_v-t_i}{t_j-t_i}
\end{equation}
Translation is linearly interpolated:
\begin{equation}
\hat{\mathbf{p}}(t_v)
=(1-\beta)\mathbf{p}_i+\beta\mathbf{p}_j
\end{equation}
whereas rotation uses spherical linear interpolation (SLERP) of the associated unit quaternions, returned as a rotation matrix:
\begin{equation}
\hat{\mathbf{R}}(t_v)
=\mathrm{SLERP}(\mathbf{R}_i,\mathbf{R}_j,\beta)
\end{equation}
The synchronized transform is
\begin{equation}
\hat{\mathbf{T}}(t_v)
=
\begin{bmatrix}
\hat{\mathbf{R}}(t_v) & \hat{\mathbf{p}}(t_v)\\
\mathbf{0}_{1\times3} & 1
\end{bmatrix}
\end{equation}
If no valid pose exists within $\Delta t_{\max}$, the overlay is marked stale and either frozen at the last valid transform or disabled according to the safety policy. 

For a bracketing association, the supporting sample offsets are $t_v-t_i$ and $t_j-t_v$, with $t_i<t_j$. Their maximum $a_v=\max(t_v-t_i,t_j-t_v)$ and the bracket width $h_v=t_j-t_i$ characterize temporal support, matching Section~\ref{sec:temporal_sync}. For nearest-sample association, the offset is $|t_v-t_i|$. An interpolated pose assigned the timestamp $t_v$ does not independently establish physical synchronization accuracy.

 \section{AR Video Overlay and Stereo Calibration Details}
\label{app:ar_overlay}

\subsection{Overlay Pipeline}
The overlay framework integrates real-time optical tracking, camera calibration, and OpenGL-based stereo visualization. The pipeline operates on an Exaact PC equipped with a Kronos K8 SDI I/O card and an RTX 6000 Ada GPU and comprises the following stages:
\begin{enumerate}
\item \textbf{Tracker input:} The optical tracker acquires a 6-DoF homogeneous pose for each tracked tool.
\item \textbf{Model loading:} Patient/phantom CT data, instrument models, and registration transforms are loaded from YAML files.
\item \textbf{Calibration parameters:} Camera intrinsics $\mathbf{K}$ and left/right extrinsics are parsed.
\item \textbf{Pose smoothing:} Tracker output is filtered to reduce jitter.
\item \textbf{Rendering:} The current model poses are rendered independently for the left and right eyes.
\item \textbf{Image fusion:} Each virtual frame is blended with its corresponding video frame.
\item \textbf{User interaction:} GUI panels control mesh visibility, group visibility, alignment offsets, opacity, and display modes.
\end{enumerate}

\subsection{Optical Tracking and Pose Smoothing}
Let $\TF{PO}{CT}$ be the fixed CT-to-tracker registration, $\TF{CM_i}{PO}$ the tracker-to-eye transform for eye $CM_i\in\{LE,RE\}$, and $\TF{PO}{tool}$ the current tracked tool pose. Dynamic objects are updated per frame by
\begin{equation}
\TF{CM_i}{tool}
=
\TF{CM_i}{PO}\,
\TF{PO}{tool}
\end{equation}

To stabilize the overlay, the estimated pose is exponentially smoothed. For overlay update index $k_{\rm AR}$, the intermediate matrix is:
\begin{equation}
\widetilde{\TF{CM_i}{tool}}^{(k_{\rm AR})}
=
\alpha_{\rm AR}\widehat{\TF{CM_i}{tool}}^{(k_{\rm AR}-1)}
+(1-\alpha_{\rm AR})\TF{CM_i}{tool}^{(k_{\rm AR})}
\end{equation}
Here $\alpha_{\rm AR}\in[0,1]$ is the previous-estimate weight (e.g., $0.9$). The rigid smoothed pose $\widehat{\TF{CM_i}{tool}}^{(k_{\rm AR})}$ retains the blended translation and uses the determinant-corrected $SO(3)$ projection of the rotation block specified in Appendix~\ref{app:pose_filtering}.

\subsection{Camera Projection and Rendering}
For a CT-referenced scene, the left-eye view matrix uses the inverse left-eye pose in CT space. With image coordinates $u$ rightward and $v$ downward, the optical-to-OpenGL axis conversion is $\TF{GL_{LE}}{LE}=\operatorname{diag}(1,-1,-1,1)$, giving
\begin{equation}
\mathbf{V}_{\rm view,LE}=\TF{GL_{LE}}{LE}\,\TF{LE}{CT}=\TF{GL_{LE}}{LE}\left(\TF{CT}{LE}\right)^{-1}
\end{equation}
Given the corresponding left-eye intrinsic matrix $\mathbf K=\mathbf K_L$, viewport width $W_{\rm img}$, viewport height $H_{\rm img}$, and near clipping plane $z_{\rm near}$, the focal lengths and principal point are
\begin{align}
f_x=\mathbf{K}_{00},
\qquad
f_y=\mathbf{K}_{11},\\
c_x=\mathbf{K}_{02},
\qquad
c_y=\mathbf{K}_{12}
\end{align}
The corresponding OpenGL frustum parameters are
\begin{align}
\mathrm{left}=-\frac{c_xz_{\rm near}}{f_x},\\
\mathrm{right}=\frac{(W_{\rm img}-c_x)z_{\rm near}}{f_x},\\
\mathrm{bottom}=-\frac{(H_{\rm img}-c_y)z_{\rm near}}{f_y},\\
\mathrm{top}=\frac{c_yz_{\rm near}}{f_y}
\end{align}
These parameters define the projection matrix used to render the anatomical model into each endoscope view.

The CT, tracker, and camera coordinate systems are connected by homogeneous transforms of the form
\begin{equation}
\TF{A}{B}
=
\begin{bmatrix}
\RF{A}{B} & \TV{A}{B}\\
\mathbf{0}_{1\times3} & 1
\end{bmatrix}
\end{equation}
where $\RF{A}{B}\in SO(3)$ and $\TV{A}{B}\in\mathbb{R}^{3}$. The main paper retains the composite CT-to-camera transform and alpha-blending equation.
\subsection{Real-Time Execution}
Three dedicated threads run concurrently. The rendering thread performs OpenGL/ImGui updates, renders the stereo scene, and writes output buffers. The tracker thread queries the hardware, smooths measurements, and updates tool poses. The video-I/O thread transfers captured and rendered frames through the Bluefish444 KRONOS K8\textsuperscript{\tiny{\textregistered}} card. This organization supports independent input-video, virtual-render, and fused-overlay modes for both endoscope eyes.

\subsection{Checkerboard Acquisition and Projection}
For every calibration frame, the stereo images and tracker poses of the endoscope and checkerboard are recorded simultaneously. CLAHE preprocessing, corner detection, and subpixel refinement yield checkerboard observations $\{\mathbf{c}_{i,k}^{(f)}\}$. The checkerboard geometry is synthesized in 3D and transformed into camera coordinates using the current tracker-to-camera and checkerboard transforms.

After applying the camera-frame transform retained in the main paper, let
\begin{equation}
\begin{bmatrix}x\\y\\z\end{bmatrix}
=\begin{bmatrix}\mathbf I_3&\mathbf0\end{bmatrix}\bar{\mathbf X}_{i,k}^{(f)}
\end{equation}
The normalized undistorted image coordinates are,
\begin{equation}
\begin{bmatrix}x'\\y'\end{bmatrix}
=\begin{bmatrix}x/z\\y/z\end{bmatrix}
\end{equation}
\subsection{Explicit Reprojection Model}
For $\bar{\mathbf{X}}_k=[X,Y,Z,1]^{\mathrm{T}}$, let $\bar{\mathbf X}_{i,k}^{(f)}=[X^{(C)},Y^{(C)},Z^{(C)},1]^{\mathrm{T}}$. For this observation, $x=X^{(C)}$, $y=Y^{(C)}$, and $z=Z^{(C)}$. The complete radial/tangential distortion model is,
\begin{align}
x'&=\frac{X^{(C)}}{Z^{(C)}},
\qquad
y'=\frac{Y^{(C)}}{Z^{(C)}},\\
r_{\rm dist}^{2}&=x'^{2}+y'^{2},\\
x_d
&=
x'\!\left(1+k_1r_{\rm dist}^2+k_2r_{\rm dist}^4+k_3r_{\rm dist}^6\right)
+2p_1x'y'
\notag\\
&\qquad
+p_2\!\left(r_{\rm dist}^2+2x'^{2}\right),\\
y_d
&=
y'\!\left(1+k_1r_{\rm dist}^2+k_2r_{\rm dist}^4+k_3r_{\rm dist}^6\right)
+p_1\!\left(r_{\rm dist}^2+2y'^{2}\right)
\notag\\
&\qquad
+2p_2x'y',\\
u&=f_xx_d+c_x,\\
v&=f_yy_d+c_y
\end{align}
Here $(u,v)$ is the predicted pixel coordinate, $(f_x,f_y)$ are focal lengths, $(c_x,c_y)$ are principal-point coordinates, and
\begin{equation}
\mathbf{D}_{i}
=
\begin{bmatrix}
k_1 & k_2 & k_3 & p_1 & p_2
\end{bmatrix}
\end{equation}
contains the distortion coefficients for eye $i$. In the scalar formulas above, the eye index is suppressed; the focal lengths, principal point, and distortion coefficients belong to the same $\mathbf K_i,\mathbf D_i$ used in $\boldsymbol{\theta}_{\rm calib}$.

\subsection{Optimization Implementation}
The reprojection objective and $SE(3)$ exponential-map parameterization are retained in the main paper. The cost function is implemented as one residual block with two parameter groups. The first is the nine-dimensional intrinsic/distortion vector
\begin{equation}
\boldsymbol{\theta}_{{\rm int},i}
=
\begin{bmatrix}
f_x & f_y & c_x & c_y & k_1 & k_2 & k_3 & p_1 & p_2
\end{bmatrix}^{\mathrm{T}}
\end{equation}
and the second parameterizes the fixed marker-to-eye transform $\TF{CM_i}{BF}$ by the six-dimensional pose vector $(\bm{\omega},\mathbf{v}_{\rm twist})$. At each iteration, the current intrinsics and pose are inserted into the forward model, checkerboard points are reprojected, and pixel-wise residuals are formed against the observed corners. The optimized intrinsics and extrinsics are exported in YAML format, together with per-frame and average reprojection RMSE.

 \section{Ultrasound Streaming Details}
\label{app:ultrasound_streaming}

\subsection{Acquisition}
Two acquisition paths are supported:
\begin{itemize}
\item \emph{DICOM:} A single multi-frame DICOM or a folder of per-frame DICOM files is parsed. Valid slices are stacked to produce $\mathcal V_{\rm US}\in\mathbb{R}^{N_{\rm US}\times H_{\rm img}\times W_{\rm img}}$ with spacing $(s_r,s_c,s_z)$.
\item \emph{Live streaming via OpenIGTLink:} Frames $\mathbf{I}_t\in\mathbb{R}^{H_{\rm img}\times W_{\rm img}}$ arrive as IMAGE messages. Intensities are converted to 8-bit when required for display, and each frame is timestamped upon reception.
\end{itemize}
The GUI maintains a recent cine buffer of $N_{\rm cine}$ frames
\begin{equation}
\mathcal B_{\rm cine}
=
\left[
\mathbf{I}_{t_0},\ldots,\mathbf{I}_{t_{N_{\rm cine}-1}}
\right],
\qquad
N_{\rm cine}\leq256
\end{equation}

\subsection{Threading and Latency Control}
Live streaming runs in a dedicated thread. New frames are pushed into a bounded queue:
\begin{equation}
\mathcal Q_{\rm US}
\leftarrow
(\mathbf{I}_t,t),
\qquad
\text{FIFO drop on overflow}
\end{equation}
When the queue is full, the oldest pending item is discarded. A GUI timer polls $\mathcal Q_{\rm US}$ at approximately 1-ms intervals and renders only the most recent item obtained during the current poll cycle.

\subsection{DICOM and Physical Coordinates}
The main paper retains the mapping from pixel/slice indices to the metric ultrasound frame $\{US\}$. When DICOM orientation and origin tags are available, the standard IJK-to-LPS mapping is
\begin{equation}
\bar{\mathbf x}_{LPS}=\AF{LPS}{IJK}\,\bar{\mathbf q}_{IJK},
\qquad \bar{\mathbf q}_{IJK}=[c,r,k,1]^{\mathrm{T}}
\end{equation}
where
\begin{equation}
\AF{LPS}{IJK}=\begin{bmatrix}
\mathbf E_{\rm DICOM}\mathbf S_{\rm US}&\mathbf o_{\rm LPS}\\
\mathbf0_{1\times3}&1
\end{bmatrix},\qquad\mathbf S_{\rm US}=\operatorname{diag}(s_c,s_r,s_z)
\end{equation}
For a uniformly spaced parallel stack, $\mathbf E_{\rm DICOM}=[\mathbf d_{\rm row},\mathbf d_{\rm col},\mathbf d_{\rm slice}]$ contains unit direction vectors: the first two come from Image Orientation (Patient), and the third follows the slice ordering. $\mathbf o_{\rm LPS}$ is Image Position (Patient) for the first slice, in millimeters. The affine map $\AF{LPS}{IJK}$ applies spacing once to the unscaled voxel indices; it is not a rigid $\mathbf T$ transform. Nonuniform or nonparallel slices require their individual positions and orientations~\cite{DICOM2026}. Coordinates are converted to centimeters for visualization and multimodal fusion, with translations and other model coordinates converted to the same unit before composing rigid transforms.

\subsection{ROI Definition and Binary Mask}
A user-drawn polygon
\begin{equation}
\mathcal P_{\rm ROI}
=
\{(c_i,r_i)\}_{i=1}^{N_{\rm ROI}}
\end{equation}
on the active frame induces the binary mask
\begin{equation}
\mathcal M_{\rm ROI}(c,r)
=
\begin{cases}
1, & (c,r)\in\mathrm{poly}(\mathcal P_{\rm ROI}),\\
0, & \mathrm{otherwise}.
\end{cases}
\end{equation}
Here $N_{\rm ROI}$ is the number of polygon vertices. The point-in-polygon mask can be smoothed by morphological opening and hole filling:
\begin{equation}
\mathcal M_{\rm ROI}
\leftarrow
\mathrm{Fill}\!\left(
\mathrm{Open}(\mathcal M_{\rm ROI};\rho_{\rm morph})
\right)
\end{equation}
Here $\rho_{\rm morph}$ is the morphological structuring-element radius. The masked image is
\begin{equation}
\mathbf{I}^{ROI}
=
\mathcal M_{\rm ROI}\odot\mathbf{I}
\end{equation}

\subsection{Interactive Landmarks}
Each landmark is stored as $(f,c,r,\ell)$, where $f$ is the frame index within the ultrasound stream, $(c,r)$ are image coordinates, and $\ell$ is the user-selected anatomical label. The landmark list is exported as a CSV file.

\subsection{Pseudo-Volume Assembly}
For a multi-frame DICOM or cine buffer, the pseudo-volume uses the slice-first array order defined in Section~\ref{ultrasound}; $N_{\rm US}=N_{\rm cine}$ for a cine-buffer reconstruction. With zero-based slice indexing,
\begin{equation}
\mathcal V_{\rm US}(k,:,:)
=
\mathbf{I}_{t_k},
\qquad
k=0,\ldots,N_{\rm US}-1
\end{equation}
A robust marching-cubes isosurface level is selected as
\begin{equation}
\tau_{\rm US}
=
\mathrm{quantile}\!\left(
\mathcal V_{\rm US},0.90
\right)
\end{equation}
Let $\mathbf X_{{\rm mesh},vox}\in\mathbb R^{3\times N_{\rm vert}}$ contain the extracted vertices as columns in $(c-c_0,r-r_0,k)$ order, relative to the local ultrasound origin. Metric scaling and conversion to centimeters are
\begin{equation}
\mathbf X_{{\rm mesh},cm}
=
\frac{1}{10}\,
\mathbf S_{\rm US}\mathbf X_{{\rm mesh},vox},
\qquad
\mathbf S_{\rm US}
=
\mathrm{diag}(s_c,s_r,s_z)
\end{equation}
Without tracked probe motion, this reconstruction is a stacked or cine pseudo-volume; true 3D compounding requires the external probe pose. The complete ultrasound-to-CT transform chain is retained in the main paper.

\subsection{Control Flow and Outputs}
The user loads a DICOM file/folder or starts an OpenIGTLink stream, displays the latest live frame or browses the offline stack, draws and exports ROIs, places and exports landmarks, and renders a marching-cubes surface from either a stored stack or the recent cine buffer. Input images are converted to 8-bit for display when needed, whereas physical dimensions are derived from DICOM metadata. Outputs comprise (i) a landmark CSV file, (ii) an optional ROI mask for each frame, and (iii) 3D meshes.
The key parameters used in ultrasound mode are summarized in Table~\ref{table:us_params_appendix}.
\begin{table}[H]
\centering
\caption{Key parameters in ultrasound mode.}
\label{table:us_params_appendix}
\begin{tabularx}{\linewidth}{>{\raggedright\arraybackslash}p{0.29\linewidth} >{\raggedright\arraybackslash}p{0.20\linewidth} >{\raggedright\arraybackslash}X}
\toprule
\textbf{Parameter} & \textbf{Symbol / Field} & \textbf{Typical Value / Note}\\
\midrule
Pixel spacing (row, col) & $(s_r,s_c)$ & DICOM Pixel Spacing (mm)\\
Slice spacing & $s_z$ & Spacing Between Slices (mm)\\
Live host/port & --- & OpenIGTLink\\
Cine buffer size & $N_{\rm cine}$ & $\leq256$ most recent frames\\
Queue policy & --- & Drop oldest on overflow\\
Isosurface level & $\tau_{\rm US}$ & $\mathrm{quantile}(\mathcal V_{\rm US},0.90)$\\
Units for fusion & --- & centimeters (cm)\\
\bottomrule
\end{tabularx}
\end{table}

\clearpage\bibliographystyle{asmems4}\bibliography{Ref}

\begin{thebibliography}{10}

\bibitem{george2018}
George, E.~I., Brand, C. T.~C., Marescaux, J., et~al., 2018,
\newblock ``Origins of robotic surgery: from skepticism to standard of care,''
\newblock {\em JSLS: Journal of the Society of Laparoendoscopic Surgeons, {\bf 22}}(4), pp.~e2018--00039. DOI: \url{10.4293/JSLS.2018.00039}.

\bibitem{morris2005}
Morris, B., 2005,
\newblock ``Robotic surgery: applications, limitations, and impact on surgical education,''
\newblock {\em Medscape General Medicine, {\bf 7}}(3), p.~72. \url{https://pmc.ncbi.nlm.nih.gov/articles/PMC1681689/}.

\bibitem{Lanfranco2004}
Lanfranco, A.~R., Castellanos, A.~E., Desai, J.~P., and Meyers, W.~C., 2004,
\newblock ``Robotic surgery: A current perspective,''
\newblock {\em Annals of Surgery, {\bf 239}}(1), Jan., pp.~14--21. DOI: \url{10.1097/01.sla.0000103020.19595.7d}.

\bibitem{Reddy2023}
Reddy, K., Gharde, P., Tayade, H., Patil, M., Reddy, L.~S., and Surya, D., 2023,
\newblock ``Advancements in robotic surgery: A comprehensive overview of current utilizations and upcoming frontiers,''
\newblock {\em Cureus, {\bf 15}}(12), p.~e50415
\newblock Published Dec 12, 2023. DOI: \url{10.7759/cureus.50415}.

\bibitem{Chuchulo2023}
Chuchulo, A., and Ali, A., 2023,
\newblock ``Is robotic-assisted surgery better?,''
\newblock {\em AMA Journal of Ethics, {\bf 25}}(8), pp.~E598--E604
\newblock Medical Education, Aug 2023, Peer-Reviewed. DOI: \url{10.1001/amajethics.2023.598}.

\bibitem{UFUrology2023}
{University of Florida, Department of Urology}, 2023,
\newblock Robotic nerve-sparing radical prostatectomy
\newblock \url{https://ufhealth.org/conditions-and-treatments/uf-health-robotic-prostatectomy}
\newblock Accessed: September 10, 2023.

\bibitem{Bessen2021}
Bessen, S.~Y., Wu, X., Sramek, M.~T., Shi, Y., Pastel, D., Halter, R., and Paydarfar, J.~A., 2021,
\newblock ``Image-guided surgery in otolaryngology: A review of current applications and future directions in head and neck surgery,''
\newblock {\em Head \& Neck, {\bf 43}}(8), Aug., pp.~2534--2553
\newblock Epub 2021 May 25. DOI: \url{10.1002/hed.26743}.

\bibitem{Cleary2010}
Cleary, K., and Peters, T.~M., 2010,
\newblock ``Image-guided interventions: technology review and clinical applications,''
\newblock {\em Annual Review of Biomedical Engineering, {\bf 12}}, August, pp.~119--142. DOI: \url{10.1146/annurev-bioeng-070909-105249}.

\bibitem{su2022s}
Su, H., Kwok, K.-W., Cleary, K., Iordachita, I., Cavusoglu, M.~C., Desai, J.~P., and Fischer, G.~S., 2022,
\newblock ``State of the art and future opportunities in mri-guided robot-assisted surgery and interventions,''
\newblock {\em Proceedings of the IEEE, {\bf 110}}(7), pp.~968--992. DOI: \url{10.1109/JPROC.2022.3169146}.

\bibitem{vanLeeuwen2024}
van Leeuwen, F. W.~B., Buckle, T., van Oosterom, M.~N., and Rietbergen, D. D.~D., 2024,
\newblock ``The rise of molecular image-guided robotic surgery,''
\newblock {\em Journal of Nuclear Medicine, {\bf 65}}(10), pp.~1505--1511. DOI: \url{10.2967/jnumed.124.267783}.

\bibitem{Wendler2021}
Wendler, T., van Leeuwen, F. W.~B., Navab, N., and van Oosterom, M.~N., 2021,
\newblock ``How molecular imaging will enable robotic precision surgery: The role of artificial intelligence, augmented reality, and navigation,''
\newblock {\em European Journal of Nuclear Medicine and Molecular Imaging, {\bf 48}}(13), December, pp.~4201--4224
\newblock Epub 2021 Jun 29. DOI: \url{10.1007/s00259-021-05445-6}.

\bibitem{Navaratnam2018}
Navaratnam, A., Abdul-Muhsin, H., and Humphreys, M., 2018,
\newblock ``Updates in urologic robot assisted surgery,''
\newblock {\em F1000Research, {\bf 7}}, p.~1948. DOI: \url{10.12688/f1000research.15480.1}.

\bibitem{Porpiglia2019}
Porpiglia, F., Checcucci, E., Amparore, D., Piramide, F., Volpi, G., Granato, S., et~al., 2019,
\newblock ``Augmented-reality robot-assisted radical prostatectomy using hyper-accuracy three-dimensional reconstruction (ha3d) technology: a radiological and pathological study,''
\newblock {\em BJU International, {\bf 123}}(5), pp.~834--845. DOI: \url{10.1111/bju.14549}.

\bibitem{Pratt2018}
Pratt, P., and Arora, A., 2018,
\newblock ``Transoral robotic surgery: Image guidance and augmented reality,''
\newblock {\em ORL Journal for Oto-Rhino-Laryngology and its Related Specialties, {\bf 80}}(3-4), pp.~204--212. DOI: \url{10.1159/000489467}.

\bibitem{liao2010}
Liao, H., Inomata, T., Sakuma, I., and Dohi, T., 2010,
\newblock ``3-d augmented reality for mri-guided surgery using integral videography autostereoscopic image overlay,''
\newblock {\em IEEE Transactions on Biomedical Engineering, {\bf 57}}(6), pp.~1476--1486. DOI: \url{10.1109/TBME.2010.2040278}.

\bibitem{chan2021}
Chan, H. H.~L., Haerle, S.~K., Daly, M.~J., Zheng, J., Philp, L., Ferrari, M., Douglas, C.~M., and Irish, J.~C., 2021,
\newblock ``An integrated augmented reality surgical navigation platform using multi-modality imaging for guidance,''
\newblock {\em PLoS One, {\bf 16}}(4), p.~e0250558. DOI: \url{10.1371/journal.pone.0250558}.

\bibitem{schiavina2021}
Schiavina, R., Bianchi, L., Lodi, S., Cercenelli, L., Chessa, F., Bortolani, B., Gaudiano, C., Casablanca, C., Droghetti, M., Porreca, A., Romagnoli, D., Golfieri, R., Giunchi, F., Fiorentino, M., Marcelli, E., Diciotti, S., and Brunocilla, E., 2021,
\newblock ``Real-time augmented reality three-dimensional guided robotic radical prostatectomy: Preliminary experience and evaluation of the impact on surgical planning,''
\newblock {\em European Urology Focus, {\bf 7}}(6), pp.~1260--1267
\newblock Epub 2020 Sep 1. DOI: \url{10.1016/j.euf.2020.08.004}.

\bibitem{Amparore2024}
Amparore, D., Sica, M., Verri, P., Piramide, F., Checcucci, E., De~Cillis, S., Piana, A., Campobasso, D., Burgio, M., Cisero, E., Busacca, G., Di~Dio, M., Piazzolla, P., Fiori, C., and Porpiglia, F., 2024,
\newblock ``Computer vision and machine-learning techniques for automatic 3d virtual images overlapping during augmented reality guided robotic partial nephrectomy,''
\newblock {\em Technology in Cancer Research \& Treatment, {\bf 23}}, p.~15330338241229368. DOI: \url{10.1177/15330338241229368}.

\bibitem{Piana2024}
Piana, A., Amparore, D., Sica, M., Volpi, G., Checcucci, E., Piramide, F., De~Cillis, S., Busacca, G., Scarpelli, G., Sidoti, F., Alba, S., Piazzolla, P., Fiori, C., Porpiglia, F., and Di~Dio, M., 2024,
\newblock ``Automatic 3d augmented-reality robot-assisted partial nephrectomy using machine learning: Our pioneer experience,''
\newblock {\em Cancers, {\bf 16}}(5). DOI: \url{10.3390/cancers16051047}.

\bibitem{Giannone2021}
Giannone, F., Felli, E., Cherkaoui, Z., Mascagni, P., and Pessaux, P., 2021,
\newblock ``Augmented reality and image-guided robotic liver surgery,''
\newblock {\em Cancers, {\bf 13}}(24). DOI: \url{10.3390/cancers13246268}.

\bibitem{Checcucci2024}
Checcucci, E., Piana, A., Volpi, G., Quar{\`a}, A., De~Cillis, S., Piramide, F., Burgio, M., Meziere, J., Cisero, E., Colombo, M., Bignante, G., Sica, M., Granato, S., Verri, P., Gatti, C., Alessio, P., Di~Dio, M., Alba, S., Fiori, C., Amparore, D., and Porpiglia, F., 2024,
\newblock ``Visual extended reality tools in image-guided surgery in urology: a systematic review,''
\newblock {\em European Journal of Nuclear Medicine and Molecular Imaging, {\bf 51}}(10), pp.~3109--3134
\newblock Epub 2024 Apr 9. DOI: \url{10.1007/s00259-024-06699-6}.

\bibitem{Manfredi2022}
Manfredi, M., Piramide, F., Amparore, D., Burgio, M., Busacca, G., Colombo, M., Meziere, J., Granato, S., Piana, A., De~Cillis, S., Checcucci, E., Fiori, C., and Porpiglia, F., 2022,
\newblock ``Augmented reality: the smart way to guide robotic urologic surgery,''
\newblock {\em Mini-invasive Surgery, {\bf 6}}, p.~40. DOI: \url{10.20517/2574-1225.2022.37}.

\bibitem{Chan20}
Chan, J. Y.~K., Holsinger, F.~C., Liu, S., Sorger, J.~M., Azizian, M., and Tsang, R. K.~Y., 2020,
\newblock ``Augmented reality for image guidance in transoral robotic surgery,''
\newblock {\em Journal of Robotic Surgery, {\bf 14}}(4), pp.~579--583
\newblock Epub 2019 Sep 25. DOI: \url{10.1007/s11701-019-01030-0}.

\bibitem{Shi24}
Shi, Y., Kokko, M., Paydarfar, J., and Halter, R., 2024,
\newblock ``A surgical navigation framework for image-guided transoral robotic surgery,''
\newblock In Medical Imaging 2024: Image-Guided Procedures, Robotic Interventions, and Modeling, Vol.~12928, SPIE, p.~1292804. DOI: \url{10.1117/12.3008805}.

\bibitem{shapey2021}
Shapey, J., Dowrick, T., Delaunay, R., Mackle, E.~C., Thompson, S., Janatka, M., Guichard, R., Georgoulas, A., P{\'e}rez-Su{\'a}rez, D., Bradford, R., et~al., 2021,
\newblock ``Integrated multi-modality image-guided navigation for neurosurgery: open-source software platform using state-of-the-art clinical hardware,''
\newblock {\em International journal of computer assisted radiology and surgery, {\bf 16}}(8), pp.~1347--1356. DOI: \url{10.1007/s11548-021-02374-5}.

\bibitem{Ukimura2014}
Ukimura, O., Aron, M., Nakamoto, M., Shoji, S., Abreu, A.~L., Matsugasumi, T., Berger, A., Desai, M., and Gill, I.~S., 2014,
\newblock ``Three-dimensional surgical navigation model with tilepro display during robot-assisted radical prostatectomy,''
\newblock {\em Journal of Endourology, {\bf 28}}(6), June, pp.~625--630
\newblock Epub 2014 May 8. DOI: \url{10.1089/end.2013.0749}.

\bibitem{Samei2020}
Samei, G., Tsang, K., Kesch, C., Lobo, J., Hor, S., Mohareri, O., Chang, S., Goldenberg, S.~L., Black, P.~C., and Salcudean, S., 2020,
\newblock ``A partial augmented reality system with live ultrasound and registered preoperative mri for guiding robot-assisted radical prostatectomy,''
\newblock {\em Medical Image Analysis, {\bf 60}}, Feb., p.~101588
\newblock Epub 2019 Oct 29. DOI: \url{10.1016/j.media.2019.101588}.

\bibitem{Song2024}
Song, H., Yang, S., Wu, Z., et~al., 2024,
\newblock ``Arc-to-line frame registration method for ultrasound and photoacoustic image-guided intraoperative robot-assisted laparoscopic prostatectomy,''
\newblock {\em International Journal of Computer Assisted Radiology and Surgery, {\bf 19}}, pp.~199--208. DOI: \url{10.1007/s11548-023-02984-1}.

\bibitem{Meza2021}
Meza, J., Contreras-Ortiz, S., Romero, L., and Marrugo, A., 2021,
\newblock ``Three-dimensional multimodal medical imaging system based on freehand ultrasound and structured light,''
\newblock {\em Optical Engineering, {\bf 60}}, 05. DOI: \url{10.1117/1.OE.60.5.054106}.

\bibitem{Stoyanov2010}
Stoyanov, D., Scarzanella, M.~V., Pratt, P., and Yang, G.~Z., 2010,
\newblock ``Real-time stereo reconstruction in robotically assisted minimally invasive surgery,''
\newblock In Medical Image Computing and Computer-Assisted Intervention (MICCAI 2010), Vol.~6361 of {\em Lecture Notes in Computer Science}, pp.~275--282. DOI: \url{10.1007/978-3-642-15705-9_34}.

\bibitem{Marcus2024}
Marcus, H.~J., Ramirez, P.~T., Khan, D.~Z., Horsfall, H.~L., Hanrahan, J.~G., Williams, S.~C., Beard, D.~J., Bhat, R., Catchpole, K., Cook, A., Hutchison, K., Martin, J., Melvin, T., Stoyanov, D., Rovers, M., Raison, N., Dasgupta, P., Noonan, D., Stocken, D., Sturt, G., Vanhoestenberghe, A., Vasey, B., and McCulloch, P., 2024,
\newblock ``The ideal framework for surgical robotics: development, comparative evaluation and long-term monitoring,''
\newblock {\em Nature Medicine, {\bf 30}}, pp.~61--75. DOI: \url{10.1038/s41591-023-02732-7}.

\bibitem{pybind11}
Jakob, W., Rhinelander, J., and Moldovan, D., 2017,
\newblock {pybind11 -- Seamless operability between C++11 and Python}
\newblock \url{https://github.com/pybind/pybind11}
\newblock Accessed: 2025-07-21.

\bibitem{Husar2024}
Husar, A., Ordyan, M., Garcia, G.~C., Yancey, J.~G., Saglam, A.~S., Faeder, J.~R., Bartol, T.~M., Kennedy, M.~B., and Sejnowski, T.~J., 2024,
\newblock ``Mcell4 with bionetgen: A monte carlo simulator of rule-based reaction-diffusion systems with python interface,''
\newblock {\em PLoS Computational Biology, {\bf 20}}(4), Apr., p.~e1011800
\newblock Published 2024 Apr 24. DOI: \url{10.1371/journal.pcbi.1011800}.

\bibitem{Adorno2021}
Adorno, B.~V., and Marques~Marinho, M., 2021,
\newblock ``Dq robotics: A library for robot modeling and control,''
\newblock {\em IEEE Robotics \& Automation Magazine, {\bf 28}}(3), pp.~102--116. DOI: \url{10.1109/MRA.2020.2997920}.

\bibitem{Berenz2021}
Berenz, V., Naveau, M., Widmaier, F., Wüthrich, M., Passy, J.-C., Guist, S., and Büchler, D., 2021,
\newblock ``The o80 c++ templated toolbox: Designing customized python apis for synchronizing realtime processes,''
\newblock {\em Journal of Open Source Software, {\bf 6}}(66), p.~2752. DOI: \url{10.21105/joss.02752}.

\bibitem{Arun1987}
Arun, K.~S., Huang, T.~S., and Blostein, S.~D., 1987,
\newblock ``Least-squares fitting of two 3-d point sets,''
\newblock {\em IEEE Transactions on Pattern Analysis and Machine Intelligence, {\bf PAMI-9}}(5), pp.~698--700. DOI: \url{10.1109/TPAMI.1987.4767965}.

\bibitem{Tsai1989}
Tsai, R., and Lenz, R., 1989,
\newblock ``A new technique for fully autonomous and efficient 3d robotics hand/eye calibration,''
\newblock {\em IEEE Transactions on Robotics and Automation, {\bf 5}}(3), pp.~345--358. DOI: \url{10.1109/70.34770}.

\bibitem{Zhang2000}
Zhang, Z., 2000,
\newblock ``A flexible new technique for camera calibration,''
\newblock {\em IEEE Transactions on Pattern Analysis and Machine Intelligence, {\bf 22}}(11), pp.~1330--1334. DOI: \url{10.1109/34.888718}.

\bibitem{Tokuda2009}
Tokuda, J., Fischer, G.~S., Papademetris, X., Yaniv, Z., Ibanez, L., Cheng, P., Liu, H., Blevins, J., Arata, J., Golby, A.~J., Kapur, T., Pieper, S., Burdette, E.~C., Fichtinger, G., Tempany, C.~M., and Hata, N., 2009,
\newblock ``Openigtlink: An open network protocol for image-guided therapy environment,''
\newblock {\em The International Journal of Medical Robotics and Computer Assisted Surgery, {\bf 5}}(4), pp.~423--434. DOI: \url{10.1002/rcs.274}.

\bibitem{Lasso2014}
Lasso, A., Heffter, T., Rankin, A., Pinter, C., Ungi, T., and Fichtinger, G., 2014,
\newblock ``{PLUS}: Open-source toolkit for ultrasound-guided intervention systems,''
\newblock {\em IEEE Transactions on Biomedical Engineering, {\bf 61}}(10), pp.~2527--2537. DOI: \url{10.1109/TBME.2014.2322864}.

\bibitem{Fedorov2012}
Fedorov, A., Beichel, R., Kalpathy-Cramer, J., Finet, J., Fillion-Robin, J.-C., Pujol, S., Bauer, C., Jennings, D., Fennessy, F., Sonka, M., Buatti, J., Aylward, S., Miller, J.~V., Pieper, S., and Kikinis, R., 2012,
\newblock ``{3D Slicer} as an image computing platform for the quantitative imaging network,''
\newblock {\em Magnetic Resonance Imaging, {\bf 30}}(9), pp.~1323--1341. DOI: \url{10.1016/j.mri.2012.05.001}.

\bibitem{Macenski2022}
Macenski, S., Foote, T., Gerkey, B., Lalancette, C., and Woodall, W., 2022,
\newblock ``Robot operating system 2: Design, architecture, and uses in the wild,''
\newblock {\em Science Robotics, {\bf 7}}(66), p.~eabm6074. DOI: \url{10.1126/scirobotics.abm6074}.

\bibitem{kokko2023}
Kokko, M.~A., Shi, Y., Paydarfar, J.~A., and Halter, R.~J., 2023,
\newblock ``Initial image-plane assessment of kinematic-based stereo overlay in trans-oral robotic surgery,''
\newblock In Medical Imaging 2023: Image-Guided Procedures, Robotic Interventions, and Modeling, Vol.~12466, SPIE, pp.~607--612. DOI: \url{10.1117/12.2655867}.

\bibitem{DICOM2026}
{National Electrical Manufacturers Association}, 2026,
\newblock Digital imaging and communications in medicine (dicom) standard, ps3.3: Information object definitions
\newblock DICOM Standard. \url{https://dicom.nema.org/medical/dicom/current/output/html/part03.html}.

\bibitem{Lorensen1987}
Lorensen, W.~E., and Cline, H.~E., 1987,
\newblock ``Marching cubes: A high resolution 3d surface construction algorithm,''
\newblock {\em ACM SIGGRAPH Computer Graphics, {\bf 21}}(4), pp.~163--169. DOI: \url{10.1145/37402.37422}.

\bibitem{lipson2021}
Lipson, L., Teed, Z., and Deng, J., 2021,
\newblock ``Raft-stereo: Multilevel recurrent field transforms for stereo matching,''
\newblock In 2021 International Conference on 3D Vision (3DV), IEEE, pp.~218--227. DOI: \url{10.1109/3DV53792.2021.00032}.

\bibitem{chang2018}
Chang, J.-R., and Chen, Y.-S., 2018,
\newblock ``Pyramid stereo matching network,''
\newblock In Proceedings of the IEEE conference on computer vision and pattern recognition, pp.~5410--5418. DOI: \url{10.1109/CVPR.2018.00567}.

\bibitem{EditorialPybind11}
{pybind11 contributors}, n.d.,
\newblock {pybind11} documentation: {NumPy} and the buffer protocol
\newblock \url{https://pybind11.readthedocs.io/en/stable/advanced/pycpp/numpy.html}
\newblock Accessed September 25, 2026.

\bibitem{Fitzpatrick1998}
Fitzpatrick, J.~M., West, J.~B., and Maurer, Jr., C.~R., 1998,
\newblock ``Predicting error in rigid-body point-based registration,''
\newblock {\em IEEE Transactions on Medical Imaging, {\bf 17}}(5), pp.~694--702. DOI: \url{10.1109/42.736021}.

\bibitem{Torres2025}
Torres, V.~C., Streeter, S.~S., Nguyen, M., Davis, S.~C., and Samkoe, K.~S., 2025,
\newblock ``Depth imaging using multi-wavelength excitation-emission ratios,''
\newblock In Optical Molecular Probes, Imaging and Drug Delivery, Optica Publishing Group, pp.~OW1E--2. DOI: \url{10.1364/OMP.2025.OW1E.2}.

\end{thebibliography}
\end{document}